\PassOptionsToPackage{table,prologue,dvipsnames}{xcolor}
\documentclass[sigconf]{acmart}
\usepackage{amsmath,amsfonts}
\usepackage{algorithmic}
\usepackage{array}
\usepackage{textcomp}
\usepackage{stfloats}
\usepackage{url}
\usepackage{verbatim}
\usepackage{graphicx}
\usepackage[utf8]{inputenc}
\usepackage{graphicx}
\usepackage{tikz}
\usetikzlibrary{calc, positioning}
\usepackage{caption}
\usepackage{subcaption}
\usepackage{algorithm}
\usepackage{svg}
\usepackage{xcolor}

\usepackage{xspace}
\makeatletter
\DeclareRobustCommand\onedot{\futurelet\@let@token\@onedot}
\def\@onedot{\ifx\@let@token.\else.\null\fi\xspace}
\def\eg{\emph{e.g}\onedot}

\def\etal{\emph{et al}\onedot}
\AtBeginDocument{%
  }

\copyrightyear{2025}
\acmYear{2025}
\setcopyright{acmlicensed}
\acmConference[MM '25] {Proceedings of the 33rd ACM International Conference on Multimedia}{October 27--31, 2025}{Dublin, Ireland.}
\acmBooktitle{Proceedings of the 33rd ACM International Conference on Multimedia (MM '25), October 27--31, 2025, Dublin, Ireland}
\acmISBN{979-8-4007-2035-2/2025/10}
\acmDOI{10.1145/3746027.3755429}
\DeclareCaptionFont{Ablationfont}{\fontsize{4pt}{4.8pt}\selectfont}

\newcommand{\ablationpicsize}{1.3cm}
\newcommand{\ablationfiggap}{1.4cm}
\newcommand{\ablationhgap}{2.2cm}
\DeclareCaptionFont{ablationfont}{\fontsize{4pt}{4.8pt}\selectfont}

\newcommand{\addpicsize}{1.3cm} 
\DeclareCaptionFont{addfont}{\fontsize{6pt}{4.8pt}\selectfont}

\newcommand{\comparisonscale}{1}
\DeclareCaptionFont{comparisonfont}{\fontsize{6pt}{7.2pt}\selectfont}

\DeclareCaptionFont{expermentfont}{\fontsize{6pt}{7.2pt}\selectfont}

\DeclareCaptionFont{fristfont}{\fontsize{9pt}{9.8pt}\selectfont}

\newcommand{\locationpicsize}{1.3cm} 
\DeclareCaptionFont{locationfont}{\fontsize{4pt}{4.8pt}\selectfont}

\newcommand{\maskpicsize}{1.3cm}

\newcommand{\realpicsize}{1.3cm}
\DeclareCaptionFont{realfont}{\fontsize{5pt}{6pt}\selectfont}

\newcommand{\replacepicsize}{1.6cm}
\newcommand{\replacetextwidth}{2.2cm}

\newcommand{\stylepicsize}{1.3cm}
\DeclareCaptionFont{stylefont}{\fontsize{5pt}{6pt}\selectfont}

\DeclareCaptionFont{unseenfont}{\fontsize{6pt}{7.2pt}\selectfont}

\newcommand{\comparisionsize}{1.3cm}

\begin{document}

\title{Prompt-Softbox-Prompt: A Free-Text Embedding Control for Image Editing}

\author{Yitong Yang}
\email{2023310181@stu.sufe.edu.cn}
\affiliation{%
  \institution{Shanghai University of Finance and Economics}
  \city{Shanghai}
  \country{China}
}
\author{Yinglin Wang}
\authornote{Corresponding author.}
\email{wang.yinglin@shufe.edu.cn}
\affiliation{%
  \institution{School of Computing and Artificial Intelligence, Shanghai University of Finance and Economics}
  \city{Shanghai}
  \country{China}
}

\author{Tian Zhang}
\email{zhangtian@stu.sufe.edu.cn}
\affiliation{%
  \institution{Shanghai University of Finance and Economics}
  \city{Shanghai}
  \country{China}
}

\author{Jing Wang}
\email{wangjing0723@stu.sufe.edu.cn}
\affiliation{%
  \institution{Shanghai University of Finance and Economics}
  \city{Shanghai}
  \country{China}
}

\author{Shuting He}
\authornotemark[1]
\email{shuting.he@sufe.edu.cn}
\affiliation{%
  \institution{MoE Key Laboratory of Interdisciplinary Research of Computation and Economics, Shanghai University of Finance and Economics}
  \city{Shanghai}
  \country{China}}

\renewcommand{\shortauthors}{Yitong Yang, Yinglin Wang, Tian Zhang, Jing Wang, \& Shuting He}

\begin{abstract}
While text-driven diffusion models demonstrate remarkable performance in image editing, the critical components of their text embeddings remain underexplored. The ambiguity and entanglement of these embeddings pose challenges for precise editing. In this paper, we provide a comprehensive analysis of text embeddings in Stable Diffusion XL, offering three key insights: (1) \textit{aug embedding}~\footnote{\textit{aug embedding} is obtained by combining the pooled output of the final text encoder with the timestep embeddings. \url{https://github.com/huggingface/diffusers}} retains complete textual semantics but contributes minimally to image generation as it is only fused via the ResBlocks. More text information weakens its local semantics while preserving most global semantics. (2) \textit{BOS} and \textit{padding embedding} do not contain any semantic information. (3) \textit{EOS} holds the semantic information of all words and stylistic information. Each word embedding is important and does not interfere with the semantic injection of other embeddings. Based on these insights, we propose PSP (\textbf{P}rompt-\textbf{S}oftbox-\textbf{P}rompt), a training-free image editing method that leverages free-text embedding.  PSP enables precise image editing by modifying text embeddings within the cross-attention layers and using Softbox to control the specific area for semantic injection.  This technique enables the addition and replacement of objects without affecting other areas of the image. Additionally, PSP can achieve style transfer by simply replacing text embeddings. Extensive experiments show that PSP performs remarkably well in tasks such as object replacement, object addition, and style transfer. Our code is available at https://github.com/yangyt46/PSP.
\end{abstract}

\ccsdesc[500]{Computing methodologies~Computer vision}

\keywords{Stable diffusion; Text embedding; Image editing; Training-free; Prompt-Softbox-Prompt.}
\begin{teaserfigure}
\centering
  \includegraphics[scale=0.43]{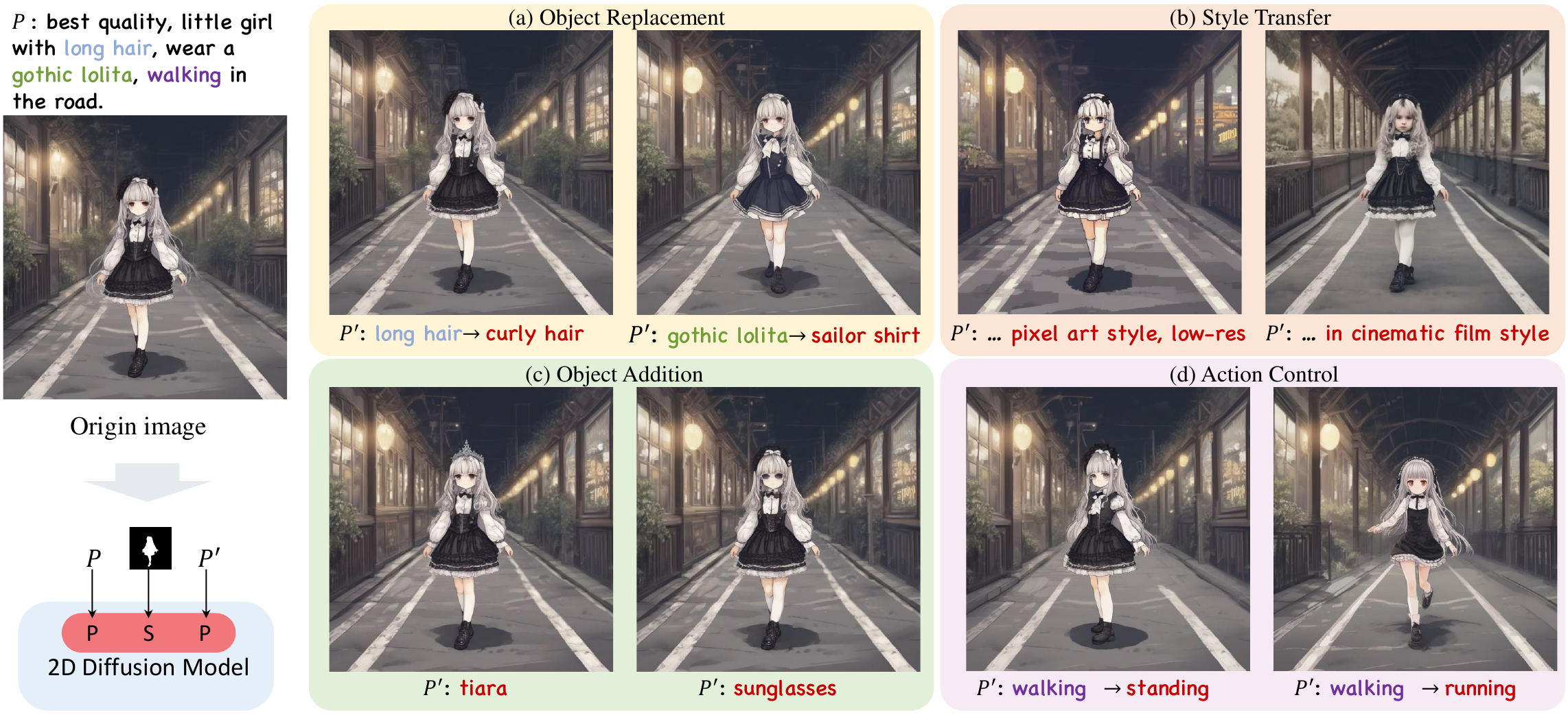}
\vspace{-10pt}
  \caption{Prompt-Softbox-Prompt for image editing. Our method allows for object replacement, object addition, action control, and style transfer by simply manipulating text embeddings. $P$ represents the source prompt, and $P^{'}$ represents the target prompt. Words of the same color indicate corresponding object changes.}
  \label{fig:teaser}
\end{teaserfigure}


\maketitle
\section{Introduction}
Text-guided diffusion models~\cite{kim2022diffusionclip,wu2023sketch,cao2023texfusion} have demonstrated exceptional performance in image generation, achieving state-of-the-art realism and diversity. By leveraging strong theoretical foundations and advanced network architectures, these models have been widely applied to diverse image editing tasks. However, previous methods~\cite{brooks2023instructpix2pix,kim2022diffusionclip,yang2023paint} typically require extensive training on large-scale datasets to achieve optimal performance in image editing. As a result, recent research has increasingly focused on developing innovative training-free approaches that effectively enable various advanced image editing functions through improved inference processes, including but not limited to object replacement~\cite{cao2023masactrl,24,mokady2023null} and style transfer~\cite{wang2023stylediffusion,hertz2024style}.

Despite significant advancements in training-free image editing methods, the intricate mapping mechanism between text embeddings and image generation remains insufficiently explored. Text embeddings inherently exhibit issues of ambiguity and entanglement, where even subtle variations can lead to substantial semantic discrepancies. While existing methods have explored word embeddings to some extent, the roles of special embeddings (such as \textit{BOS}, \textit{EOS}, etc.) remain ambiguous. Moreover, the injection of text embedding information is not only achieved through the cross-attention mechanism but also involves feature fusion within ResBlocks. These complexities present significant challenges for precise image editing based on text embeddings.


To address these challenges, we conduct a comprehensive analysis of how text embeddings influence image generation, highlighting their immense potential for controllable editing. We specifically investigate the relationship between text embeddings and image features in the Stable Diffusion XL (SDXL) model, uncovering three key insights: (1) \textit{aug embedding} contains complete textual semantics but contributes little to image generation. As the density of textual information increases, local semantics are lost, while global features remain largely unaffected. (2) \textit{BOS} and \textit{padding embedding} don’t contain any textual semantic information. (3) \textit{EOS} includes the semantic information of all words and style features. Each word embedding is crucial to the generation process and does not interfere with the semantic injection of others.

Based on these key insights, we propose a method called Prompt-Softbox-prompt (PSP), which enables arbitrary image editing without requiring additional optimization or training. Our core idea is to edit the source and target text embeddings within the cross-attention layers, such as by replacing or inserting them, to achieve the desired modifications. The target text embedding is a freely defined text embedding that does not need to adhere to the strict formatting constraints of the source text. For the object replacement task, we replace specific object tokens in the source text embedding with those from the target embedding. A Softbox, obtained from the segmentation model, constrains the source object's region derived using the OTSU algorithm~\cite{otsu1975threshold}, and injects the semantic information from the target text embedding into this region (based on Insight 1 and Insight 3). For object addition, we insert the target text embedding into the padding part of the source embedding (based on Insight 2). For style transfer, we replace the style words, \textit{EOS}, and \textit{aug embedding} in the source embedding to achieve the desired effect effortlessly (based on Insight 1 and Insight 3). We visually demonstrate our method across various tasks, including object replacement, action control, object addition, and style transfer, as depicted in Fig.~\ref{fig:teaser}. These examples show that our method can flexibly replace or add specified targets without altering the rest of the image and can effectively adjust the overall style. The experimental results demonstrate that our method achieves excellent performance in both qualitative and quantitative evaluations. Compared to existing methods, PSP can also be extended to local object editing and novel content generation.

\vspace{-2pt}
In summary, our contributions are as follows:
\begin{itemize}
\vspace{-2pt}
	\item We conduct a comprehensive analysis of the role of key components of text embeddings in image generation and provide three key insights.
	\item We propose PSP, a training-free image editing method that leverages free-text embedding. Our method modifies images by replacing and inserting text embeddings within the cross-attention layers while using Softbox to precisely control the semantic injection of the target text embedding.
	\item Extensive experiments comprehensively demonstrate the superior performance of our method across various image editing tasks, further substantiating its effectiveness.
\end{itemize}

\def\etal{\emph{et al}\onedot}
\section{Related Works}
\textbf{Foundation Model for Text-to-Image Generation.} 
In recent years, the field of text-to-image generation has experienced a revolutionary shift, significantly driven by the rapid rise of diffusion models~\cite{1,2,3}. Compared to traditional Generative Adversarial Networks (GANs)\cite{4,5,6}, these text-to-image (T2I) diffusion models excel at generating high-quality images from text prompts. Due to their robust theoretical foundations and flexible architectural design, diffusion models have quickly been applied to a broad range of visual tasks, including image segmentation\cite{7,8}, image translation~\cite{9,10}, object detection~\cite{11,12}, and video generation~\cite{13,14}. For example, Imagen~\cite{15} enhances the fidelity of generated samples by incorporating pre-trained large language models (\eg, T5~\cite{raffel2020exploring}). DALL·E 3~\cite{16} improves prompt-following ability by training with highly descriptive image captions. Compared to earlier versions of Stable Diffusion~\cite{2}, SDXL~\cite{17} utilizes a UNet backbone three times larger and two text encoders, along with several new adjustment schemes. The emergence of these advanced T2I base models has led researchers to further explore how to precisely control image generation through additional conditions such as masks~\cite{couairon2022diffedit,wang2023instructedit} and depth maps~\cite{duan2024diffusiondepth,zhang2023adding}.

\begin{figure}[t]
	\centering
	\includegraphics[width=\linewidth]{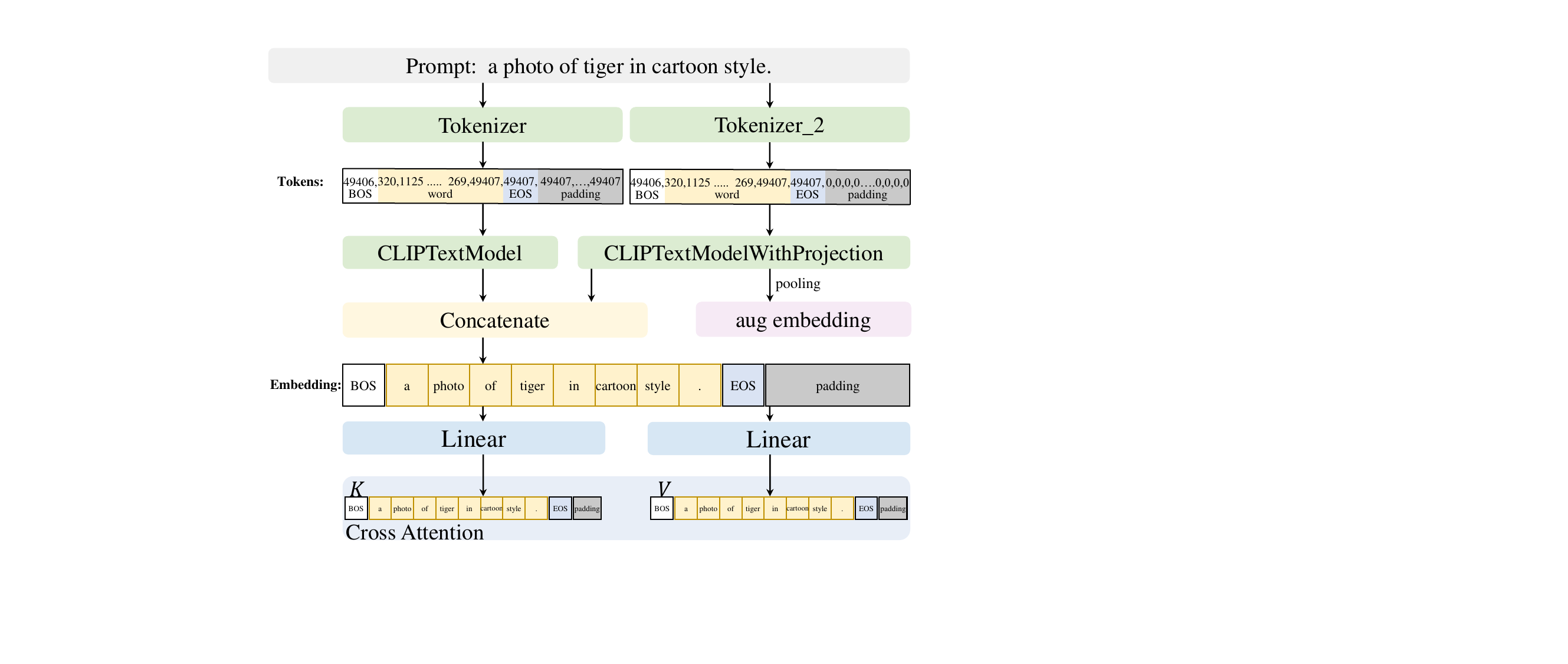}
        \vspace{-20pt}
	\caption{Text-driven image generation process in SDXL.}
	\label{fig:text_embedding}
    \vspace{-20pt}
\end{figure}
\noindent \textbf{Image Generation Conditioned on Text Embeddings.} Text encoders~\cite{radford2021learning} transform text conditions into embeddings to guide T2I models. Since minor embedding variations can substantially affect outputs, text embedding research has become crucial. Prior work has demonstrated their broad applicability~\cite{19,20,21,22,23,butt2024colorpeel}. For instance, Gal~\etal~\cite{19} facilitated personalized creation by converting images into text embeddings and combining them with other text embeddings. Han~\etal~\cite{23} decomposed the CLIP embedding space for personalization and content manipulation. Shi~\etal~\cite{21} utilized image encoders to convert input images into global embeddings and integrated new adapter layers into pre-trained models to enhance their ability to capture complex identity details while maintaining linguistic coherence. However, these methods do not sufficiently explore the intrinsic relationship between text embeddings and images. Recent studies~\cite{24,25,26} have imposed strict constraints on embeddings (e.g., cross-attention modification~\cite{24}, SVD analysis~\cite{25}, or region-specific injection~\cite{26}), limiting flexible combinations. Our work relaxes these constraints while investigating text embeddings' intrinsic properties more deeply, enabling flexible embedding manipulation for efficient image editing.

\noindent \textbf{MASK-Guided and Attention-Controlled Diffusion Models.} In diffusion models, MASK-guided~\cite{27,28,29,hertz2024style,31} techniques and attention control~\cite{32,chen2024training,34,35,36,chefer2023attend,yang2023dynamic} have emerged as key research directions. These approaches enable precise adjustment of specific image regions without altering the background through effective MASK and attention mechanisms. TEWEL~\etal~\cite{35} ensured thematic consistency by sharing internal activations of pre-trained models with theme-driven shared attention blocks and feature injections. Hertz~\etal~\cite{hertz2024style} proposed StyleAligned, maintaining style consistency in T2I models via minimal ``attention sharing" during the diffusion process. Chen~\etal~\cite{chen2024training} manipulated attention maps within the cross-attention layers of the model to guide image generation based on user-specified layouts. Chen~\etal~\cite{31} introduced MGPF, which integrates object masks into both aligned and misaligned parts of visual controls and prompts, designing a network that utilizes these masks to generate objects in misaligned visual control regions. While these methods perform well in their tasks, they still struggle with complex or unknown scenarios.

\section{Methods}
In the following section, we conduct a comprehensive analysis and discussion of text embeddings in SDXL. This provides critical insights for our subsequent work. Based on these insights, we develop the PSP method to enable controllable image editing.

\subsection{Comprehensive Analysis of Text Embedding}
Our method uses text embeddings for image editing, requiring a thorough analysis of their relationship with generated images. While most existing approaches guide latent optimization through backpropagation, the role of text embeddings in diffusion models, especially within the Stable Diffusion series, remains insufficiently explored. To bridge this gap, we analyze text embeddings in SDXL to enhance editing efficiency and effectiveness. As shown in Fig.~\ref{fig:text_embedding}, SDXL processes a text prompt through two tokenizers (differing primarily in padding) and two CLIP text encoders, producing embeddings of varying dimensions ($B\times77\times768$ and $B\times77\times1280$, where $B$ denotes batch size). These embeddings are concatenated to form the final text embedding, which is then mapped to Key and Value in the cross-attention layers to inject rich textual context into the latent. Unlike earlier versions of Stable Diffusion, SDXL also extracts semantic information from the pooled output (with dimension $B\times1280$) of the final text encoder, combining it with timestep embeddings to create the \textit{aug embedding}, which is then fused with the latent via the ResNet blocks.
\begin{figure}[!t]
	\centering
	\includegraphics[width=\linewidth]{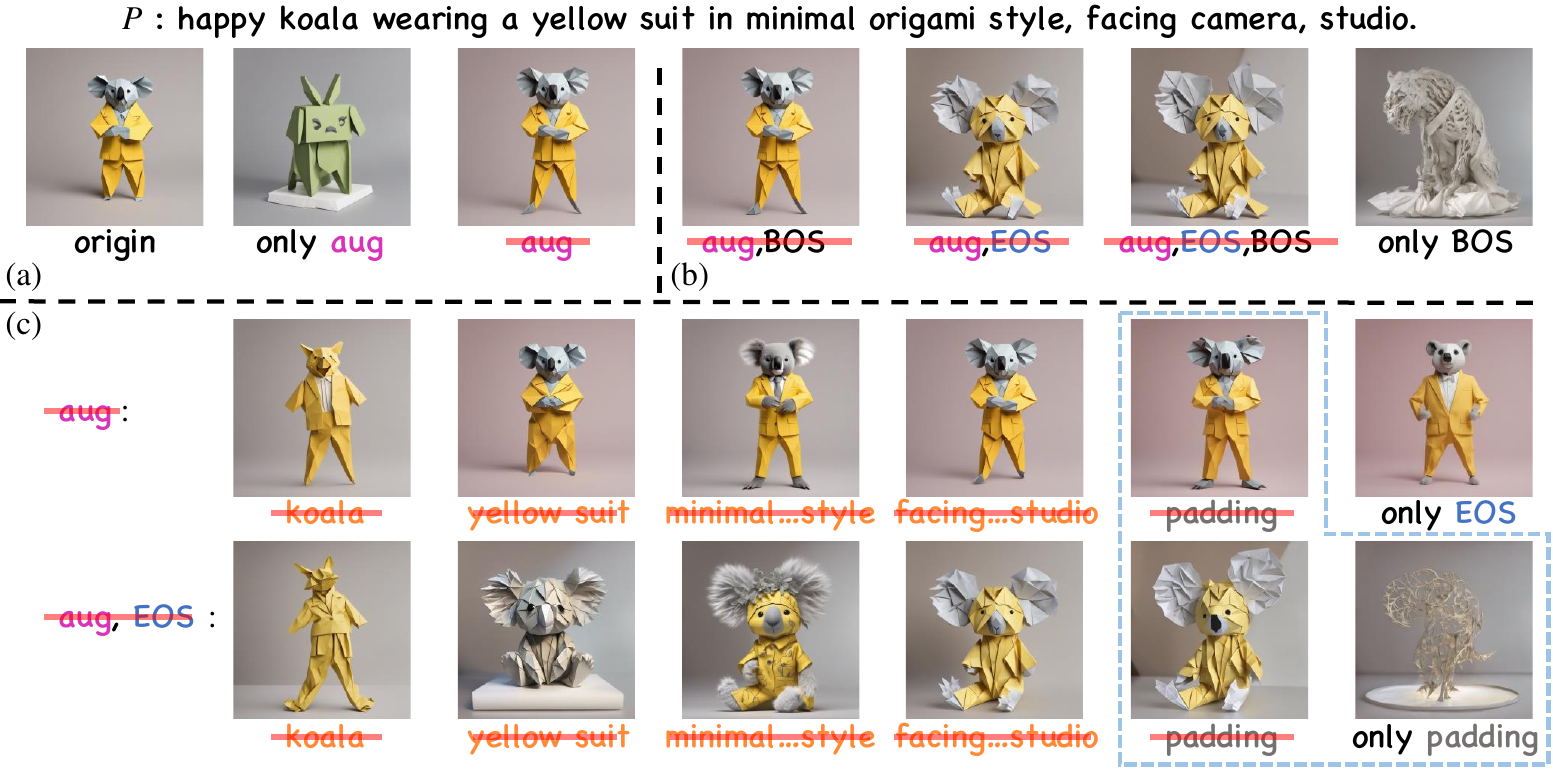}
    \vspace{-18pt}
	\caption{Impact of text embeddings on image generation by masking specified tokens. Red line: Mask; ``aug'': \textit{aug embedding}; ``only'': generate with specified tokens only.}
	\label{fig:experment}
    \vspace{-10pt}
\end{figure}
\begin{table}[!t]
\centering
\small
\setlength{\tabcolsep}{11pt}
\caption{Ablation study on text embedding. ``only'': generate with specified tokens only. ``aug'' denotes \textit{aug embedding}. ``90\%/7\%'' represent the generation probability for the primary object (cat) and secondary object (sweater), respectively.}
\vspace{-10pt}
\begin{tabular}{cccc}
\specialrule{.1em}{.05em}{.05em}
                       & Object $\uparrow$            & Style $\uparrow$          & Image Quality $\uparrow$         \\ \hline
\rowcolor[HTML]{EFEFEF} 
Origin                 & 100\%                  & 100\%              & 7.36             \\
Mask BOS               & 100\%                  & 100\%              & 7.29             \\
Only BOS               & 0                  & 0              & 2.4              \\
\rowcolor[HTML]{EFEFEF} 
Mask EOS               & 87\%               & 95\%           & 6.98             \\
\rowcolor[HTML]{EFEFEF} 
Only EOS               & 93\%               & 97\%           & 7.11             \\
Mask object            & 84\%               & 100\%              & 6.97             \\
Mask style             & 98\%               & 99\%           & 6.91             \\
\rowcolor[HTML]{EFEFEF} 
Mask padding           & 99\%               & 99\%           & 7.11             \\
\rowcolor[HTML]{EFEFEF} 
Only padding           & 0                  & 0              & 2.6              \\
Mask aug               & 100\%                  & 98\%           & 7.38             \\
Only aug               & 39\%               & 72\%           & 6.04             \\ \hline
\multicolumn{4}{c}{\cellcolor[HTML]{FFFFFF}Only aug (Different prompt)} \\ \hline
1 token               & 99\%               & -              & 6.14             \\
9 tokens               & 90\%/7\%           & 74\%           & 5.88             \\
65 tokens              & 0                  & 70\%            & 5.28             \\ 
\specialrule{.1em}{.05em}{.05em}
\end{tabular}
\label{tab:motivation}
\vspace{-18pt}
\end{table}

\begin{figure*}[h!]
	\centering
	\includegraphics[width=\linewidth]{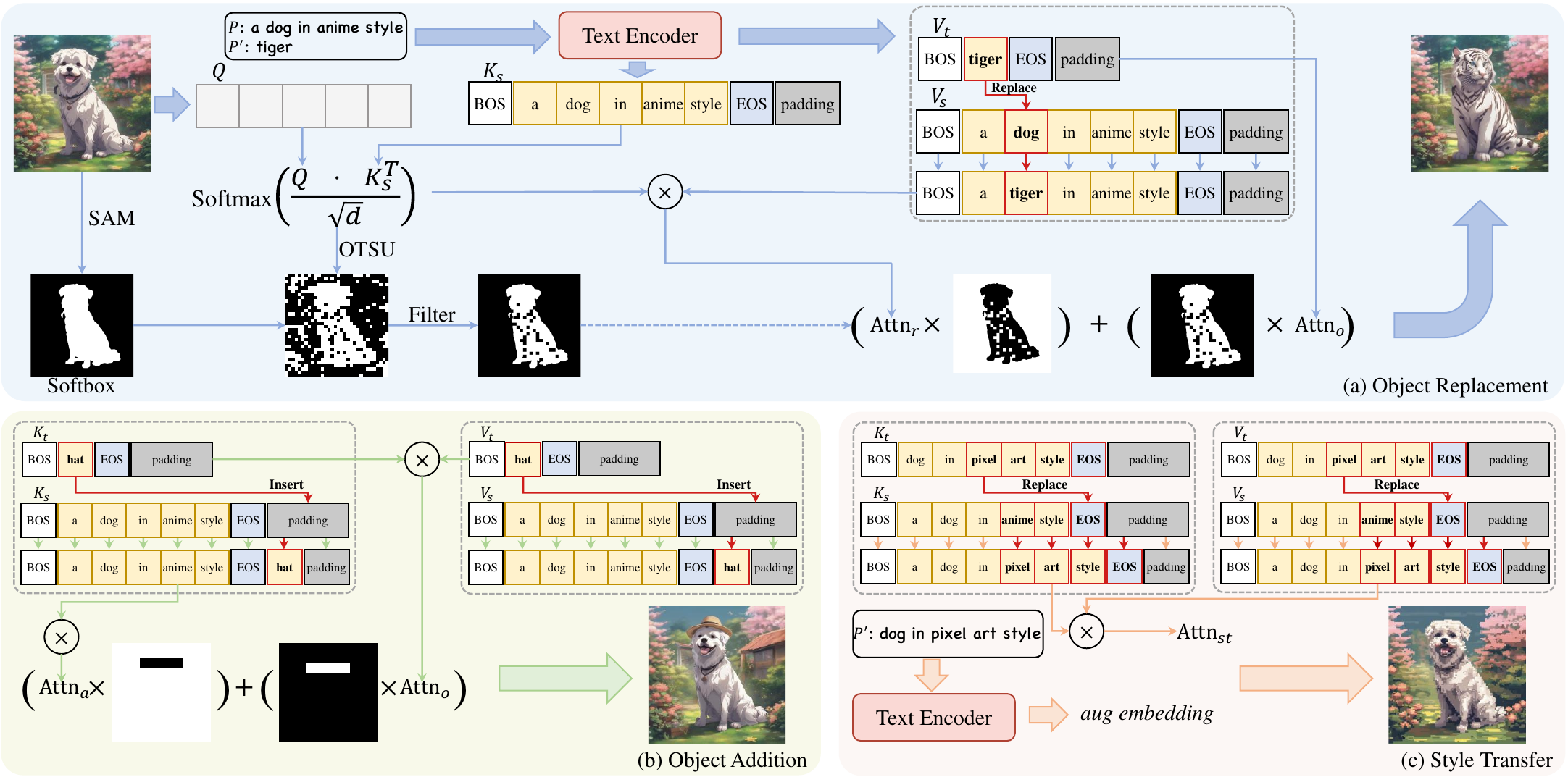}
	\caption{Overview of our proposed method across different tasks. The source prompt $P$ and target prompt $P'$ are transformed into $K_{s},K_{t}$ and $V_{s},V_{t}$ using different linear layers. SAM~\cite{kirillov2023segment} is the segmentation model.}
	\label{fig:model}
\end{figure*}

Inspired by~\cite{25}, we conduct detailed experiments, but our findings differ significantly from theirs. To analyze the roles of word semantics and style semantics within text embeddings, we conduct an ablation study by masking specific words or phrases in the prompts. As shown in Tab.~\ref{tab:motivation}, we evaluate multiple prompts and random seeds, and assess the probability of successfully generating the object or style using a Multimodal Large Model, even when certain tokens are masked. “Image Quality” is rated on a scale from 0 to 10. In addition, we use prompts of varying lengths to further evaluate the contribution of the \textit{aug embedding} (see Appendix A.1 for more detailed results). Visualization results are presented in Fig.~\ref{fig:experment}. Through extensive experimentation, we obtain several key insights. To interpret these insights, we begin by defining each component of the text embedding: the embeddings of all words represent word embeddings, padding is represented by \textit{padding embedding}, and \textit{aug embedding} is obtained by combining the pooled output of the final text encoder with the timestep embeddings. \textit{BOS/EOS} denotes the beginning/end of a sequence.

\vspace{4pt}
\noindent \textbf{Insight 1:} \textit{aug embedding} contains the complete textual semantic information but contributes relatively little to image generation. As the density of textual information increases, local semantics (e.g., certain objects and their textures) are significantly lost, whereas global features (e.g., style) are less affected.

\vspace{2pt}
\noindent \textbf{Discussion:} The text semantics in \textit{aug embedding} originate from the pooled output of the text encoder, thus encompassing complete semantic information. However, the pooling process compresses and entangles semantic features, potentially causing local semantic loss. As shown in the Tab.~\ref{tab:motivation}, when text information is low (fewer tokens), \textit{aug embedding} alone can generate coherent images; whereas with high text information (more tokens), local semantic loss becomes inevitable. Style information, being a global feature, remains largely preserved after compression (Fig.~\ref{fig:experment}~(a)). In the diffusion model, \textit{aug embedding} is fused with the latent through ResNet blocks, but contributes minimally to the final output since text information is primarily injected via cross-attention.

\vspace{4pt}
\noindent \textbf{Insight 2:} \textit{BOS} and \textit{padding embedding} don't contain any textual semantic information.

\vspace{2pt}
\noindent \noindent \textbf{Discussion:} As shown in Fig.~\ref{fig:experment}~(b) and Tab.~\ref{tab:motivation}, the presence or absence of \textit{BOS} doesn't affect the final image generation. Similarly, \textit{padding embedding} doesn't impact the generation of objects within the image but can slightly degrade the overall image quality (Fig.~\ref{fig:experment}~(c) blue box). This degradation is unrelated to text embeddings and is nearly imperceptible to the human eye.

\vspace{4pt}
\noindent \textbf{Insight 3:}  \textit{EOS} contains the semantic information of all words and stylistic information. Each word embedding is important and does not interfere with the semantic injection of other embeddings.

\vspace{2pt}
\noindent \noindent \textbf{Discussion:} Tab.~\ref{tab:motivation} demonstrates that \textit{EOS} is crucial for the quality and fidelity of the generated image, as it encompasses all semantic information and stylistic features. In Fig.~\ref{fig:experment}~(c), using only \textit{EOS} is sufficient to generate an image that aligns with the expected semantics (row 1 column 6). \textit{EOS} ensures that masking the phrase ``yellow suit'' does not affect the generation of this content (row 1 column 2). However, removing \textit{EOS} leads to a decline in image style and quality, while continuously masking stylistic words causes even more severe degradation (row 2 column 3). Additionally, each word embedding is important and does not interfere with others. When \textit{EOS} is removed, masking relevant words hinders the generation of the corresponding semantics (row 2 column 2).

\subsection{Controllable Edit}
\textbf{Preliminaries:} Diffusion models inject the semantic information of the text into deep features to generate the corresponding image through cross-attention layers. In the cross-attention layer, the input latent and text embeddings are transformed through different linear layers to obtain query, key, and value ($Q$, $K$, and $V$), respectively. The attention is computed using the following formula:
\begin{align}
	\text{Attn}_{o} = \text{Softmax}\left(\frac{QK^T}{\sqrt{d}}\right) \cdot V,
\end{align}
where $d$ is the dimension of $Q$ and $K$. CLIP limits the maximum length of text tokens to 77. The softmax function calculates the relevance between $Q$ and $K$, which is then multiplied by $V$ to inject specific textual semantics into the latent.

\noindent \textbf{Task Description:} For a source prompt $P$ and a target prompt $P'$, previous methods \cite{24,25} imposed strict formatting requirements on these prompts. Aside from the words to be modified, all other words had to remain unchanged. Essentially, this approach involved inserting the attention map from the source prompt into the target prompt to achieve the desired modification. Unlike existing approaches, our method relaxes the format constraints and enables direct editing using free-form text (target prompt). As illustrated in Fig.~\ref{fig:model}, PSP modifies the image content by manipulating the key and value in the cross-attention layers. We will describe our method through several sub-tasks, demonstrating its effectiveness and flexibility in various application scenarios.

\noindent \textbf{Object Replacement.} In the cross-attention layer, the text embeddings from source prompt $P$ and target prompt $P'$ are transformed into key (\(K_s\) and \(K_t\)) and value (\(V_s\) and \(V_t\)) via separate linear layers. The image features are updated using the following formula:
\begin{align}
	\text{Attn}_{r}\left(Q, K_s, V_{s,t} \right) = \text{Softmax}\left(\frac{Q K_s^T}{\sqrt{d}}\right) \cdot \text{Re}(V_s, V_t),
\end{align}
where \(\text{Re}(\cdot)\) is a replacement function that substitutes the object in $P$ with the object in $P'$. For example, \(P: \text{``a photo of a dog''}\) and \(P': \text{``tiger''}\), the function replaces ``dog'' in \(P\) with ``tiger'' from \(P'\) at the text embedding level. This direct replacement method performs well for prompts containing a single object. However, for prompts with multiple objects or complex scenes, it often yields suboptimal results. As highlighted in Insight 1 and Insight 3, the information of the object is not only present in the word embeddings but also included in the \textit{aug embedding} and \textit{EOS}. This implies that merely replacing the object information is insufficient. Moreover, the entangled nature of the \textit{aug embedding} and \textit{EOS} makes it challenging to decompose and modify them independently.

To address this challenge, we design Softbox, which directly injects the semantics of $P'$ into the region of the source object, thereby minimizing interference from the source object. Softbox is essentially a mask that can be either extracted by a segmentation model~\cite{kirillov2023segment} or manually selected by users. As illustrated in Fig.~\ref{fig:model}~(a), the attention map corresponding to the source object is first determined using the following formula:
\begin{align}
	\mathcal{A}^i\left(Q, K_s^i\right) = \text{Softmax}\left(\frac{Q{K_s^i}^T}{\sqrt{d}}\right),
\end{align}
where \(K_s^i\) denotes the embedding corresponding to the \(i\)-th token (i.e., the source object) in \(K_s\). Subsequently, using the Otsu method~\cite{otsu1975threshold} and the Softbox, we obtain the source object's mask:
\begin{align}
	\mathcal{M}^i = \mathcal{B}^i * \text{OTSU}(\mathcal{A}^i),
\end{align}
where $\mathcal{B}^i$ represents the Softbox for the $i$-th token, which provides a constraint. The final formula for attention is expressed as follows:
\begin{align}
	\text{Attn} = \text{Attn}_{r} \cdot (1 - \mathcal{M}^i) + \text{Attn}_{o}(Q, K_{t}, V_{t}) \cdot \mathcal{M}^i.
\end{align}
\begin{figure}[!t]
    \centering
    \begin{subfigure}[t]{\maskpicsize}
        \includegraphics[width=\textwidth]{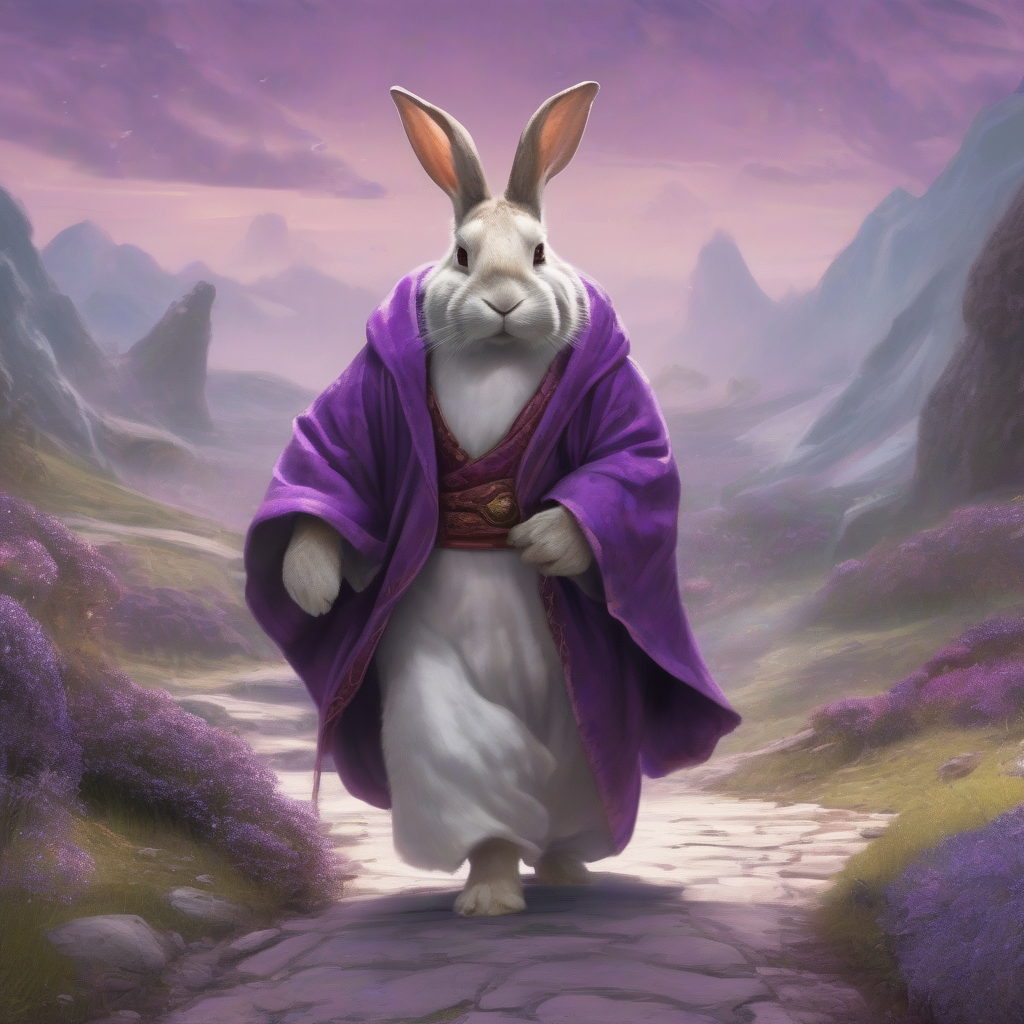}
        \vspace{-15pt}
        \caption*{origin}
    \end{subfigure}
    \begin{subfigure}[t]{\maskpicsize}
        \includegraphics[width=\textwidth]{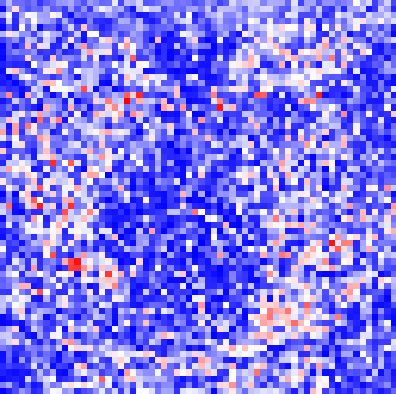}
        \vspace{-15pt}
        \caption*{Block=1}
    \end{subfigure}
    \begin{subfigure}[t]{\maskpicsize}
        \includegraphics[width=\textwidth]{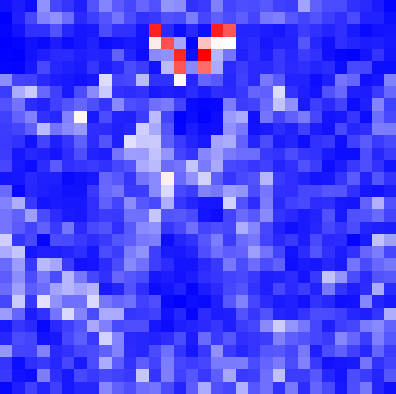}
        \vspace{-15pt}
        \caption*{Block=19}
    \end{subfigure}
    \begin{subfigure}[t]{\maskpicsize}
        \includegraphics[width=\textwidth]{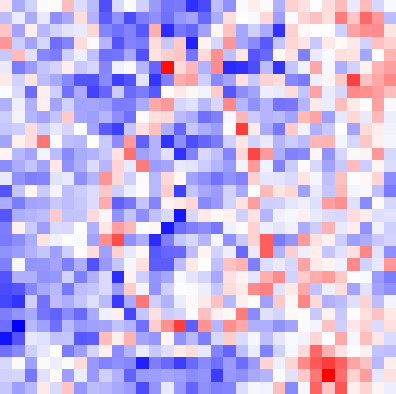}
        \vspace{-15pt}
        \caption*{Block=29}
    \end{subfigure}
    \begin{subfigure}[t]{\maskpicsize}
        \includegraphics[width=\textwidth]{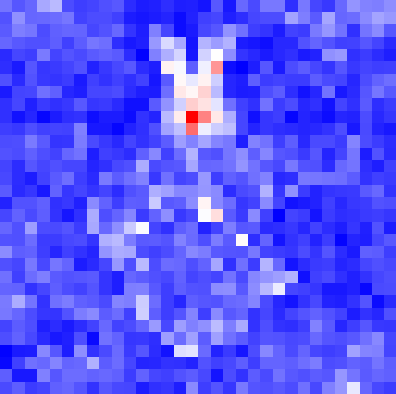}
        \vspace{-15pt}
        \caption*{Block=45}
    \end{subfigure}
    \begin{subfigure}[t]{\maskpicsize}
        \includegraphics[width=\textwidth]{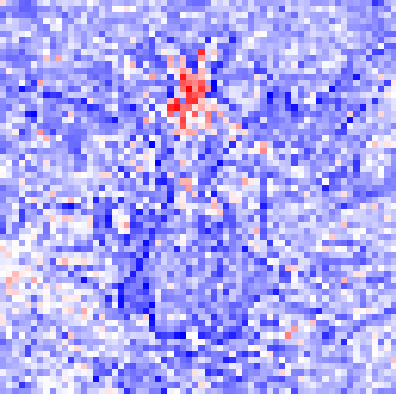}
        \vspace{-15pt}
        \caption*{Block=65}
    \end{subfigure}
    \begin{subfigure}[t]{\maskpicsize}
        \includegraphics[width=\textwidth]{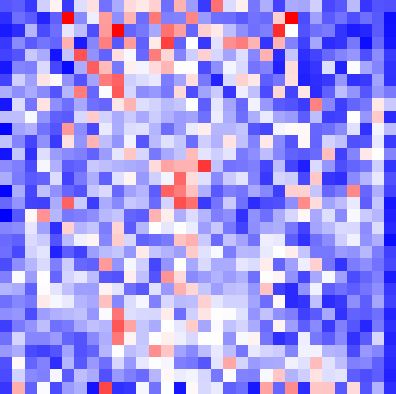}
        \vspace{-15pt}
        \caption*{T=0}
    \end{subfigure}
    \begin{subfigure}[t]{\maskpicsize}
        \includegraphics[width=\textwidth]{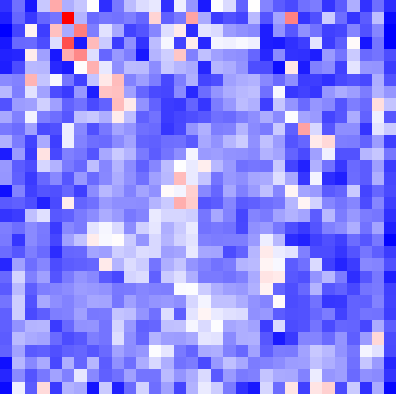}
        \vspace{-15pt}
        \caption*{T=3}
    \end{subfigure}
    \begin{subfigure}[t]{\maskpicsize}
        \includegraphics[width=\textwidth]{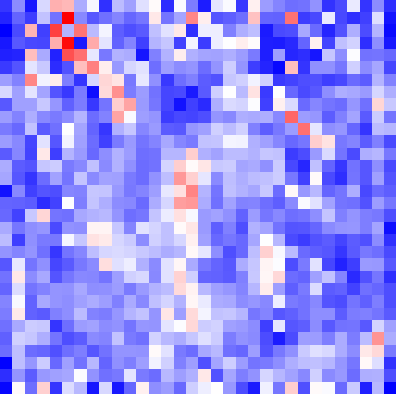}
        \vspace{-15pt}
        \caption*{T=5}
    \end{subfigure}
    \begin{subfigure}[t]{\maskpicsize}
        \includegraphics[width=\textwidth]{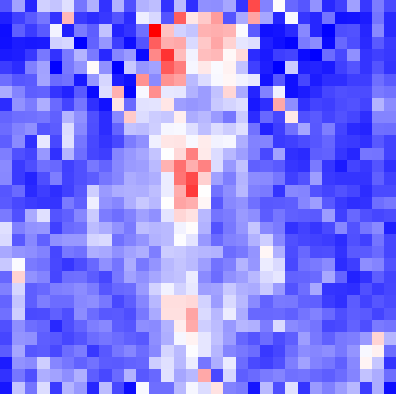}
        \vspace{-15pt}
        \caption*{T=10}
    \end{subfigure}
    \begin{subfigure}[t]{\maskpicsize}
        \includegraphics[width=\textwidth]{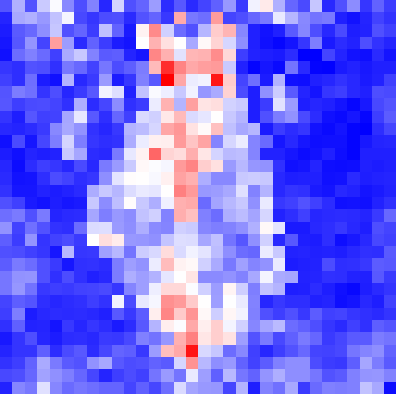}
        \vspace{-15pt}
        \caption*{T=20}
    \end{subfigure}
    \begin{subfigure}[t]{\maskpicsize}
        \includegraphics[width=\textwidth]{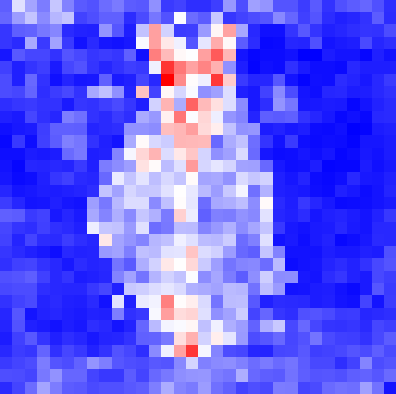}
        \vspace{-15pt}
        \caption*{T=25}
    \end{subfigure}
    \vspace{-10pt}
    \caption{Attention maps in the cross-attention layer. The first row displays the attention maps from different cross-attention blocks at time step 29, while the second row shows the attention maps across different time steps within the same cross-attention block.}
    \label{fig:mask}
    \vspace{-10pt}
\end{figure}
As shown in Fig.\ref{fig:mask}, early in the image generation process, before the object has fully emerged, the correlation between $Q$ and $K$ computed using the above formula may introduce irrelevant regions, which affect the accuracy of the results (See Appendix A.2 for Otsu-processed results). To address this issue, Softbox plays a crucial role in filtering during the early stages by constraining $\mathcal{A}^i$ to exclude irrelevant areas, thereby preventing modifications to other parts of the final image. Our approach also allows for the manipulation of an object's actions and colors through the aforementioned steps. These operations require the target prompt to contain the relevant semantic information. Examples demonstrating this capability are shown in Fig.~\ref{fig:teaser} and Fig.~\ref{fig:replace}.

\noindent \textbf{Object Addition.} Adding objects differs significantly from object replacement, as the added object does not replace an existing one in the source prompt, which makes defining the mask region more challenging. Based on Insight 2, a straightforward approach can be derived. As illustrated in Fig~\ref{fig:model}~(b), the semantic information of the target prompt is directly inserted into the padding area of the source prompt. However, this approach encounters a problem: the semantic information of the target prompt is not included in $Q$, resulting in an inability to form a weight matrix related to the added object during the interaction between $Q$ and $K$. Consequently, even when inserting the target prompt, it is often difficult to effectively introduce the target semantics. A natural solution is to embed the semantic information of the target prompt into a predefined Softbox region. This method not only facilitates the interaction between $Q$ and $K$ to form a weight matrix for the added object but also ensures seamless integration of the added target with the background. The final formula is given by:
\begin{align}
	\text{Attn}_{a}\left(Q, K_{s,t}, V_{s,t}\right) = \text{Softmax}\left(\frac{Q \cdot \text{Add}(K_{s}, K_{t})^{T}}{\sqrt{d_{}}}\right) \cdot \text{Add}(V_{s}, V_{t}) ,
\end{align}
\begin{align}
	\text{Attn}=\text{Attn}_{a} \cdot (1 - \mathcal{B}^i)+\text{Attn}_{o}(Q, K_{t}, V_{t}) \cdot \mathcal{B}^i,
\end{align}
where \( \text{Add}(\cdot) \) denotes the insert function, and $\mathcal{B}^i$ represents the Softbox region for the \( i \)-th added object.

\noindent \textbf{Style Transfer.} Considering that the \textit{EOS} and \textit{aug embedding} contain style information, it is both foreseeable and reasonable to directly operate on these components, as suggested by Insight 1 and Insight 3. Unlike object replacement, style transfer requires the joint operation of $Q$, $K$, and $V$. If the source prompt and target prompt are completely unrelated, the corresponding weight matrix cannot be retrieved from $Q$ using $K_t$, making it impossible to effectively inject the semantic style from $V_t$ into the source prompt. This suggests that modifying the source prompt using the $K$ and $V$ of the target prompt is ineffective because they fail to establish a consistent mapping. Therefore, a fundamental criterion for generating the target prompt is that it must combine a specific object from the source prompt while allowing for any stylistic variation. For example: $P$: ``happy koala wearing a yellow turtleneck, facing camera, studio, photorealistic style''. $P^{'}$: ``koala in minimal origami style''. In such a scenario, the target prompt enables retrieval of the same object information from Q, facilitating effective style transfer. The formula for this process is as follows:
\begin{align}
	\text{Attn}_{st}\left(Q, K_{s,t}, V_{s,t}\right)= 	\text{Softmax}\left(\frac{Q \cdot \text{Re}(K_{s},K_{t})^{T}}{\sqrt{d}}\right) \cdot \text{Re}(V_{s},V_{t}).
\end{align}

As illustrated in Fig.~\ref{fig:model}~(c), for style transfer, the Re($\cdot$) function is employed to replace the style phrases and \textit{EOS} from the source prompt with those from the target prompt. Subsequently, the \textit{aug embedding} of the source prompt is substituted with the \textit{aug embedding} of the target prompt to achieve complete style transfer. This approach facilitates more precise style transfer, even in complex scenes with multiple objects, ensuring both the effectiveness and accuracy of the process.

Our method represents a general and training-free image editing control technique. In this algorithm, we perform the diffusion process only once at each time step and apply the PSP operation at predefined steps to accommodate specific editing tasks. Since our approach uses two prompts applied to a single noise during processing, there are no special constraints needed for the random seed. Formally, our method is summarized in Algorithm~\ref{algorithmic1}.
\begin{algorithm}
	\caption{PSP image editing}
	\label{alg:PSP}
	\begin{algorithmic}[1]
		\REQUIRE A source prompt $P$, a target prompt $P^{'}$, $\lambda_1$ and $\lambda_2$ are the timestamp parameters.
		\ENSURE An edited image $x$
		\STATE $z_t \sim N(0,1)$
		\FOR{$t = T, T-1, \dots ,1$}
		\STATE { $z_{t-1}=\text{DM}(z_{t},P,P',\mathrm{t}) :=\begin{cases}\text{PSP}(P,P') &\quad \mathrm{if} \quad\lambda_1<\mathrm{t}<\lambda_2,\\ \text{Attn}_{o}(P) &\quad \mathrm{otherwise}.\end{cases}$} \\
		\ENDFOR
		\RETURN $z_0$
	\end{algorithmic}
	\label{algorithmic1}
\end{algorithm}
 
\section{Experiments}

\subsection{Experimental Setup}
\noindent \textbf{Baselines and datasets}. For a fair comparison, we evaluate our method against SDXL-based methods (P2P~\cite{24}, MasaCtrl~\cite{cao2023masactrl}, and InstructPix2Pix~\cite{brooks2023instructpix2pix}) and the DiT-based Flux-Fill~\cite{flux}. We build our validation set by sampling prompts from PIE-Bench~\cite{ju2023direct} and generating masks using SAM~\cite{kirillov2023segment} to evaluate these methods.

\noindent \textbf{Metrics}. To evaluate consistency between generated and original images in unmasked regions, we employ PSNR, LPIPS~\cite{zhang2018unreasonable}, and MSE. For assessing text-image alignment in masked regions, we utilize CLIP similarity (CLIP Sim)~\cite{radford2021learning}.

\noindent \textbf{Implementation Details}. We set the number of inference steps to 30 and generated images at a resolution of 1024×1025, keeping all other parameters at their default SDXL settings. All experiments are conducted on a single NVIDIA A40 GPU.

\subsection{Comparisons}
\noindent \textbf{Quantitative Comparison}. As shown in Tab.~\ref{tab:comparision}, the DiT-based Flux.Fill method achieves the best performance in background consistency preservation, benefiting from its large-scale training data and advanced network architecture. However, this method demonstrates poor semantic alignment between the generated content in masked regions and the corresponding text prompts. While the MasaCtrl method attains the highest CLIP similarity score, it comes at the cost of significant background detail degradation. In comparison, our method not only accurately generates edited content that maintains high semantic consistency with the text descriptions, but also effectively preserves the original background details, achieving superior overall performance.

\begin{figure}[t]
    \centering
    \begin{tikzpicture}
\node[anchor=base,scale=\comparisonscale] at (4.2cm, -0.0cm) { bear $\rightarrow$ {\color{red} dragon}};
        \node[anchor=north west] (img41) at (0,0) {
            \begin{subfigure}[t]{\comparisionsize}
                \includegraphics[width=\textwidth]{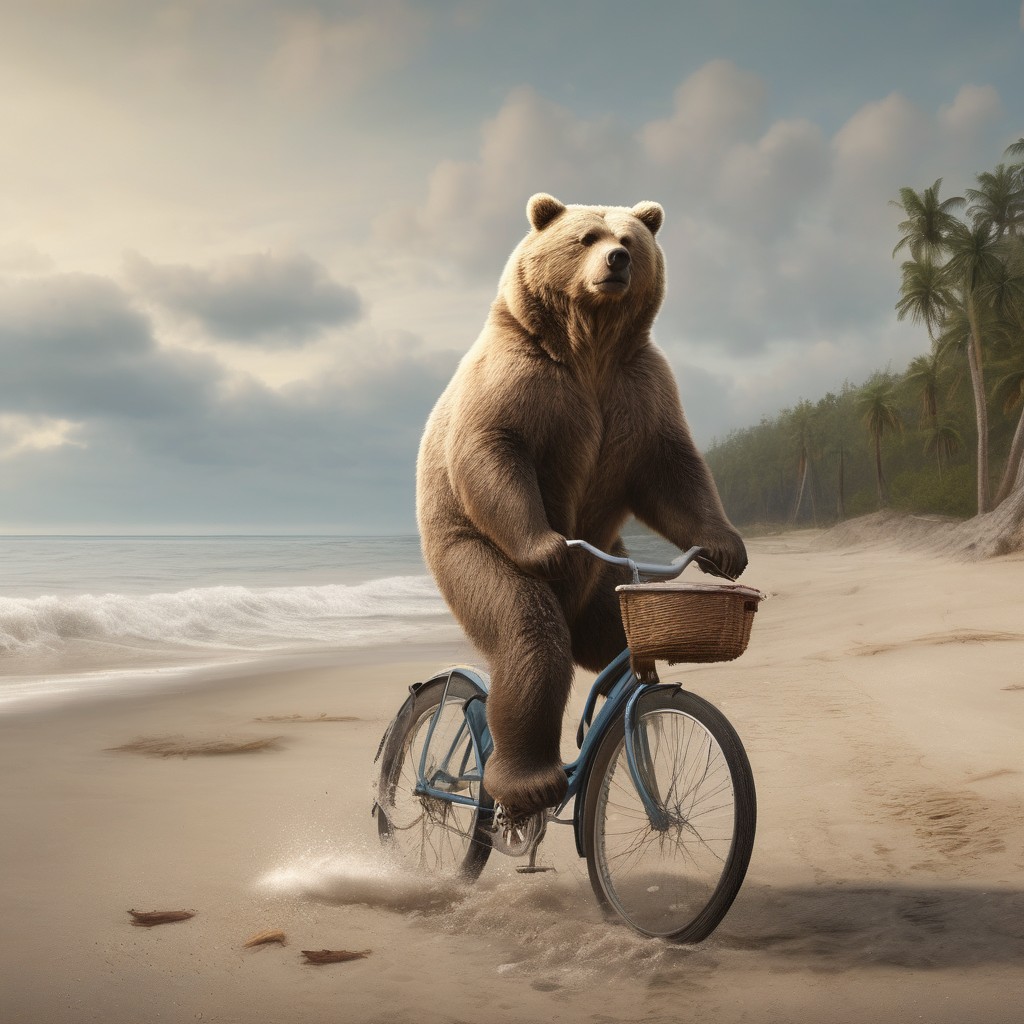}
            \end{subfigure}
        };
        \node[anchor=north west] (img42) at (1.35cm,0) {
            \begin{subfigure}[t]{\comparisionsize}
                \includegraphics[width=\textwidth]{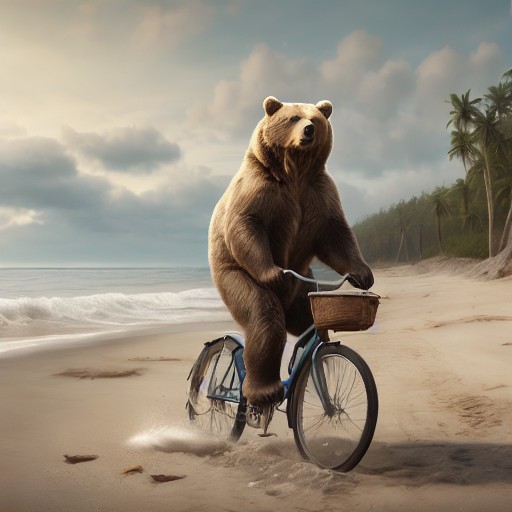}
            \end{subfigure}
        };
        \node[anchor=north west] (img43) at (2.7cm,0) {
            \begin{subfigure}[t]{\comparisionsize}
                \includegraphics[width=\textwidth]{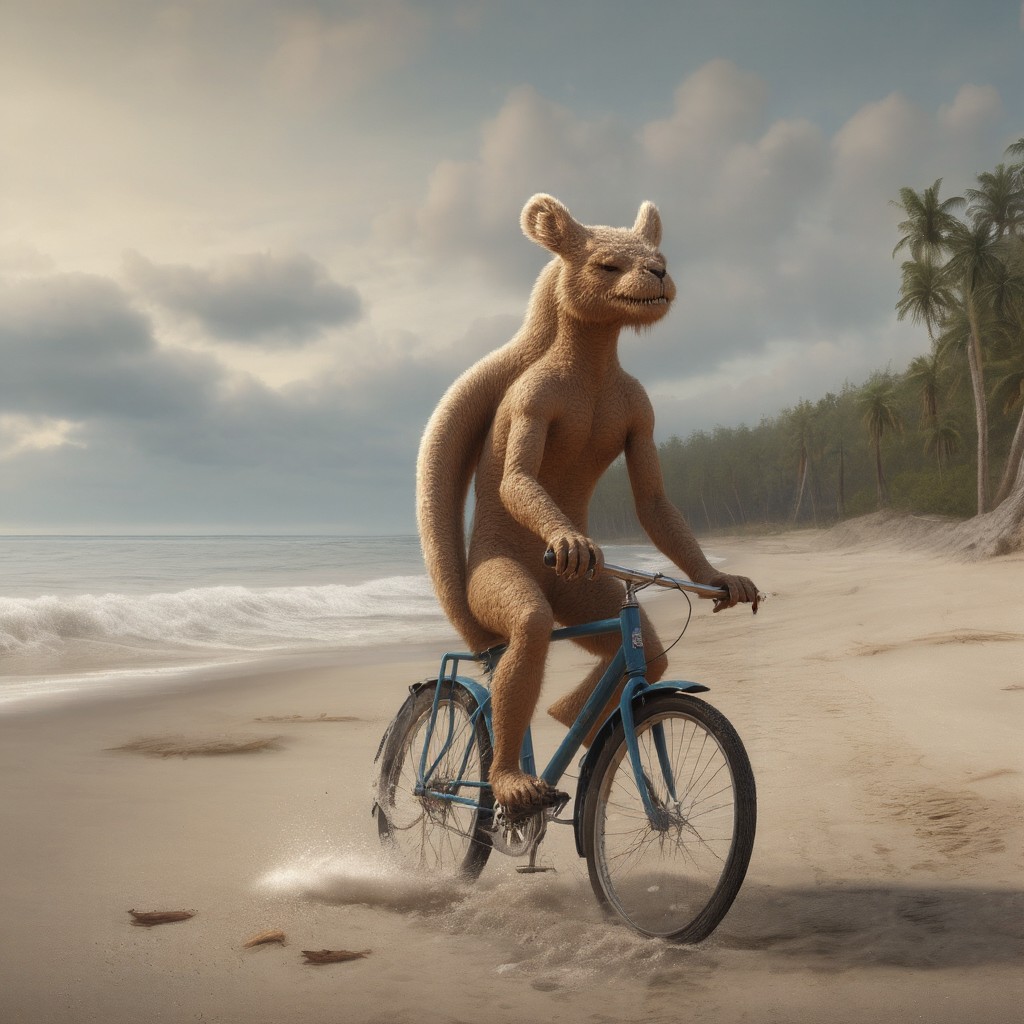}
            \end{subfigure}
        };
        \node[anchor=north west] (img44) at (4.05cm,0) {
            \begin{subfigure}[t]{\comparisionsize}
                \includegraphics[width=\textwidth]{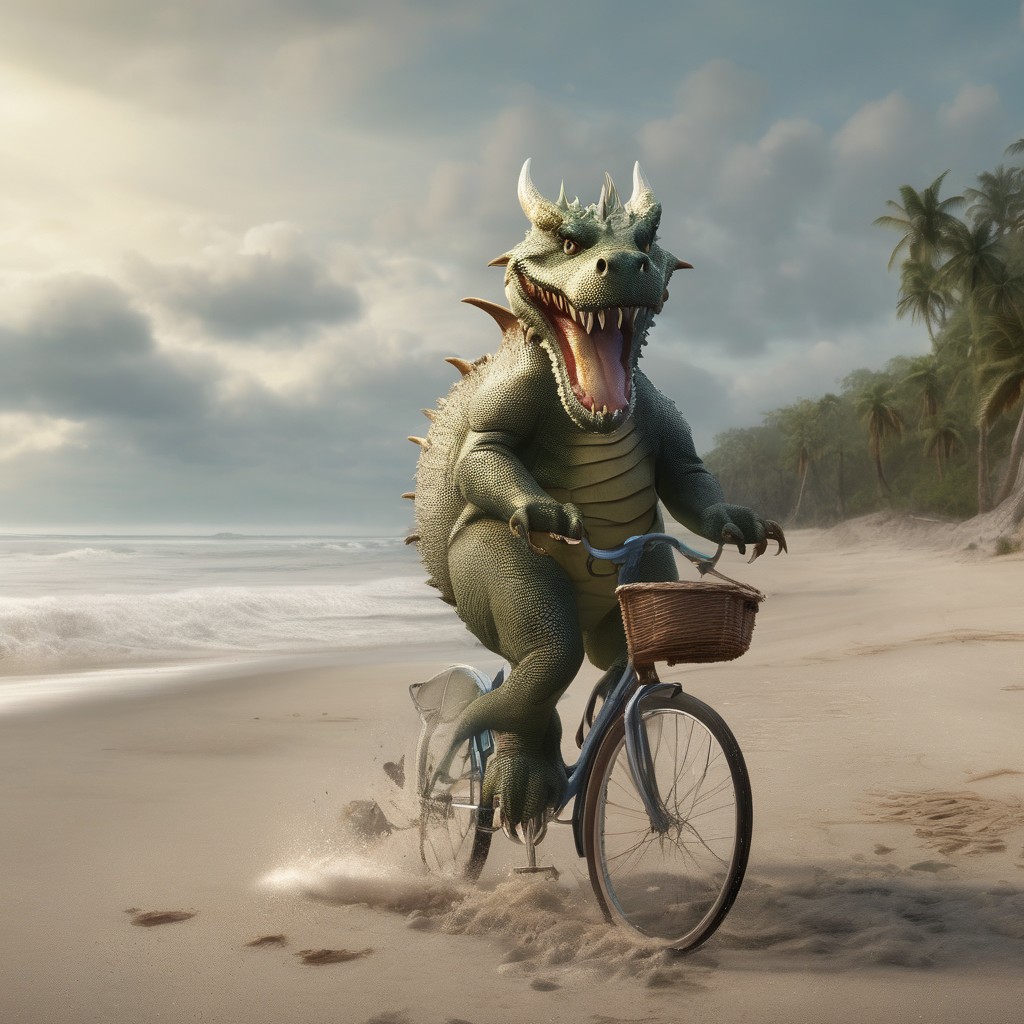}
            \end{subfigure}
        };
        \node[anchor=north west] (img44) at (5.4cm,0) {
            \begin{subfigure}[t]{\comparisionsize}
                \includegraphics[width=\textwidth]{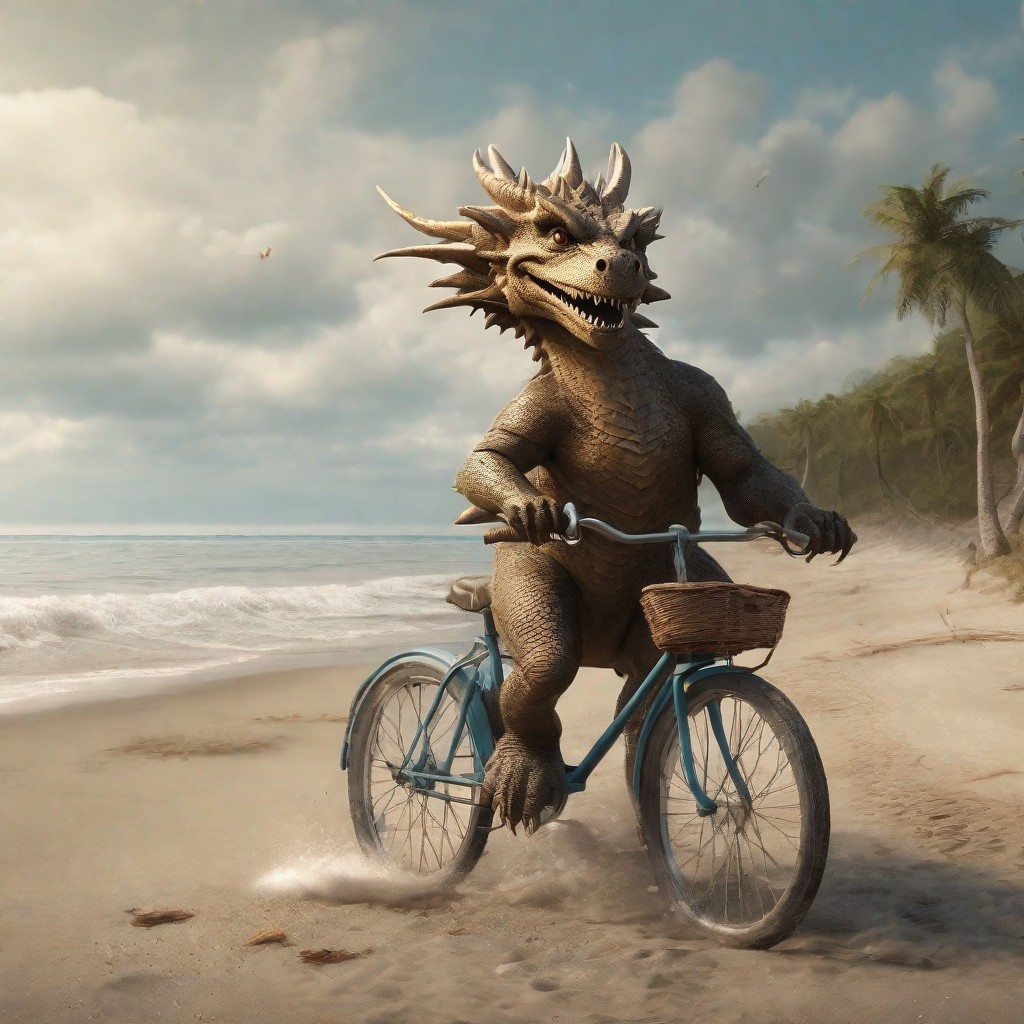}
            \end{subfigure}
        };
        \node[anchor=north west] (img45) at (6.75cm,0) {
            \begin{subfigure}[t]{\comparisionsize}
                \includegraphics[width=\textwidth]{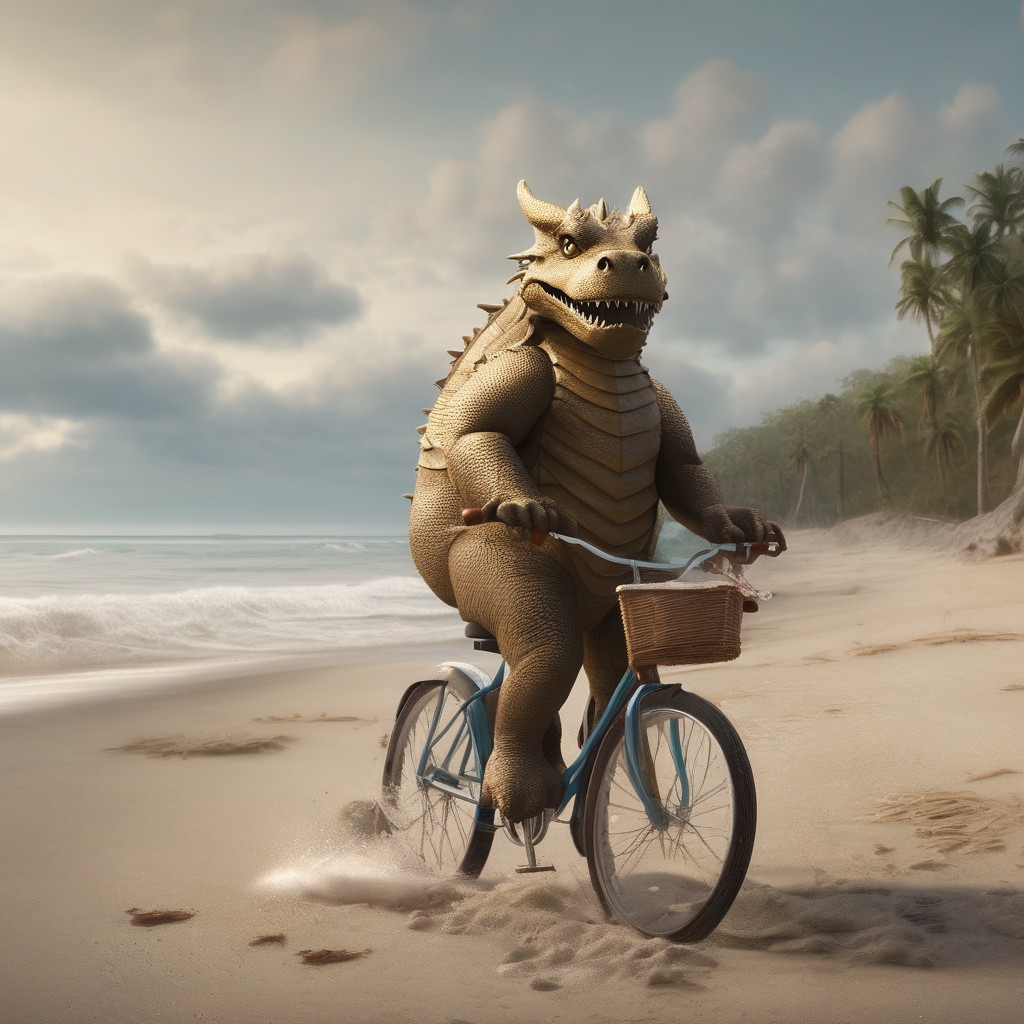}
            \end{subfigure}
        };

        \node[anchor=base,scale=\comparisonscale] at (4.2cm, -1.7cm) { Lionel Messi $\rightarrow$ {\color{red} Cristiano Ronaldo}};
        \node[anchor=north west] (img21) at (0,-1.65cm) {
            \begin{subfigure}[t]{\comparisionsize}
                \includegraphics[width=\textwidth]{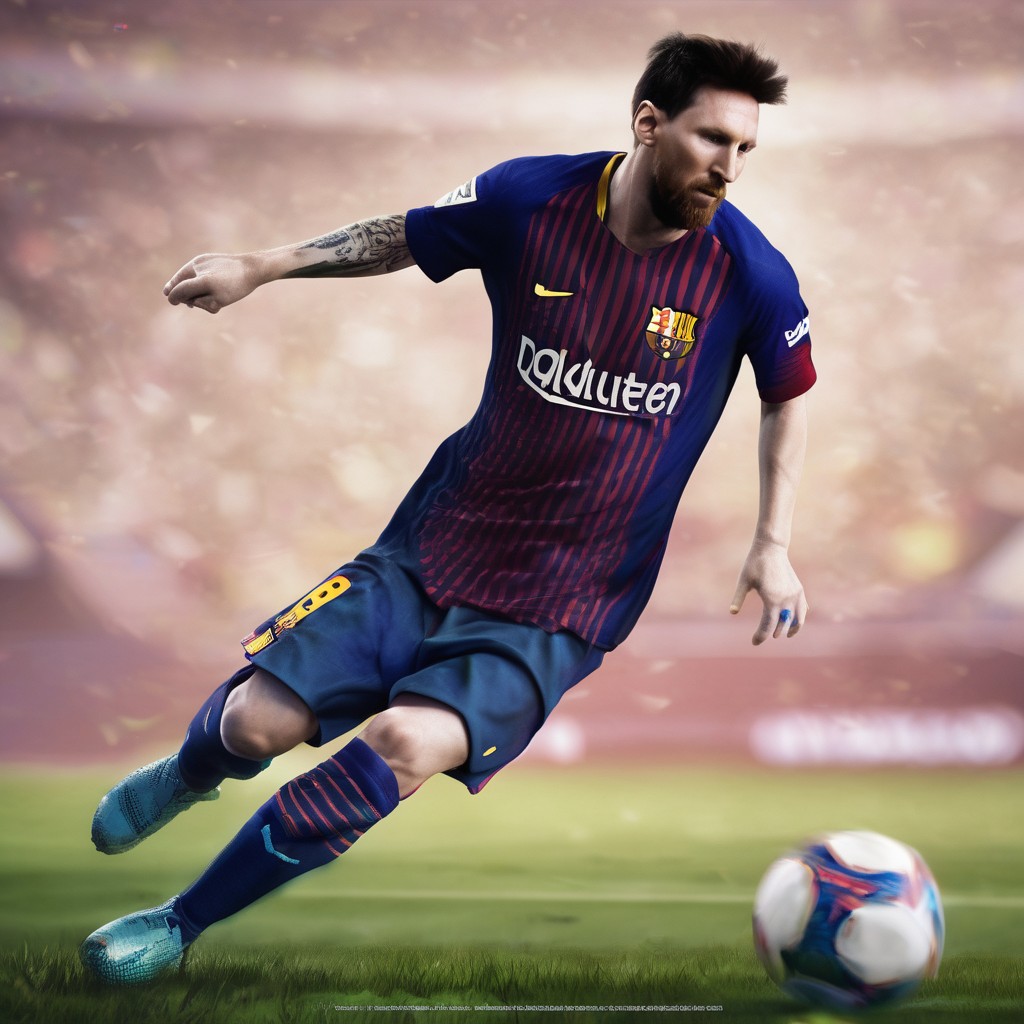}
            \end{subfigure}
        };

        \node[anchor=north west] (img22) at (1.35cm,-1.65cm) {
            \begin{subfigure}[t]{\comparisionsize}
                \includegraphics[width=\textwidth]{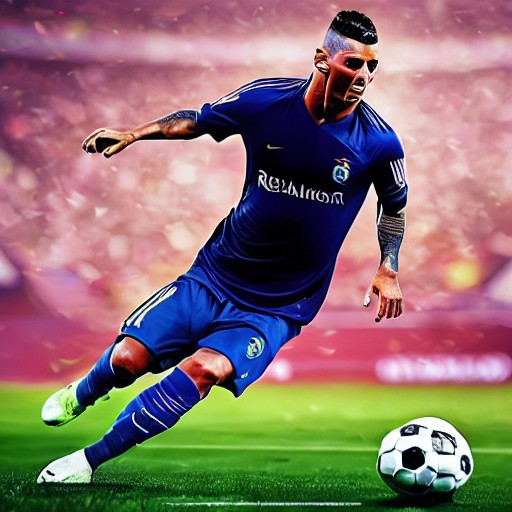}
            \end{subfigure}
        };
        \node[anchor=north west] (img23) at (2.7cm,-1.65cm) {
            \begin{subfigure}[t]{\comparisionsize}
                \includegraphics[width=\textwidth]{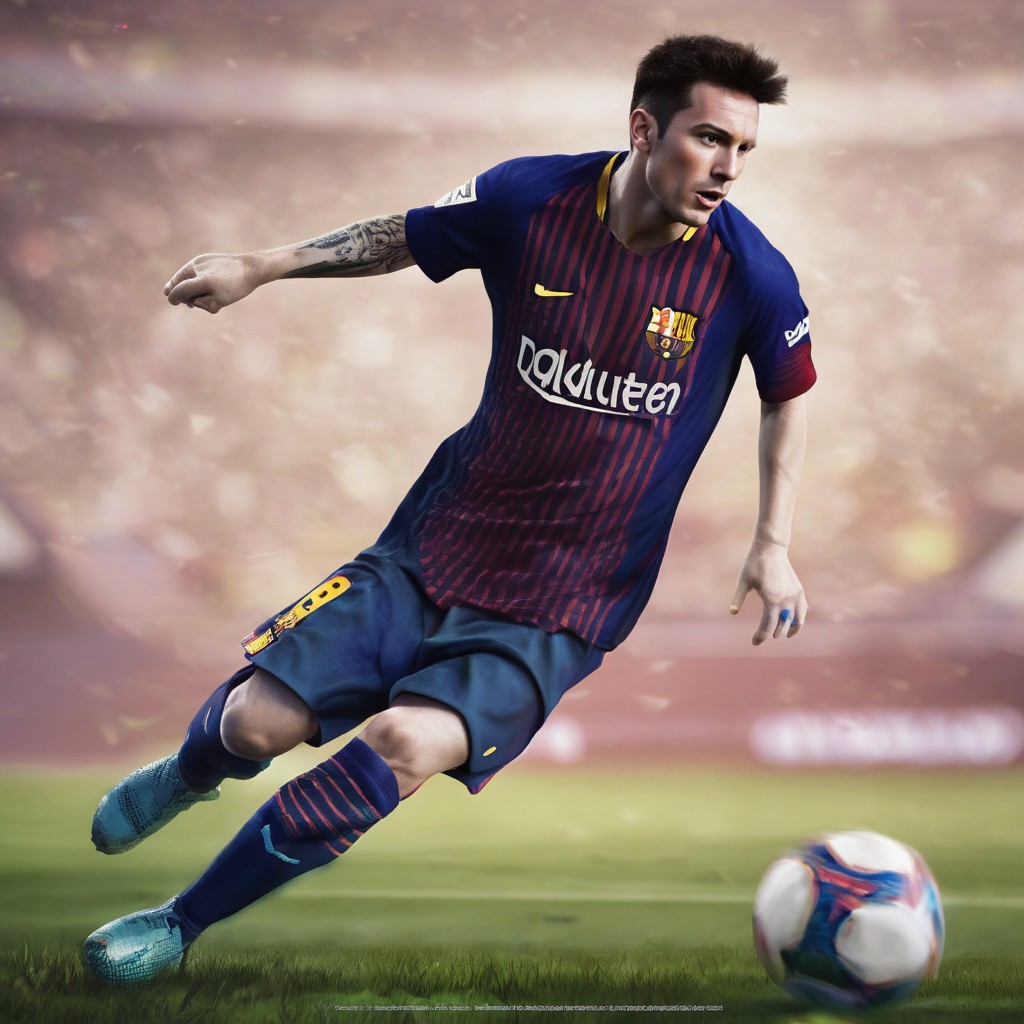}
            \end{subfigure}
        };
        \node[anchor=north west] (img24) at (4.05cm,-1.65cm) {
            \begin{subfigure}[t]{\comparisionsize}
                \includegraphics[width=\textwidth]{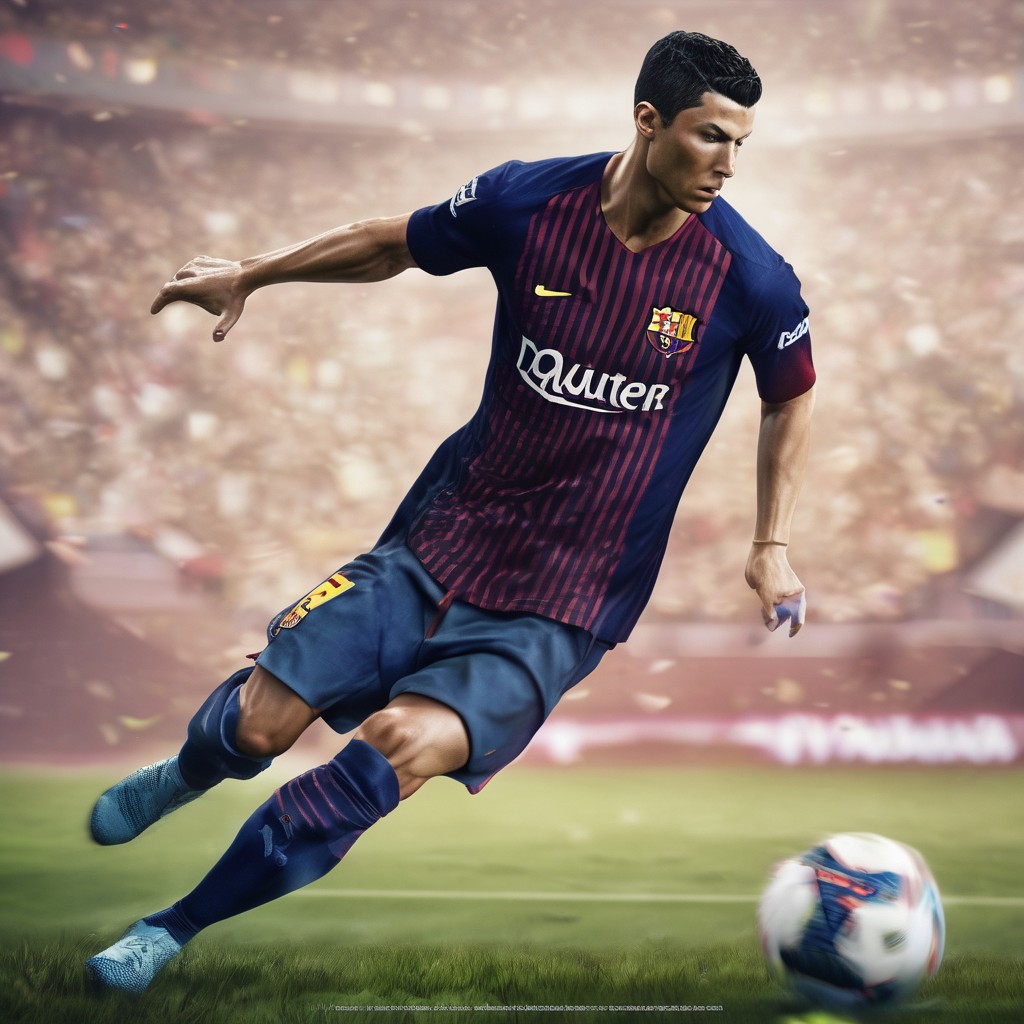}
            \end{subfigure}
        };
        \node[anchor=north west] (img25) at (5.4cm,-1.65cm) {
            \begin{subfigure}[t]{\comparisionsize}
                \includegraphics[width=\textwidth]{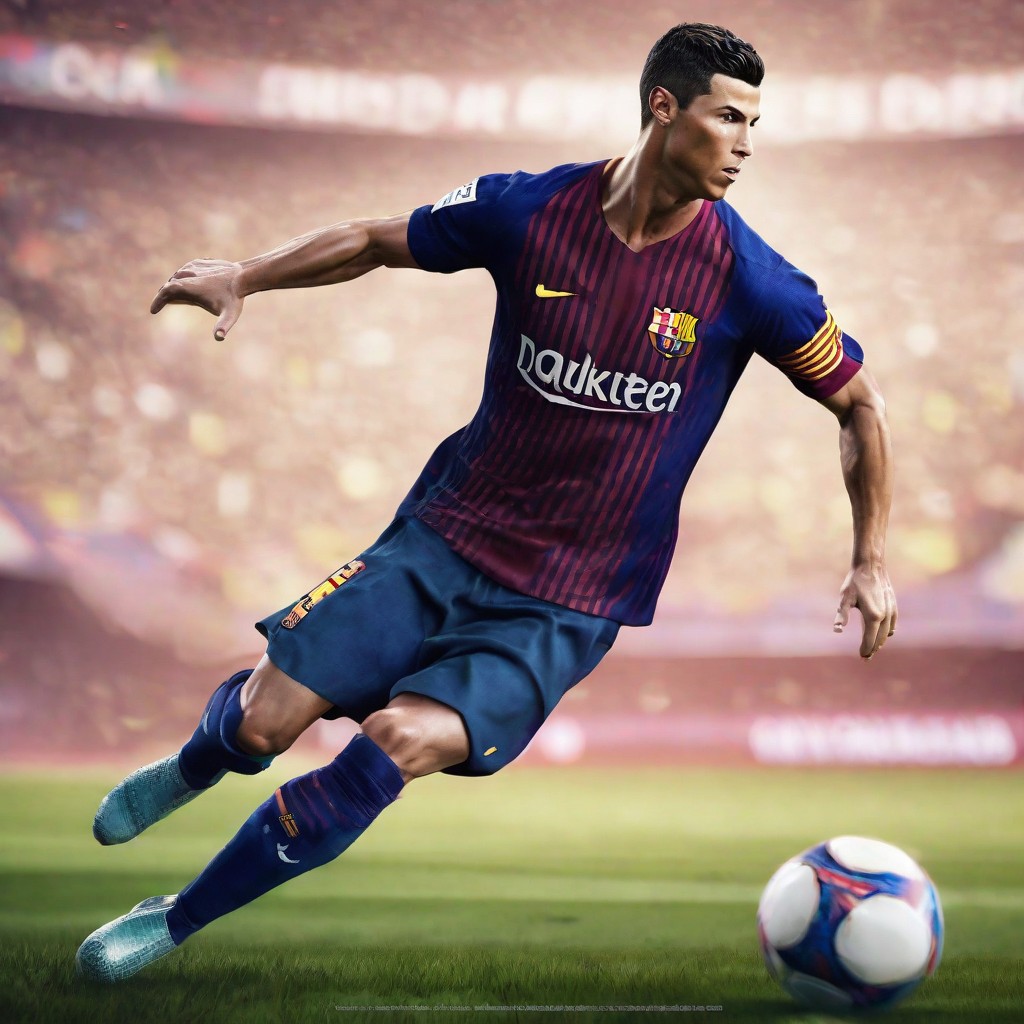}
            \end{subfigure}
        };
        \node[anchor=north west] (img45) at (6.75cm,-1.65cm) {
            \begin{subfigure}[t]{\comparisionsize}
                \includegraphics[width=\textwidth]{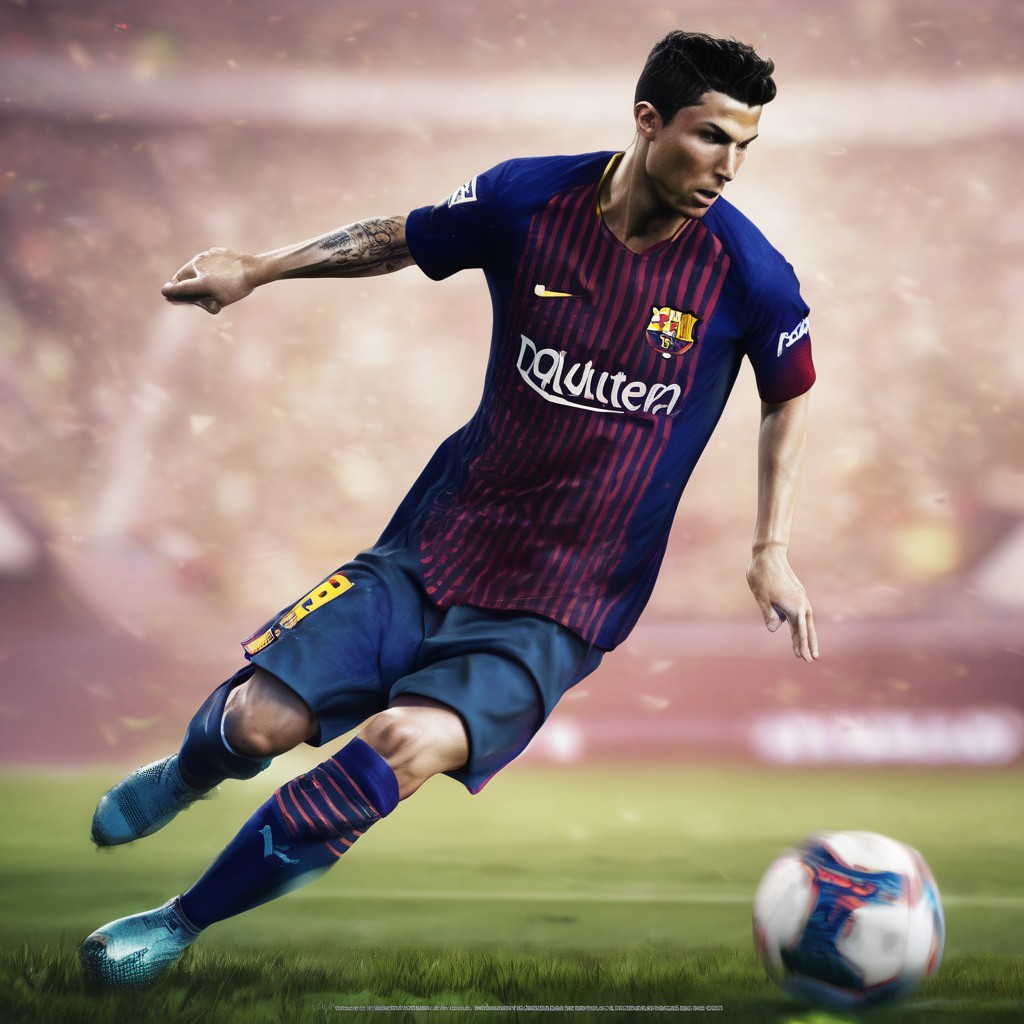}
            \end{subfigure}
        };

        \node[anchor=base,scale=\comparisonscale] at (4.2cm, -3.3cm) { happy woman $\rightarrow$ {\color{red} anger woman}};
        \node[anchor=north west] (img31) at (0,-3.3) {
            \begin{subfigure}[t]{\comparisionsize}
                \includegraphics[width=\textwidth]{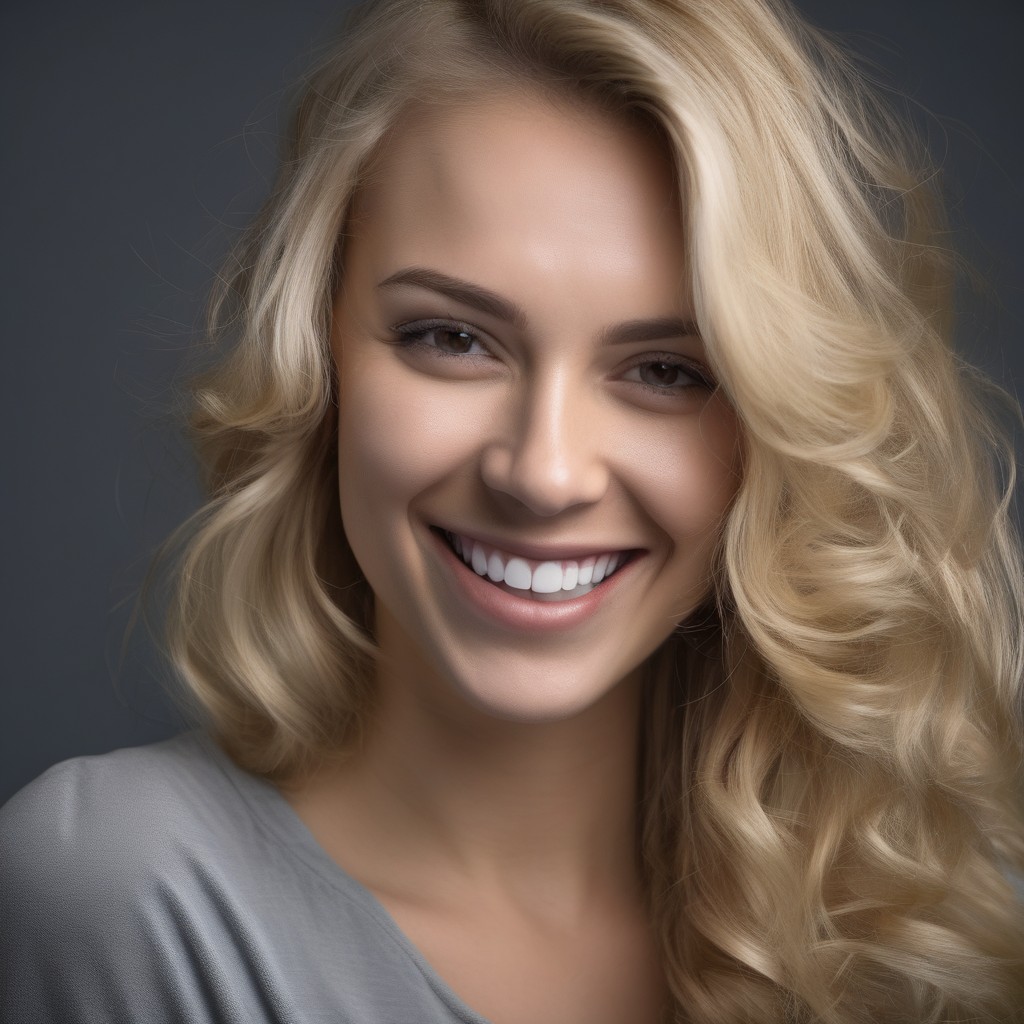}
            \end{subfigure}
        };
        \node[anchor=north west] (img32) at (1.35cm,-3.3cm) {
            \begin{subfigure}[t]{\comparisionsize}
                \includegraphics[width=\textwidth]{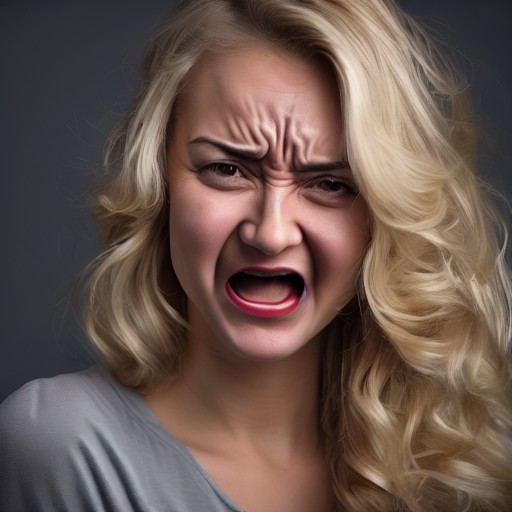}
            \end{subfigure}
        };
        \node[anchor=north west] (img33) at (2.7cm,-3.3cm) {
            \begin{subfigure}[t]{\comparisionsize}
                \includegraphics[width=\textwidth]{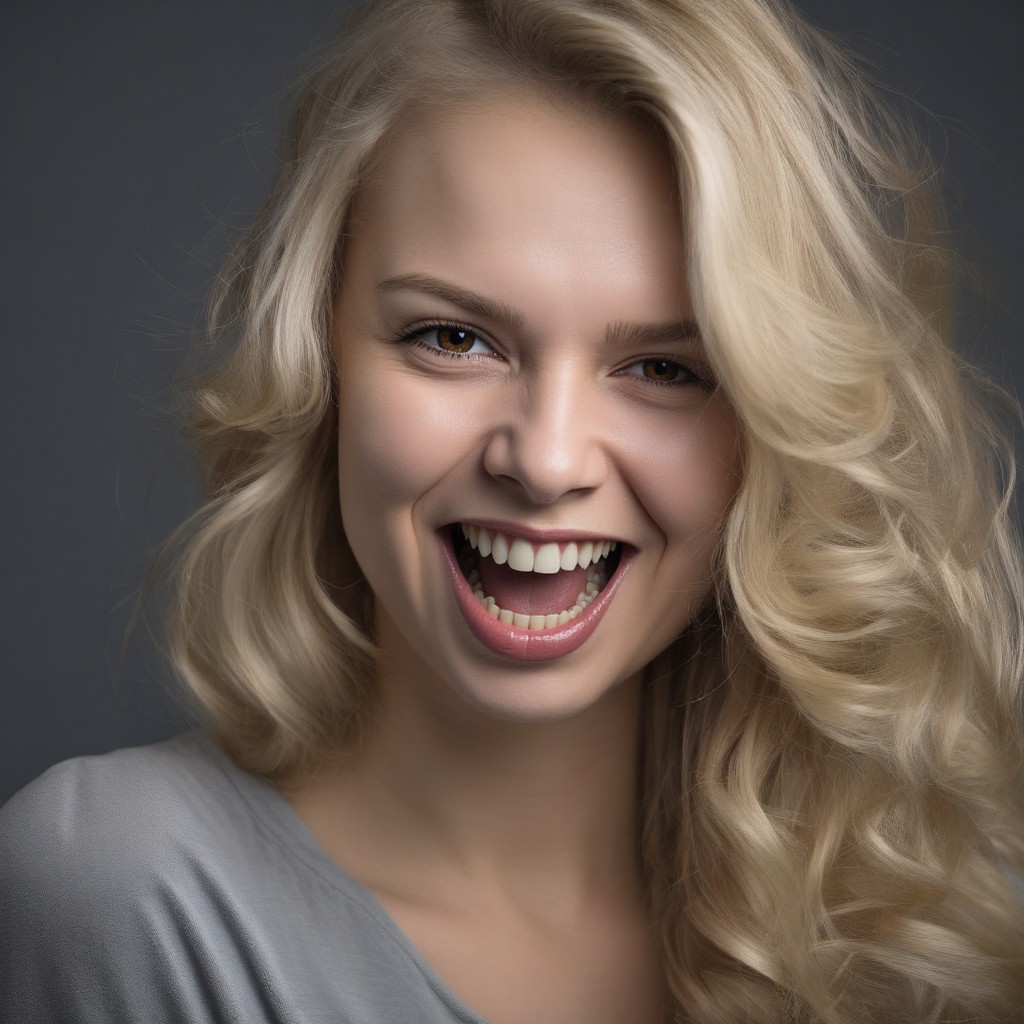}
            \end{subfigure}
        };
        \node[anchor=north west] (img34) at (4.05cm,-3.3cm) {
            \begin{subfigure}[t]{\comparisionsize}
                \includegraphics[width=\textwidth]{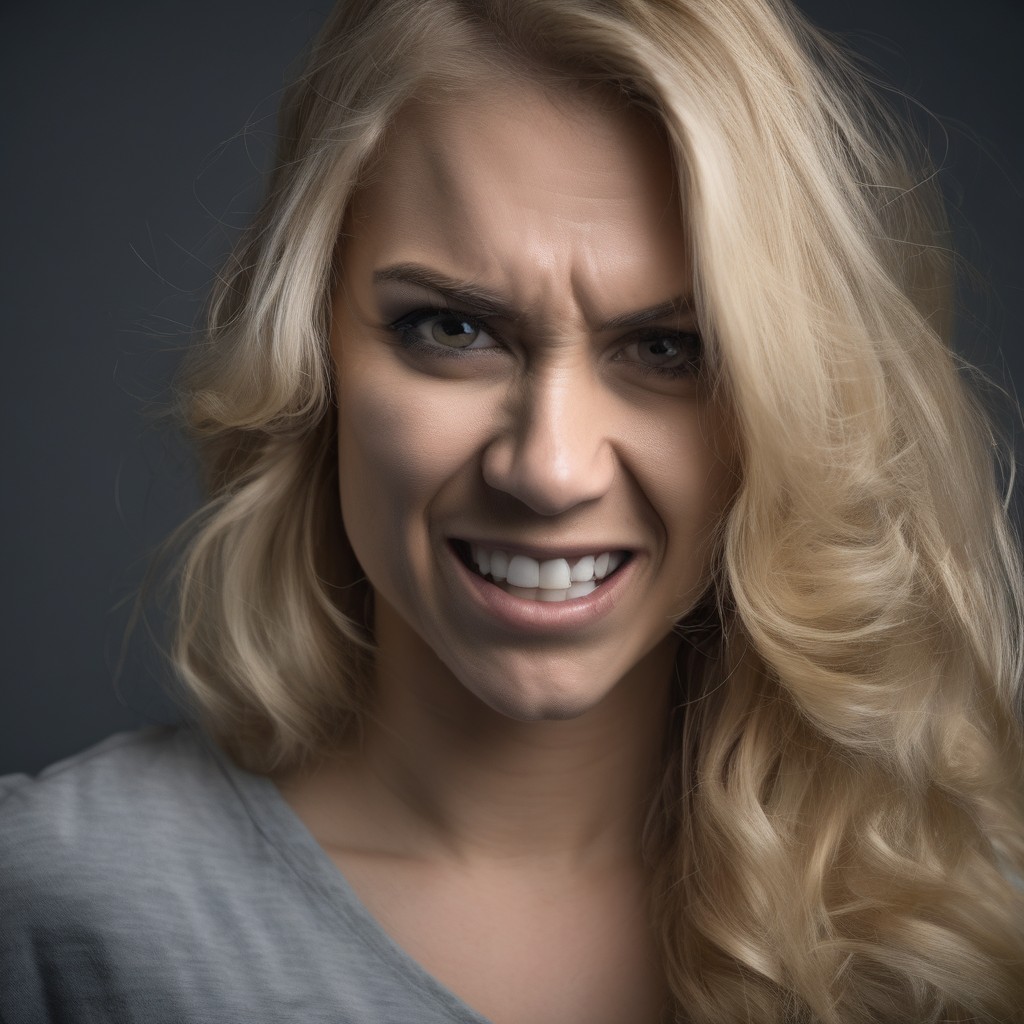}
            \end{subfigure}
        };
        \node[anchor=north west] (img35) at (5.4cm,-3.3cm) {
            \begin{subfigure}[t]{\comparisionsize}
                \includegraphics[width=\textwidth]{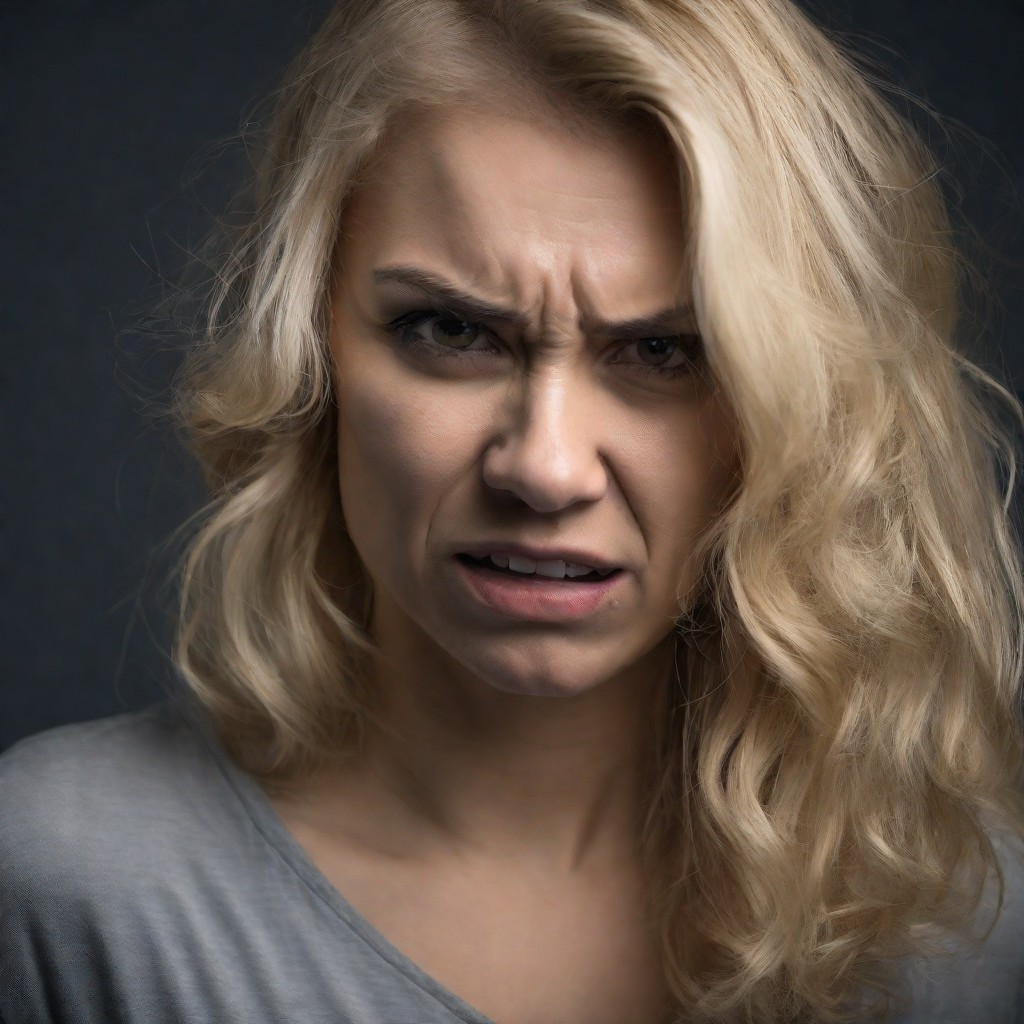}
            \end{subfigure}
        };
        \node[anchor=north west] (img35) at (6.75cm,-3.3cm) {
            \begin{subfigure}[t]{\comparisionsize}
                \includegraphics[width=\textwidth]{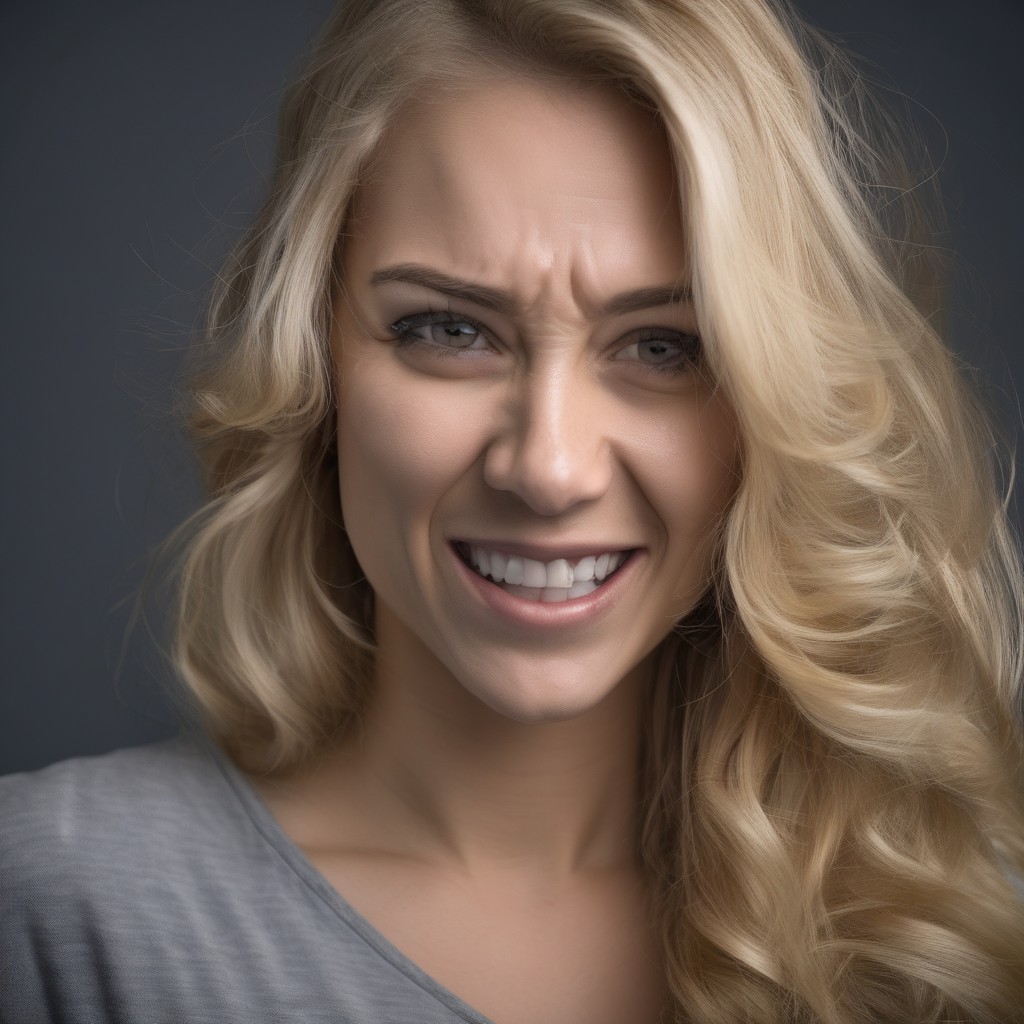}
            \end{subfigure}
        };

        \node[anchor=base,scale=\comparisonscale] at (4.2cm, -5.0cm) { shirt $\rightarrow$ {\color{red} suit}};
        \node[anchor=north west] (img11) at (0,-4.95cm) {
            \begin{subfigure}[t]{\comparisionsize}
                \includegraphics[width=\textwidth]{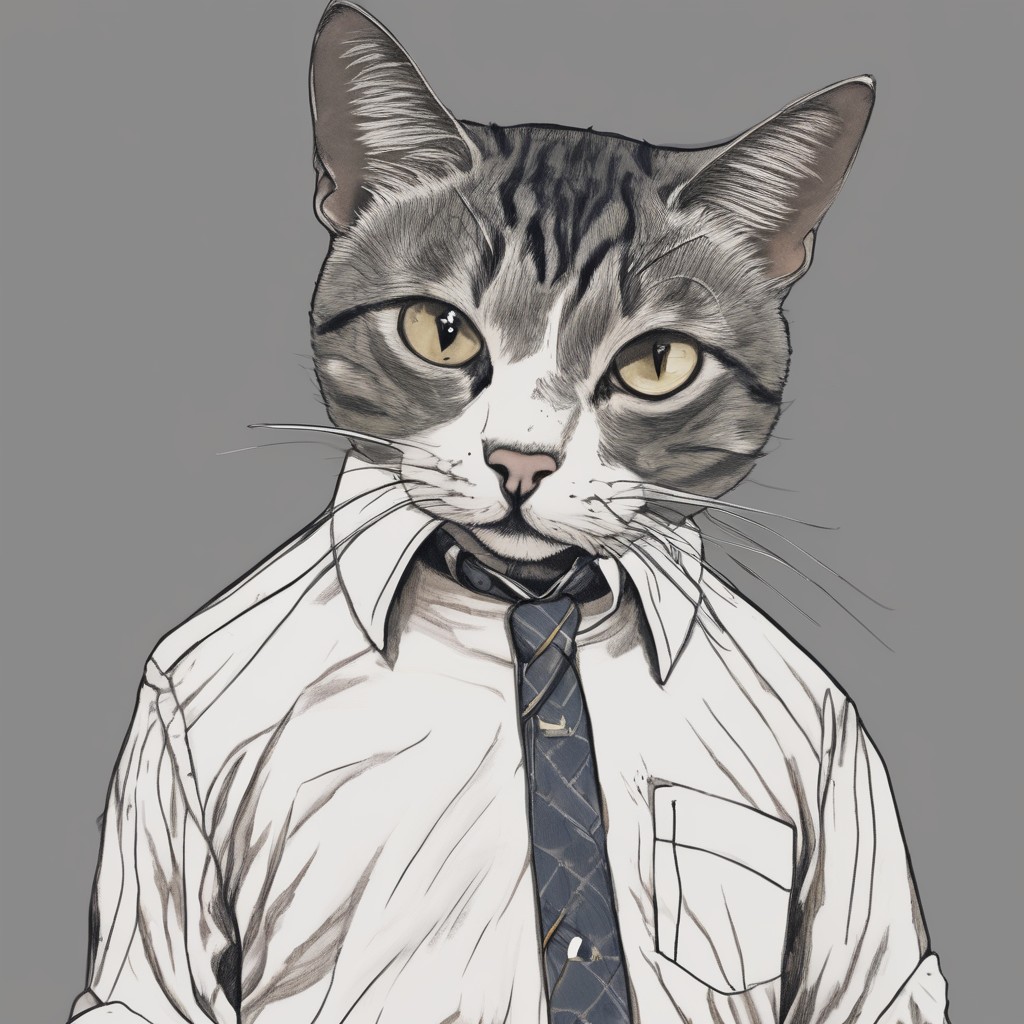}
            \end{subfigure}
        };
        \node[anchor=north west] (img12) at (1.35cm,-4.95cm) {
            \begin{subfigure}[t]{\comparisionsize}
                \includegraphics[width=\textwidth]{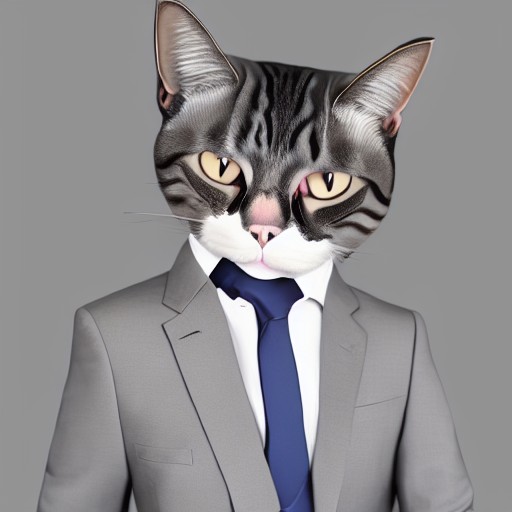}
            \end{subfigure}
        };
        \node[anchor=north west] (img13) at (2.7cm,-4.95cm) {
            \begin{subfigure}[t]{\comparisionsize}
                \includegraphics[width=\textwidth]{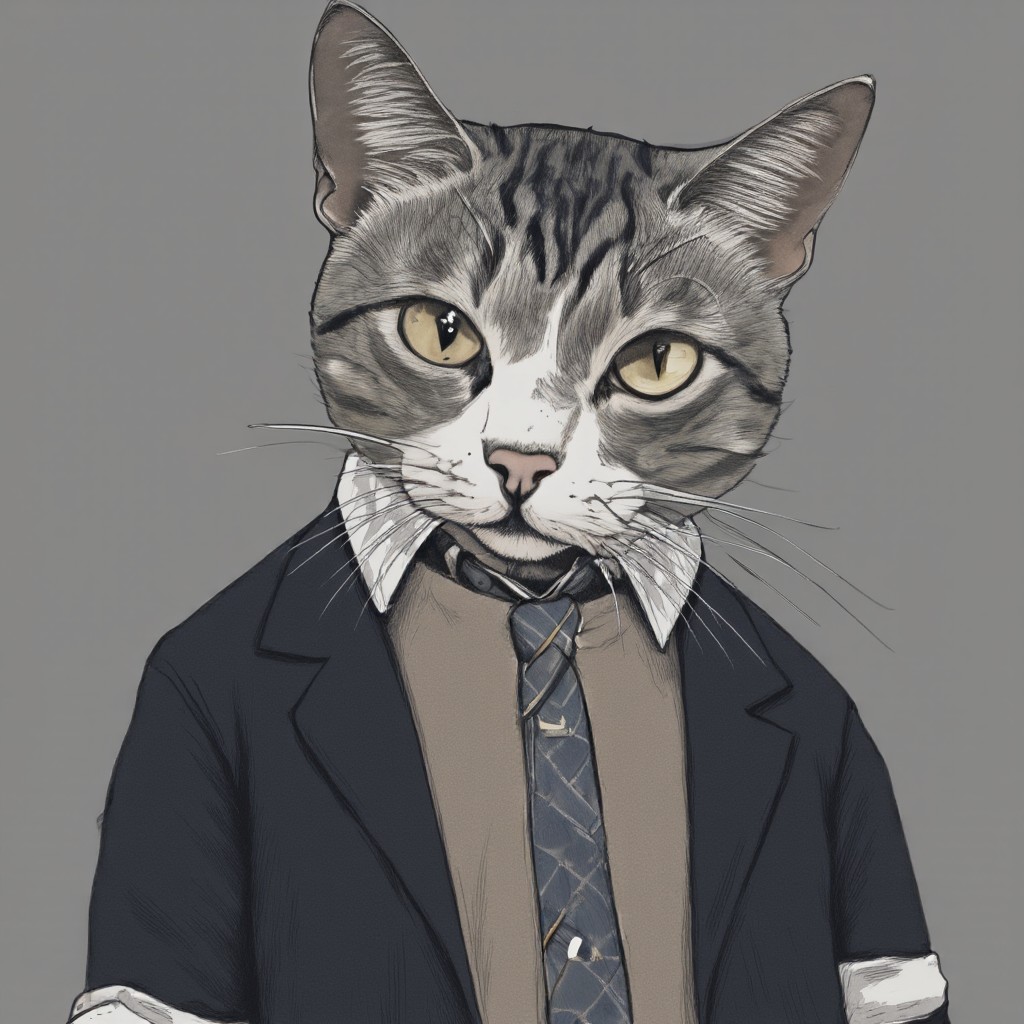}
            \end{subfigure}
        };
        \node[anchor=north west] (img14) at (4.05cm,-4.95cm) {
            \begin{subfigure}[t]{\comparisionsize}
                \includegraphics[width=\textwidth]{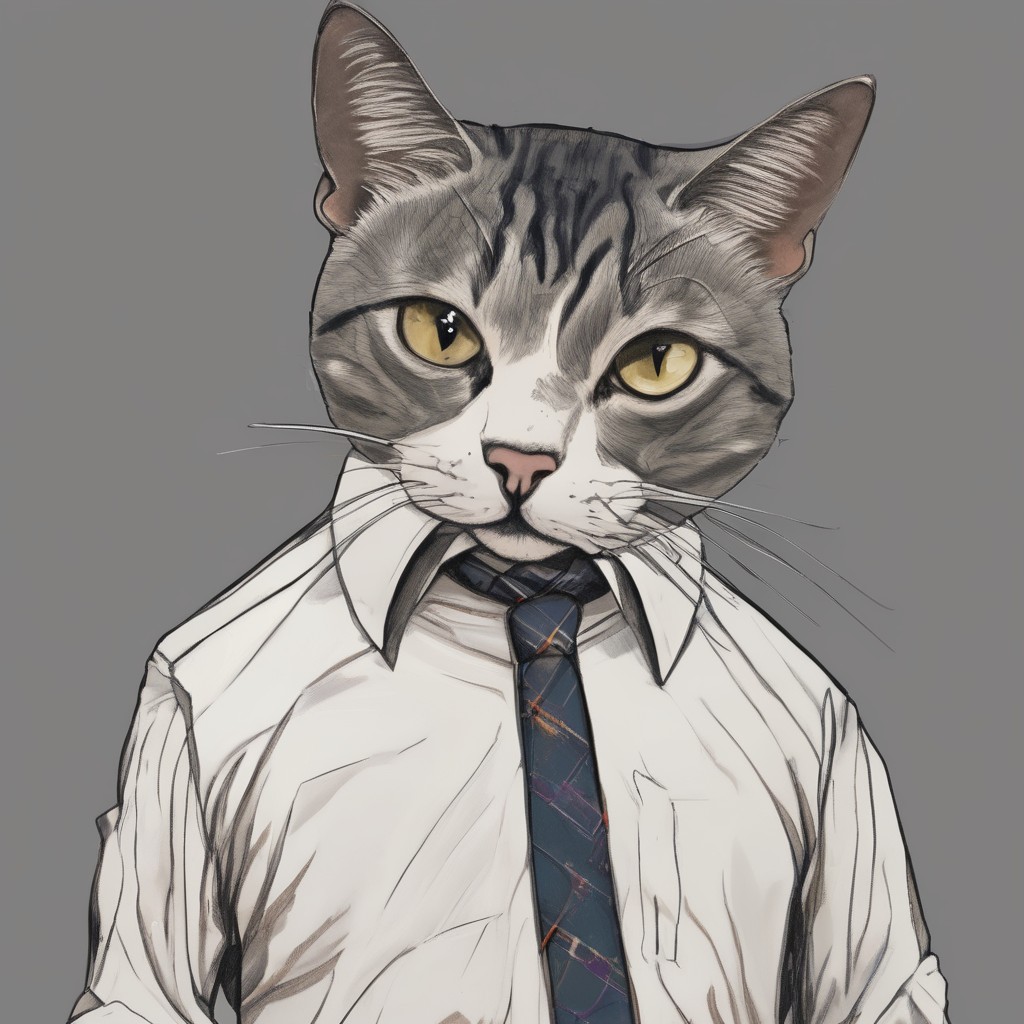}
            \end{subfigure}
        };
        \node[anchor=north west] (img15) at (5.4cm,-4.95cm) {
            \begin{subfigure}[t]{\comparisionsize}
                \includegraphics[width=\textwidth]{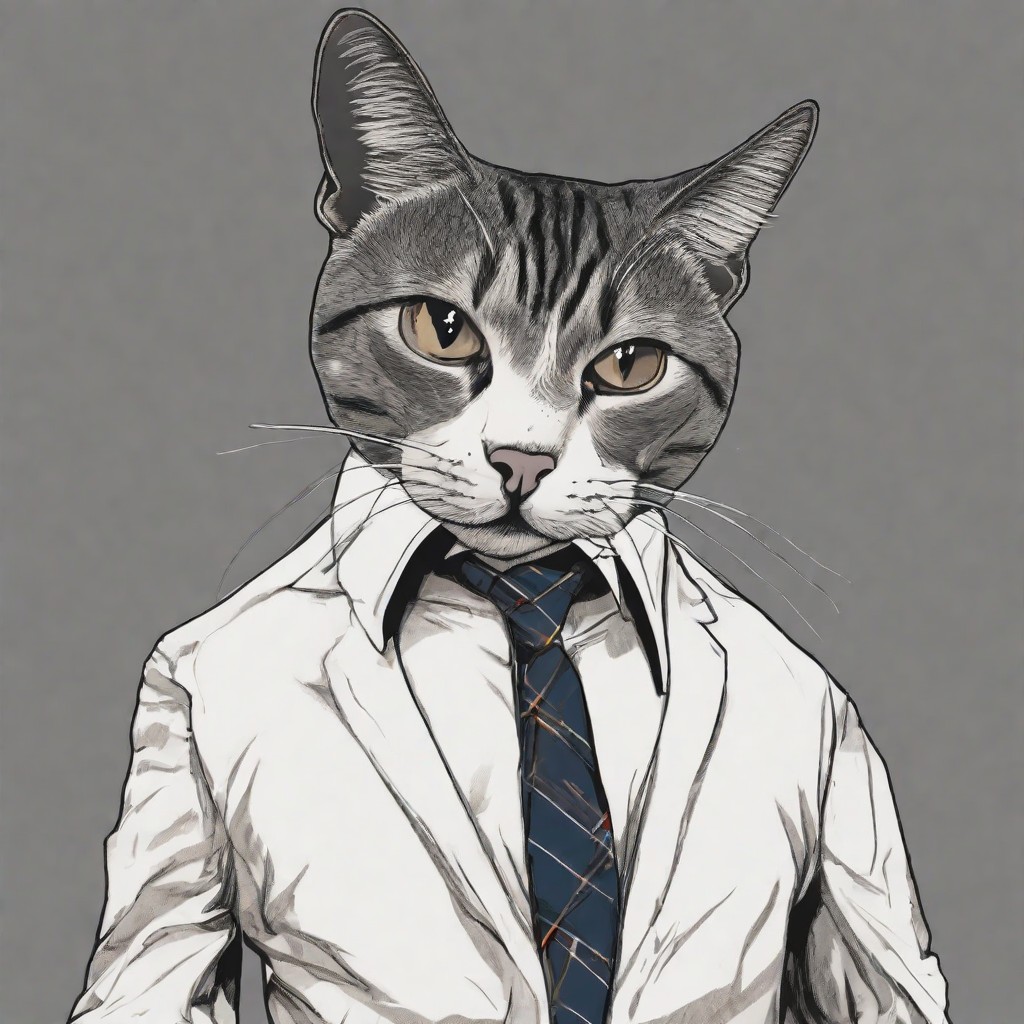}
            \end{subfigure}
        };
        \node[anchor=north west] (img15) at (6.75cm,-4.95cm) {
            \begin{subfigure}[t]{\comparisionsize}
                \includegraphics[width=\textwidth]{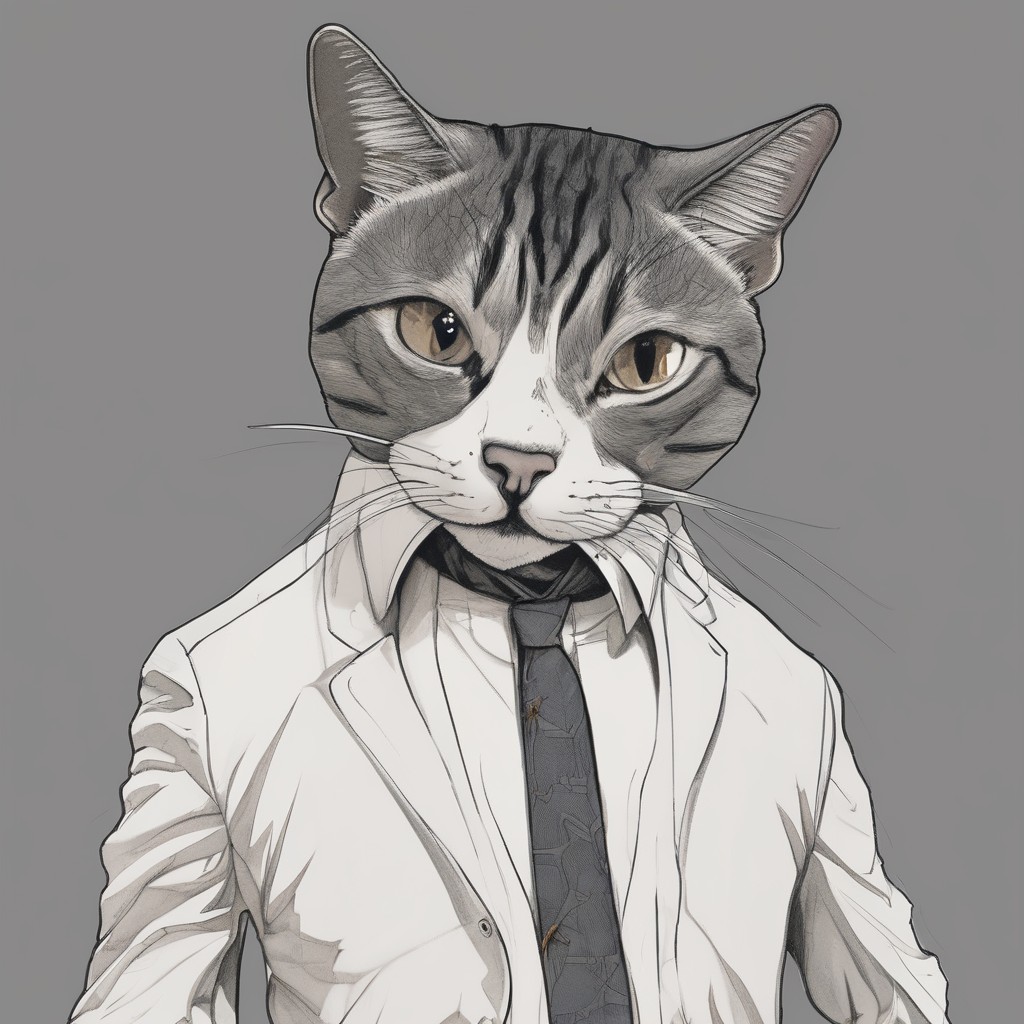}
            \end{subfigure}
        };
        \node[anchor=base,scale=\comparisonscale] at (0.8cm, -6.6cm) {\small \textbf{Origin}};
\node[anchor=base,scale=\comparisonscale] at (2.1cm, -6.6cm) {\small \textbf{InPix2Pix}};
\node[anchor=base,scale=\comparisonscale] at (3.45cm, -6.6cm) {\small \textbf{FLUX.Fill}};
\node[anchor=base,scale=\comparisonscale] at (4.8cm, -6.6cm) {\small \textbf{P2P}};
\node[anchor=base,scale=\comparisonscale] at (6.15cm, -6.6cm) {\small \textbf{Masactrl}};
\node[anchor=base,scale=\comparisonscale] at (7.55cm, -6.6cm) {\small \textbf{Ours}};

    \end{tikzpicture}
    \vspace{-5pt}
    \caption{Qualitative comparison with other methods.}
    \label{fig:comparison}
    \vspace{-5pt}
\end{figure}

\noindent \textbf{Qualitative Comparison}. As shown in Fig.~\ref{fig:comparison}, the results clearly demonstrate distinctions between our method and others. The first three rows reveal FLUX.Fill strictly confines new objects within masked regions to avoid altering other parts, leading to conflicts and unsuccessful outcomes. In contrast, our method enhances object generation accuracy by replacing text embeddings, allowing objects to extend beyond mask constraints. While P2P and Masactrl perform well in object replacement, they struggle to preserve background details. Additionally, they require strict target-source prompt alignment. Our method relaxes this constraint while achieving better results through flexible timestep intervals and Softbox. For instance, the third row shows background color shifts with other methods, whereas ours preserves the original color. See Appendix A.3 for more comparison results.
\begin{table}[!t]
\small
\setlength{\tabcolsep}{3.2pt}
\caption{Comparative results of different methods. Bold represents the best, and underline represents the second best.}
\vspace{-10pt}
\begin{tabular}{cccccc}
\specialrule{.1em}{.05em}{.05em}
Method            & train & PSNR$\uparrow$    & MSE $ \scriptscriptstyle\times 10^{4}$ $\downarrow$ & Lpips $ \scriptscriptstyle\times 10^{3}$ $\downarrow$ & CLIP Sim$\uparrow$ \\ \hline
P2P(SDXL)                       &       & 21.67 & 81.18       & 1.63         & 24.87      \\
Masactrl(SDXL)                  &       & 20.90 & 94.87       & 1.68         & \textbf{25.26}      \\
InPix2Pix(SDXL)         & \checkmark     & 18.77 & 169.91      & 2.20         & 23.58      \\
Flux.Fill(DiT)               & \checkmark     & \textbf{33.51} & \textbf{5.67}        & \textbf{0.14}        & 24.79      \\ \hline
Our(SDXL)                       &       & \underline{26.21} & \underline{35.95}       & \underline{0.99}         & \underline{25.21}     \\ 
\specialrule{.1em}{.05em}{.05em}
\end{tabular}
\label{tab:comparision}
\vspace{-5pt}
\end{table}
\begin{figure}[]
    \centering
    \begin{tikzpicture}
    \node[anchor=north west]  at (0,0) {
    \begin{subfigure}[b]{\locationpicsize}
        \includegraphics[width=\textwidth]{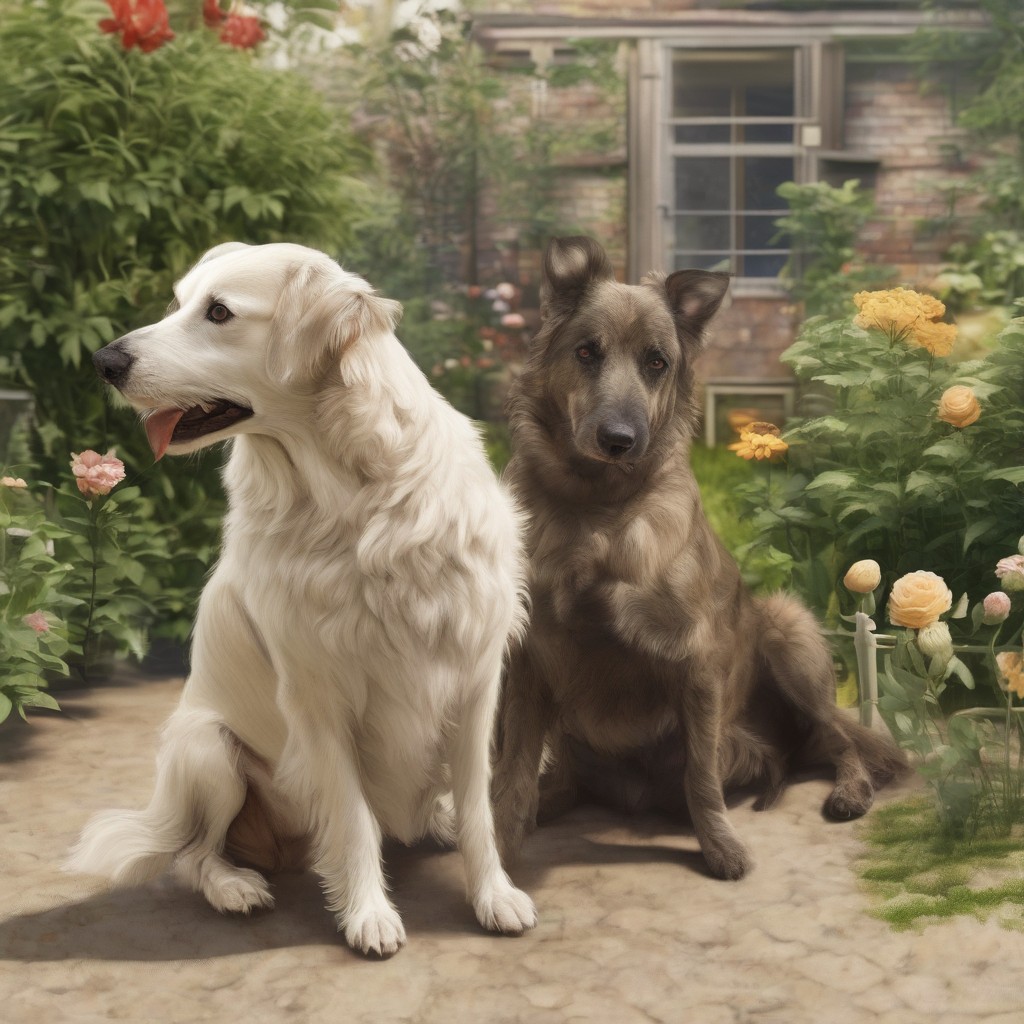}
    \end{subfigure}
    };
    \node[anchor=north west]  at (1.35cm,0) {
    \begin{subfigure}[b]{\locationpicsize}
        \includegraphics[width=\textwidth]{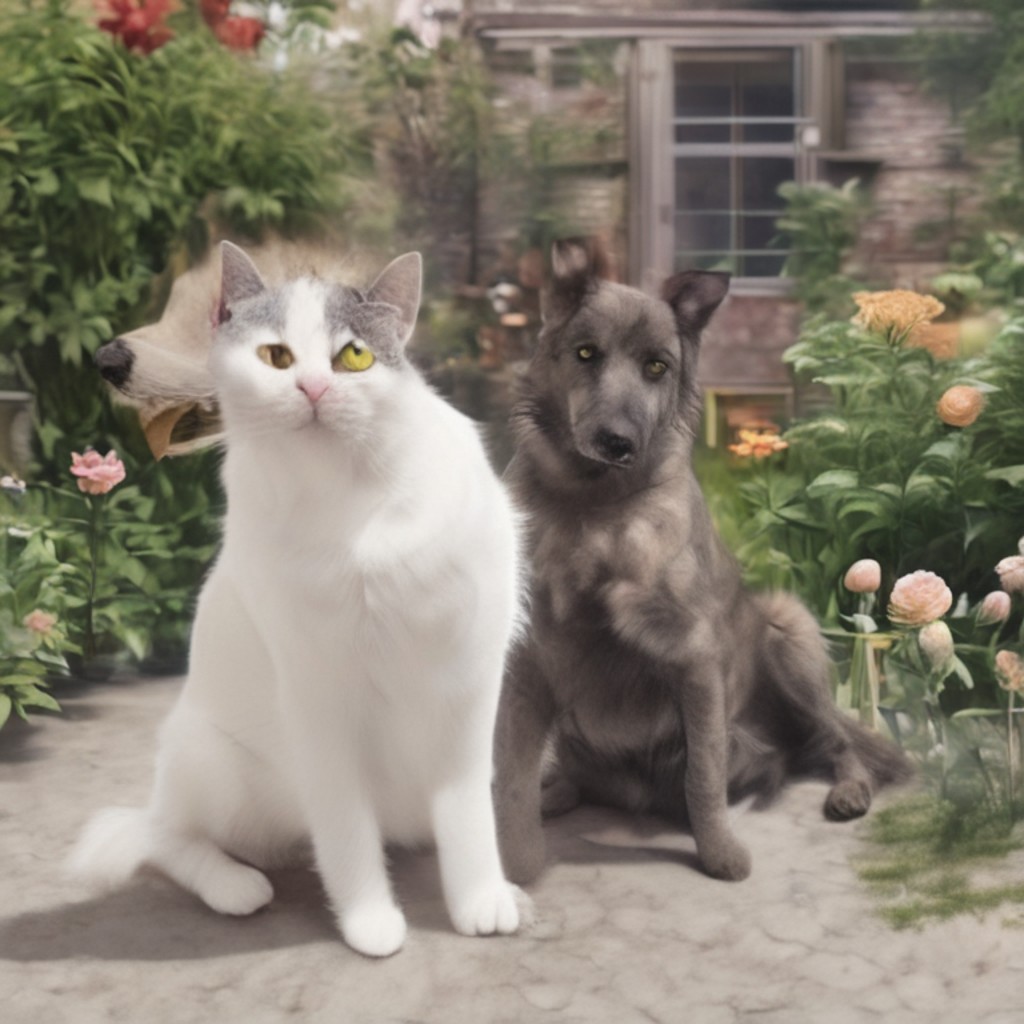}
    \end{subfigure}
    };
    \node[anchor=north west]  at (2.7cm,0) {
    \begin{subfigure}[b]{\locationpicsize}
        \includegraphics[width=\textwidth]{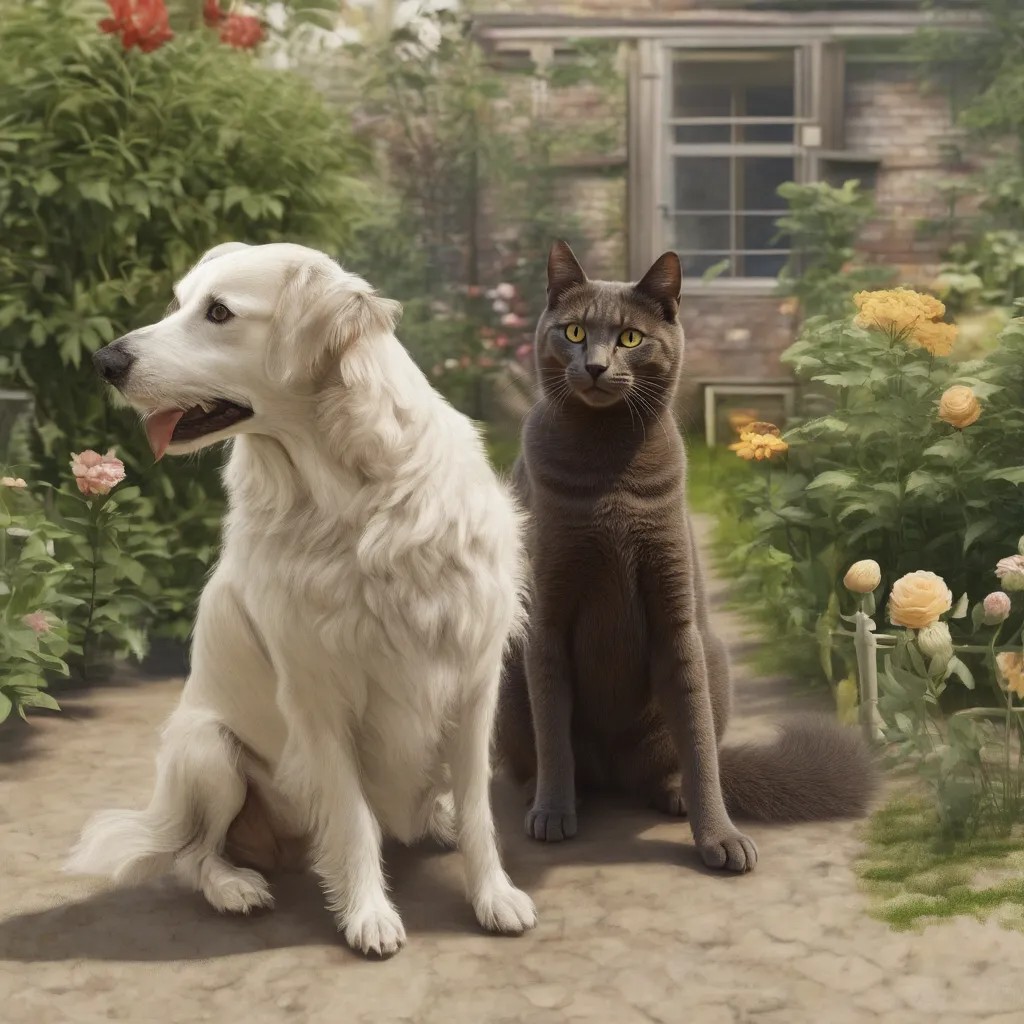}
    \end{subfigure}
    };
    \node[anchor=north west]  at (4.05cm,0) {
    \begin{subfigure}[b]{\locationpicsize}
        \includegraphics[width=\textwidth]{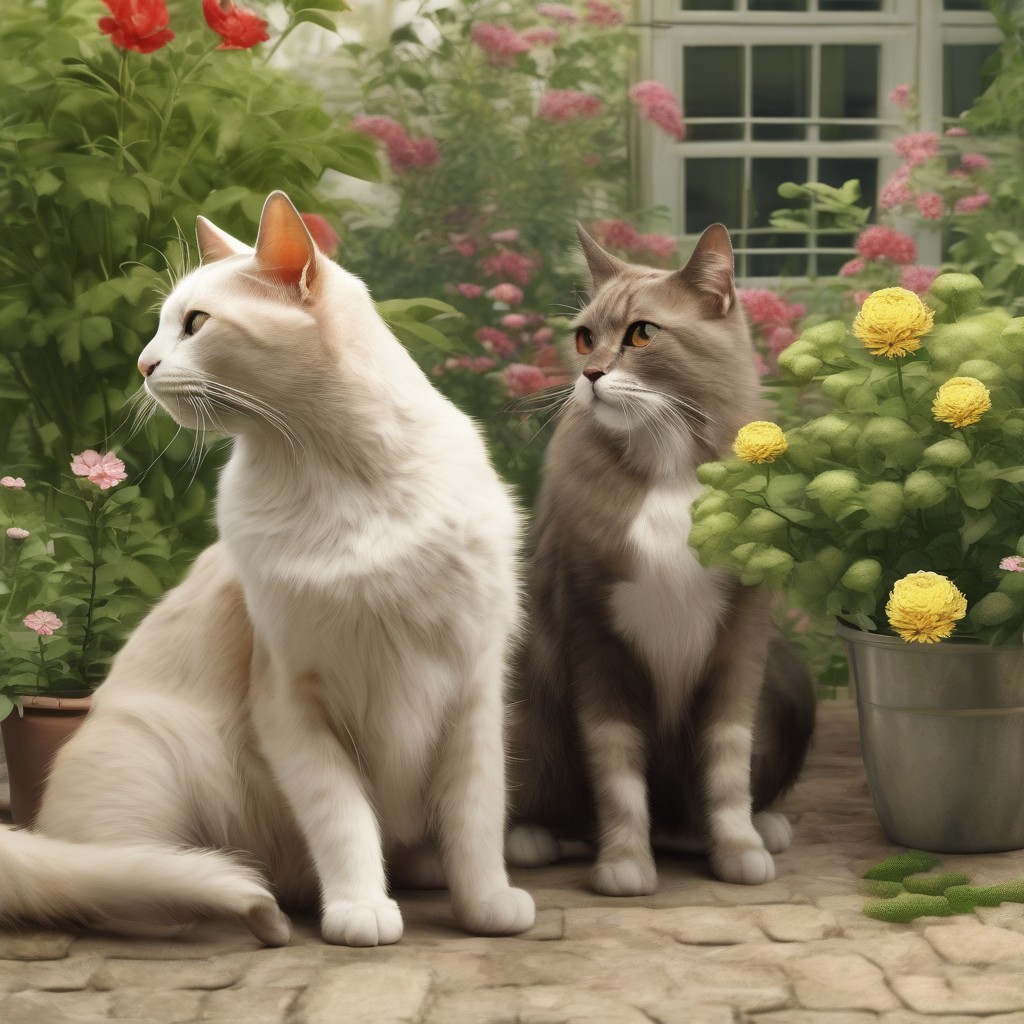}
    \end{subfigure}
    };
    \node[anchor=north west]  at (5.4cm,0) {
    \begin{subfigure}[b]{\locationpicsize}
        \includegraphics[width=\textwidth]{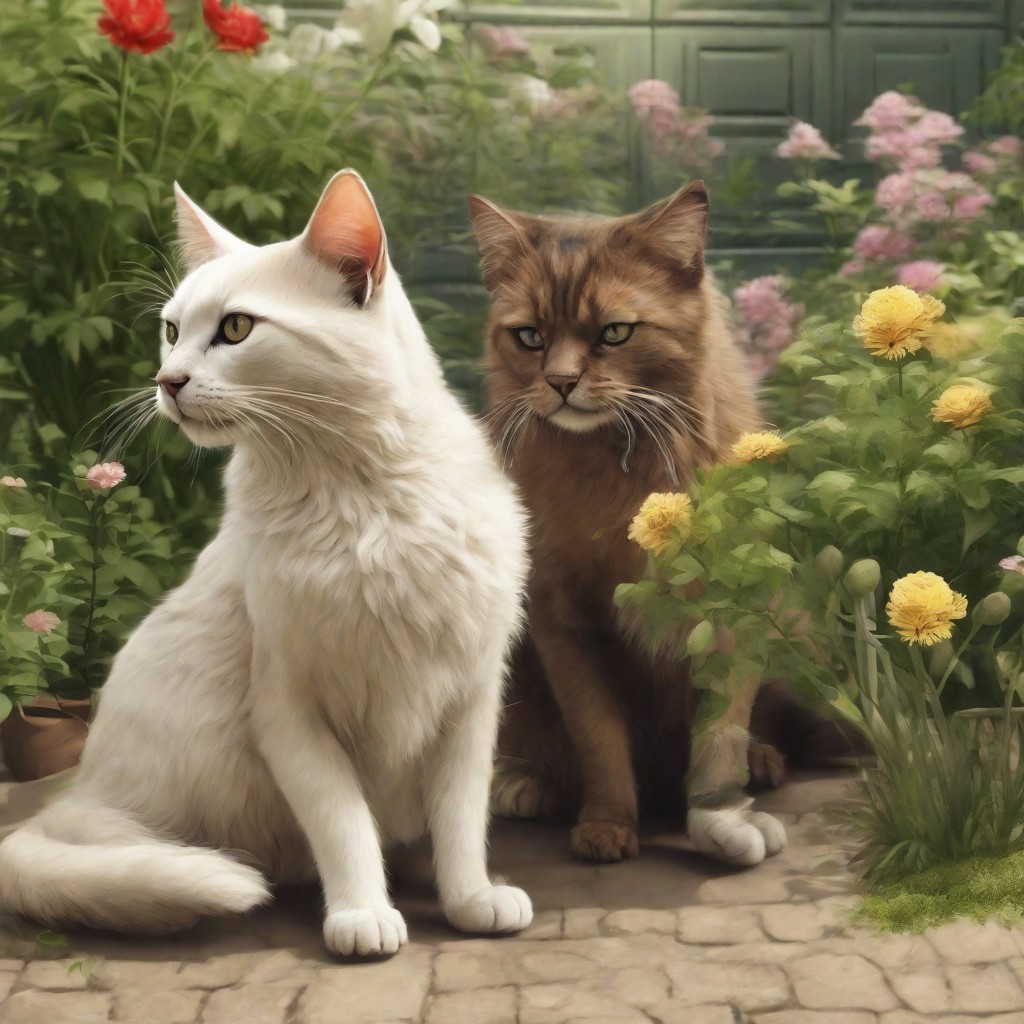}
    \end{subfigure}
    };
    \node[anchor=north west]  at (6.75cm,0) {
    \begin{subfigure}[b]{\locationpicsize}
        \includegraphics[width=\textwidth]{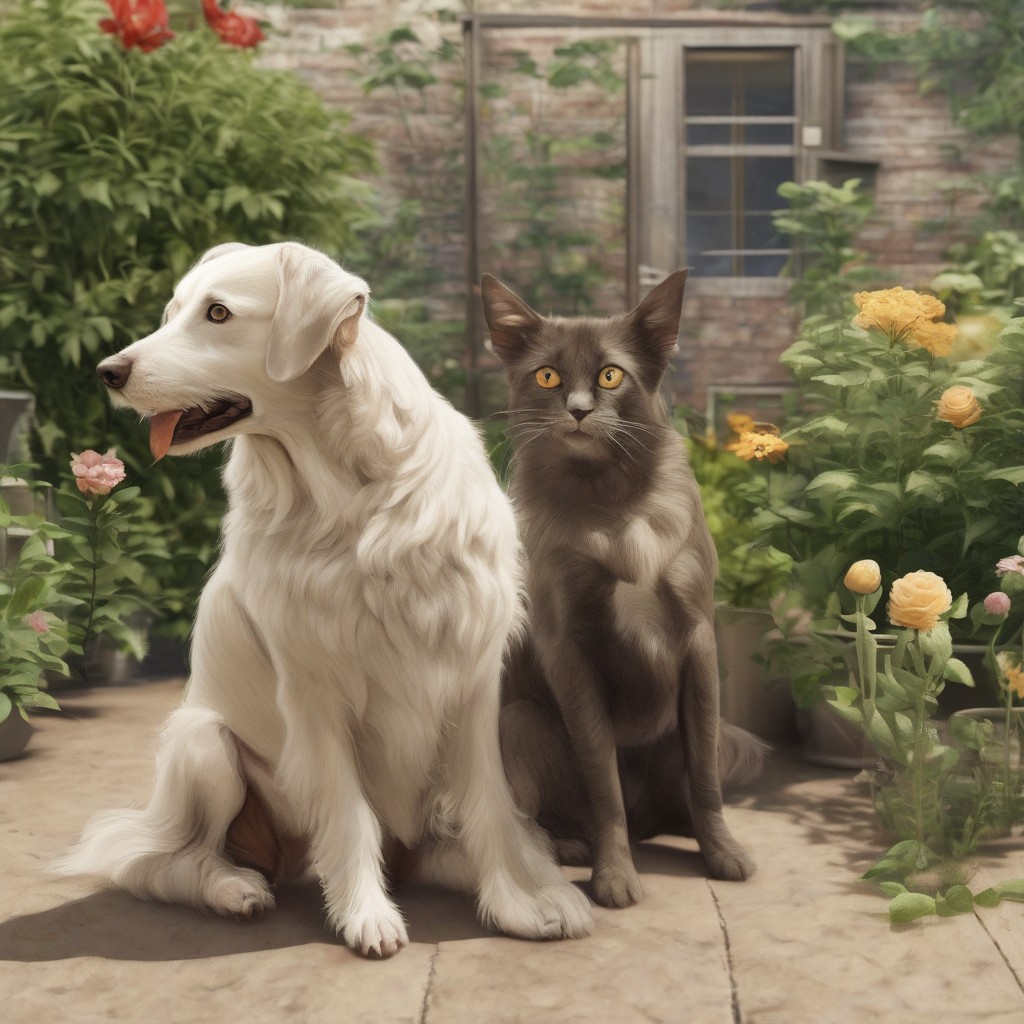}
    \end{subfigure}
    };

    \node[anchor=north west]  at (0,-1.4cm) {
    \begin{subfigure}[b]{\locationpicsize}
        \includegraphics[width=\textwidth]{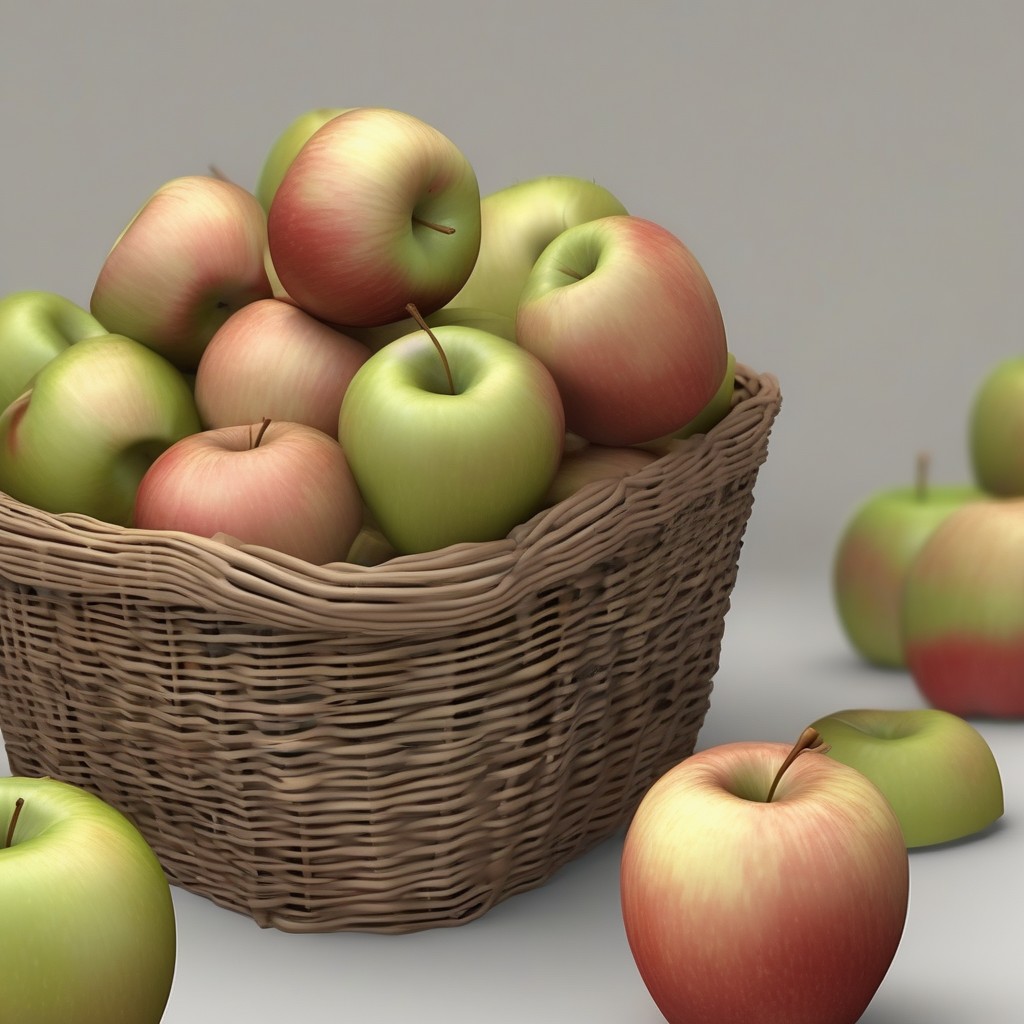}
        \vspace{-18pt}
        \caption*{Origin}
    \end{subfigure}
    };
    \node[anchor=north west]  at (1.35cm,-1.4cm) {
    \begin{subfigure}[b]{\locationpicsize}
        \includegraphics[width=\textwidth]{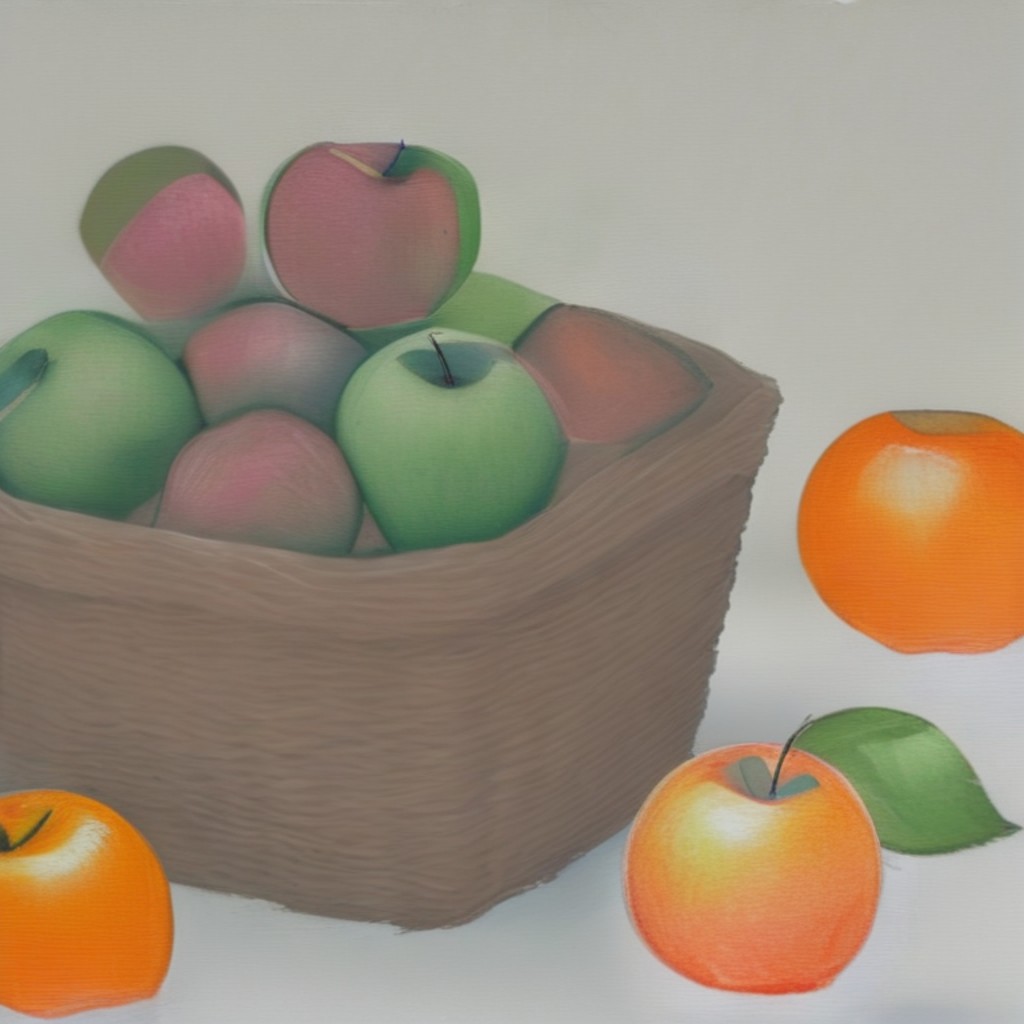}
        \vspace{-18pt}
        \caption*{InPix2Pix}
    \end{subfigure}
    };
    \node[anchor=north west]  at (2.7cm,-1.4cm) {
    \begin{subfigure}[b]{\locationpicsize}
        \includegraphics[width=\textwidth]{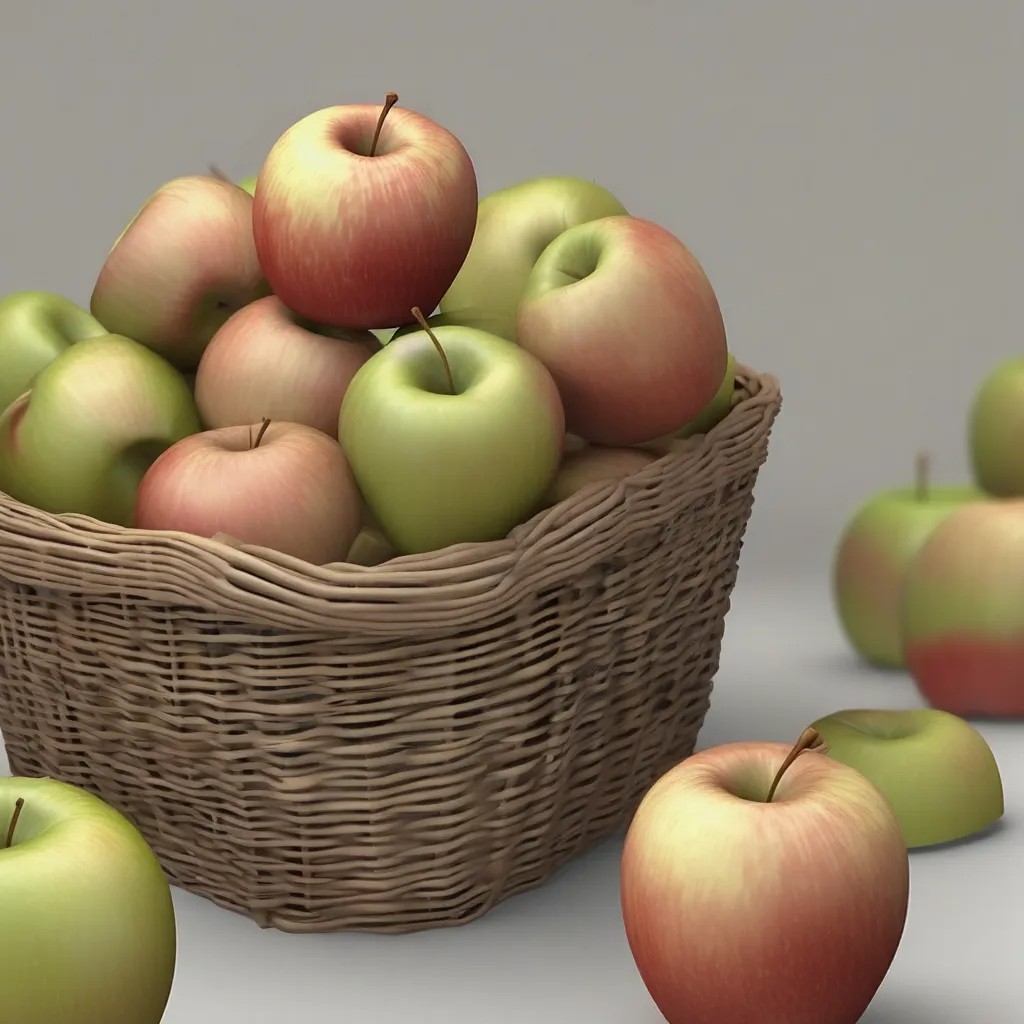}
        \vspace{-18pt}
        \caption*{FLUX.Fill}
    \end{subfigure}
    };
    \node[anchor=north west]  at (4.05cm,-1.4cm) {
    \begin{subfigure}[b]{\locationpicsize}
        \includegraphics[width=\textwidth]{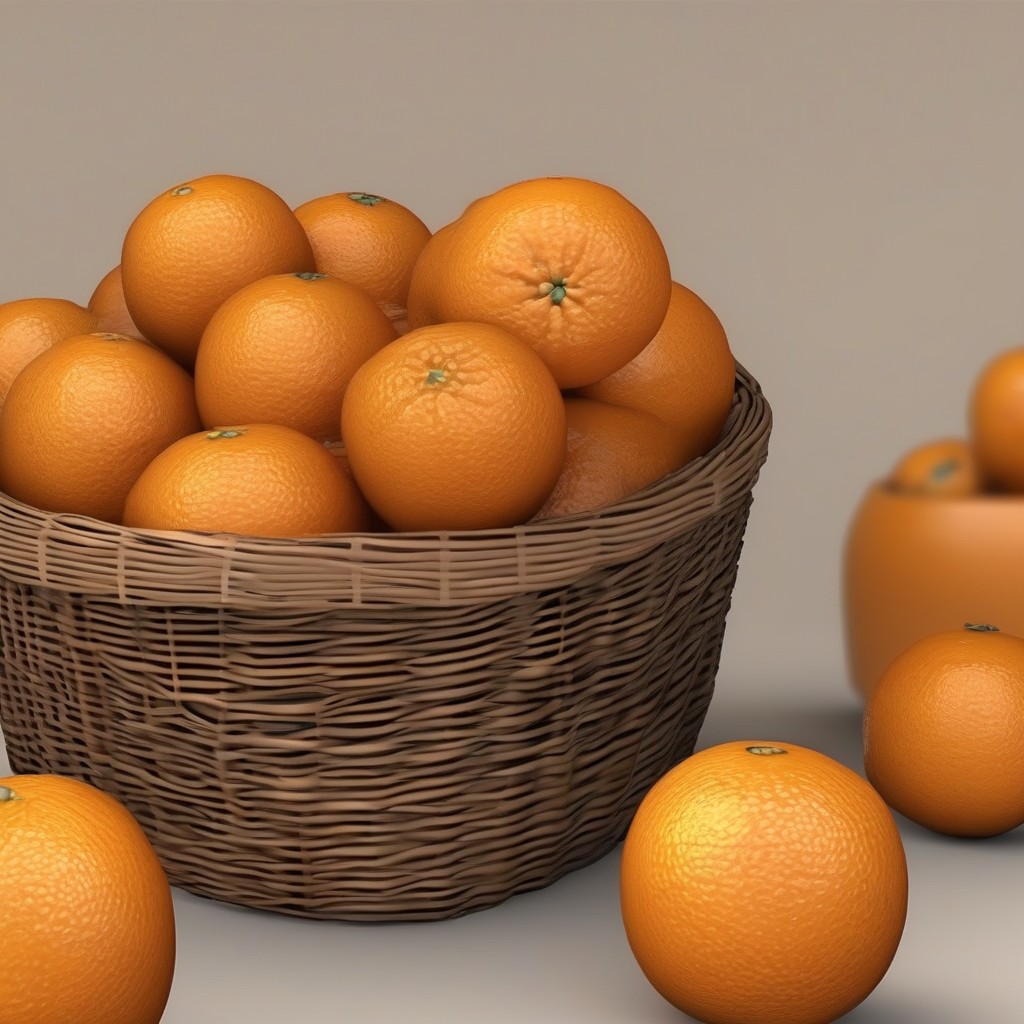}
        \vspace{-18pt}
        \caption*{P2P}
    \end{subfigure}
    };
    \node[anchor=north west]  at (5.4cm,-1.4cm) {
    \begin{subfigure}[b]{\locationpicsize}
        \includegraphics[width=\textwidth]{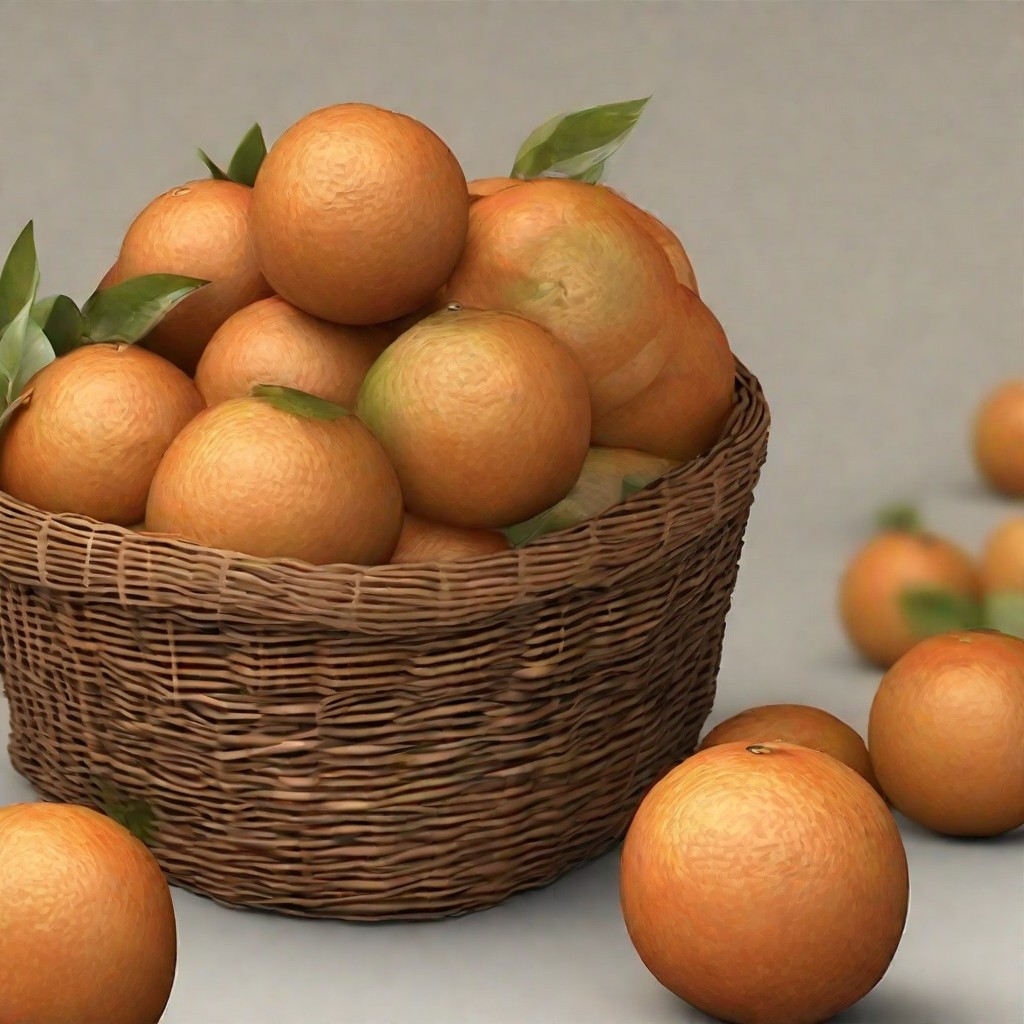}
        \vspace{-18pt}
        \caption*{Masactrl}
    \end{subfigure}
    };
    \node[anchor=north west]  at (6.75cm,-1.4cm) {
    \begin{subfigure}[b]{\locationpicsize}
        \includegraphics[width=\textwidth]{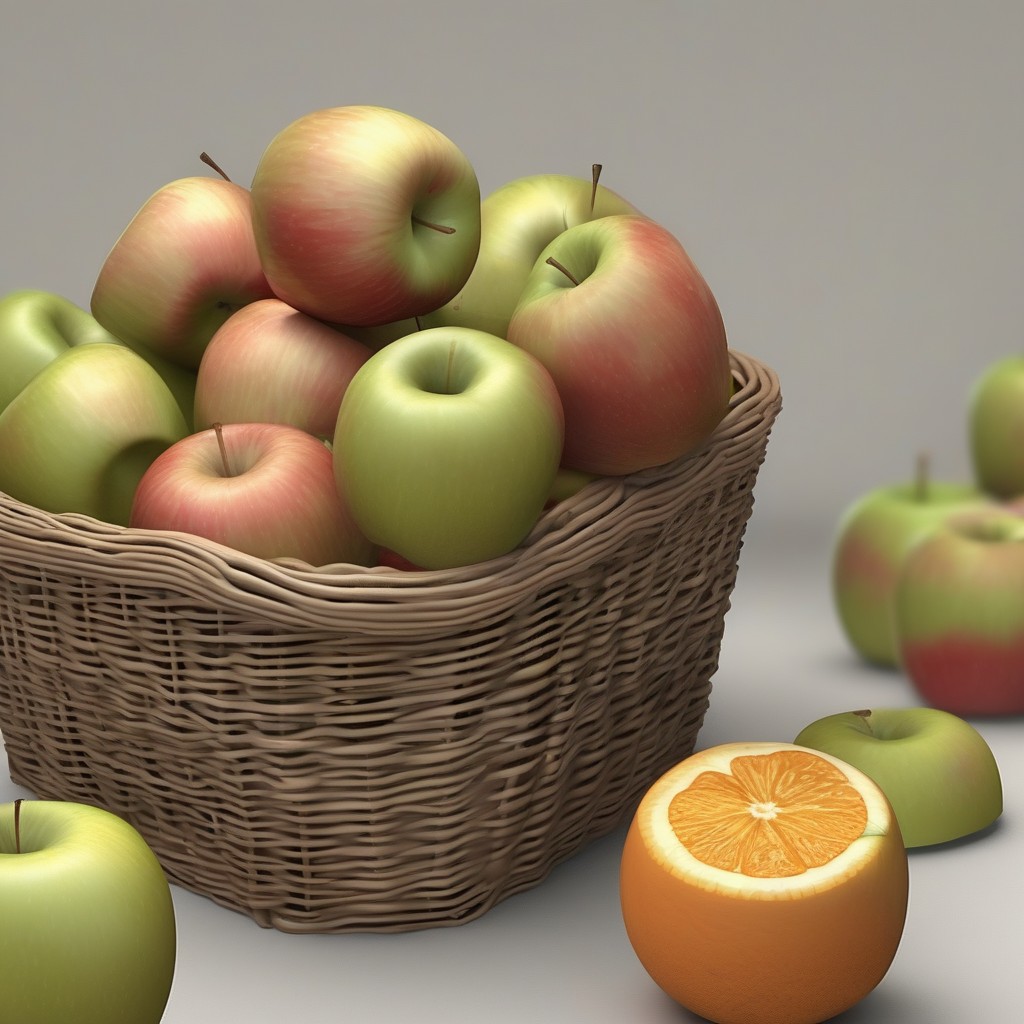}
        \vspace{-18pt}
        \caption*{Ours}
    \end{subfigure}
    };
    \end{tikzpicture}
    \vspace{-5pt}
    \caption{Comparison of local object replacement using different methods. In the first row, the dog on the right is replaced with a cat; in the second row, the apple at the bottom-right corner is replaced with an orange.}
    \label{fig:location}
    \vspace{-5pt}
\end{figure}

\begin{figure}[!t]
    \centering
    \begin{tikzpicture}
    \node[anchor=north west]  at (0,0) {
   \begin{subfigure}[b]{\locationpicsize}
        \includegraphics[width=\textwidth]{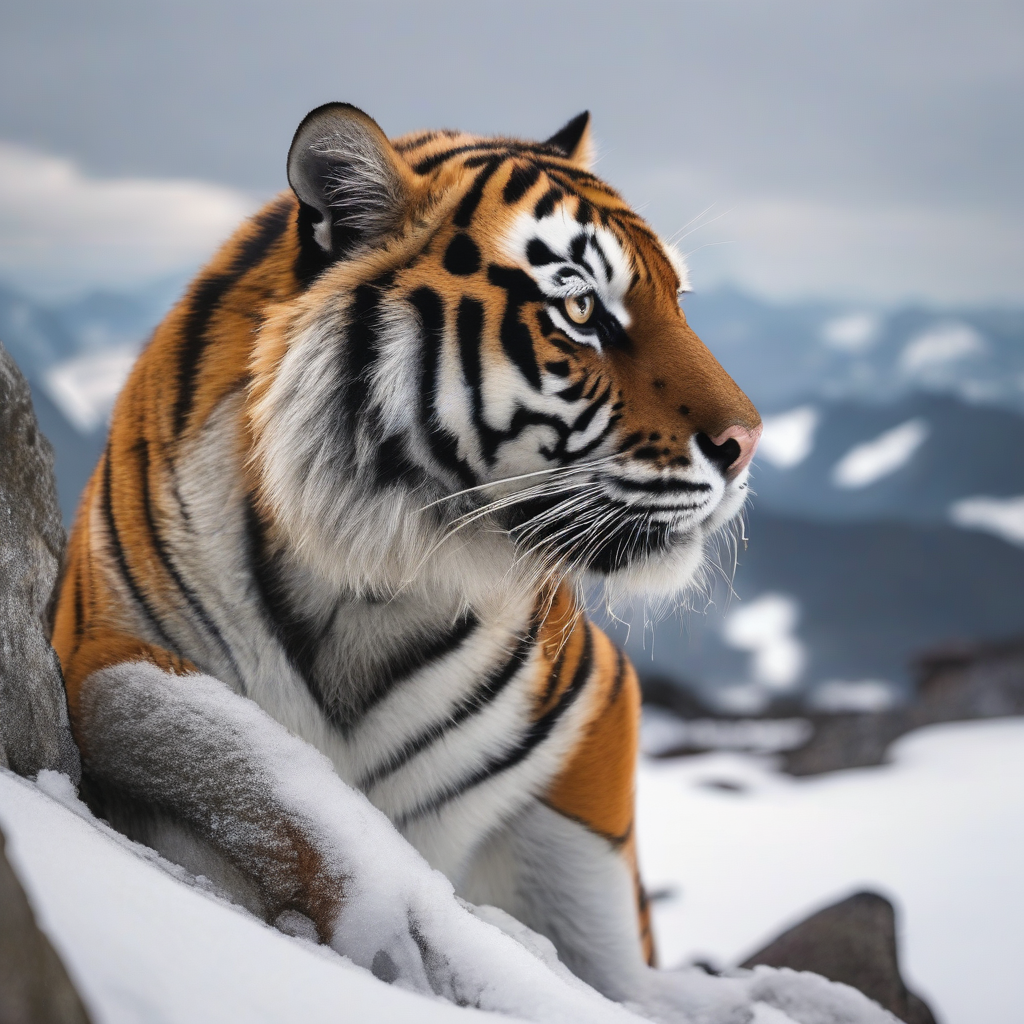}
    \end{subfigure}};
    \node[anchor=north west]  at (1.35cm,0) {
    \begin{subfigure}[b]{\locationpicsize}
        \includegraphics[width=\textwidth]{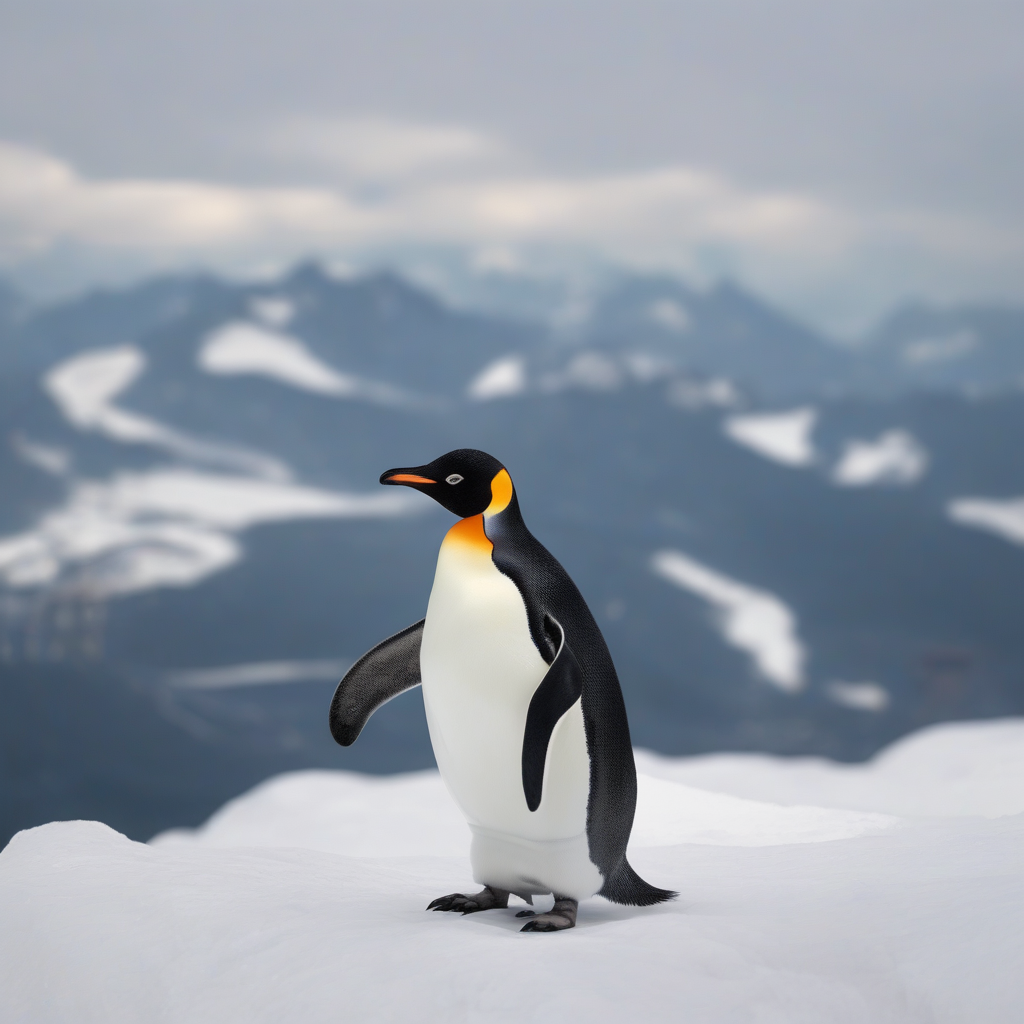}
    \end{subfigure}};
    \node[anchor=north west]  at (2.7cm,0) {
    \begin{subfigure}[b]{\locationpicsize}
        \includegraphics[width=\textwidth]{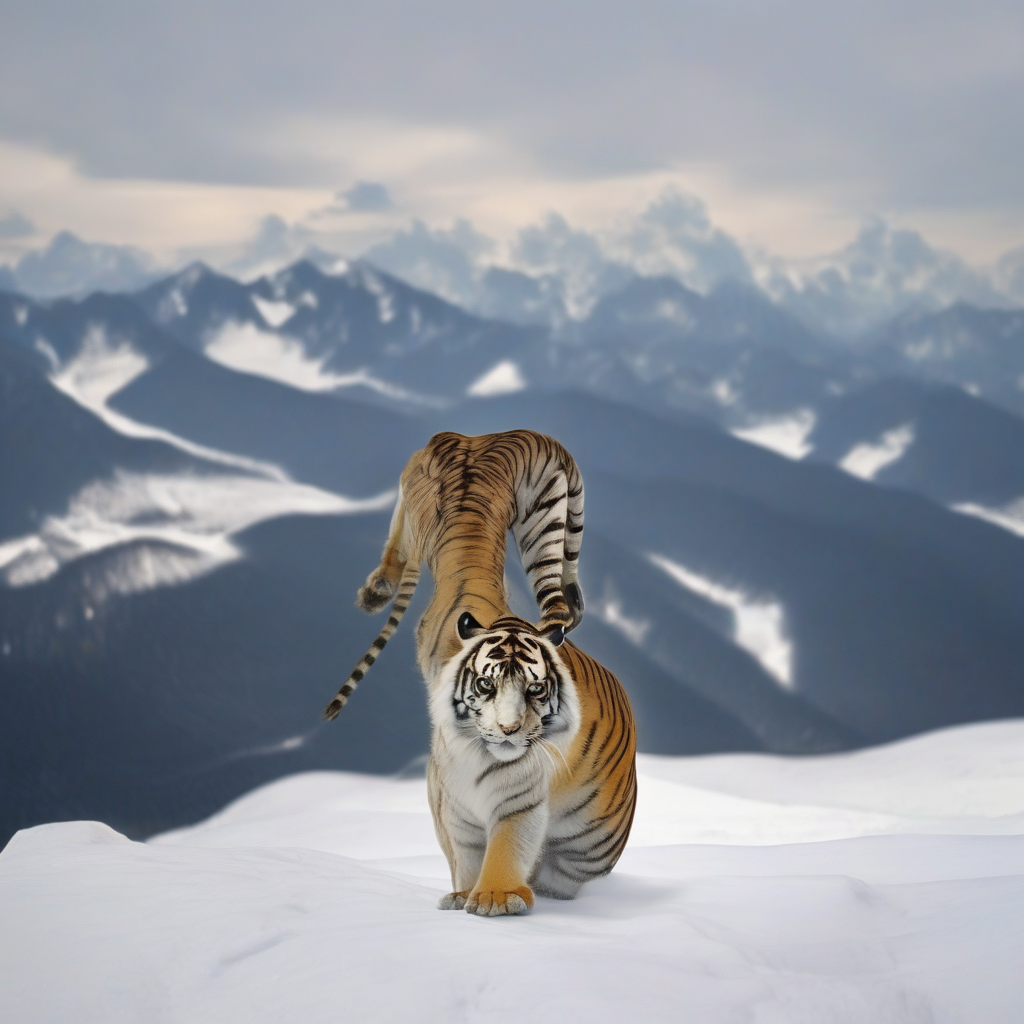}
    \end{subfigure}};
    \node[anchor=north west]  at (4.05cm,0) {
    \begin{subfigure}[b]{\locationpicsize}
        \includegraphics[width=\textwidth]{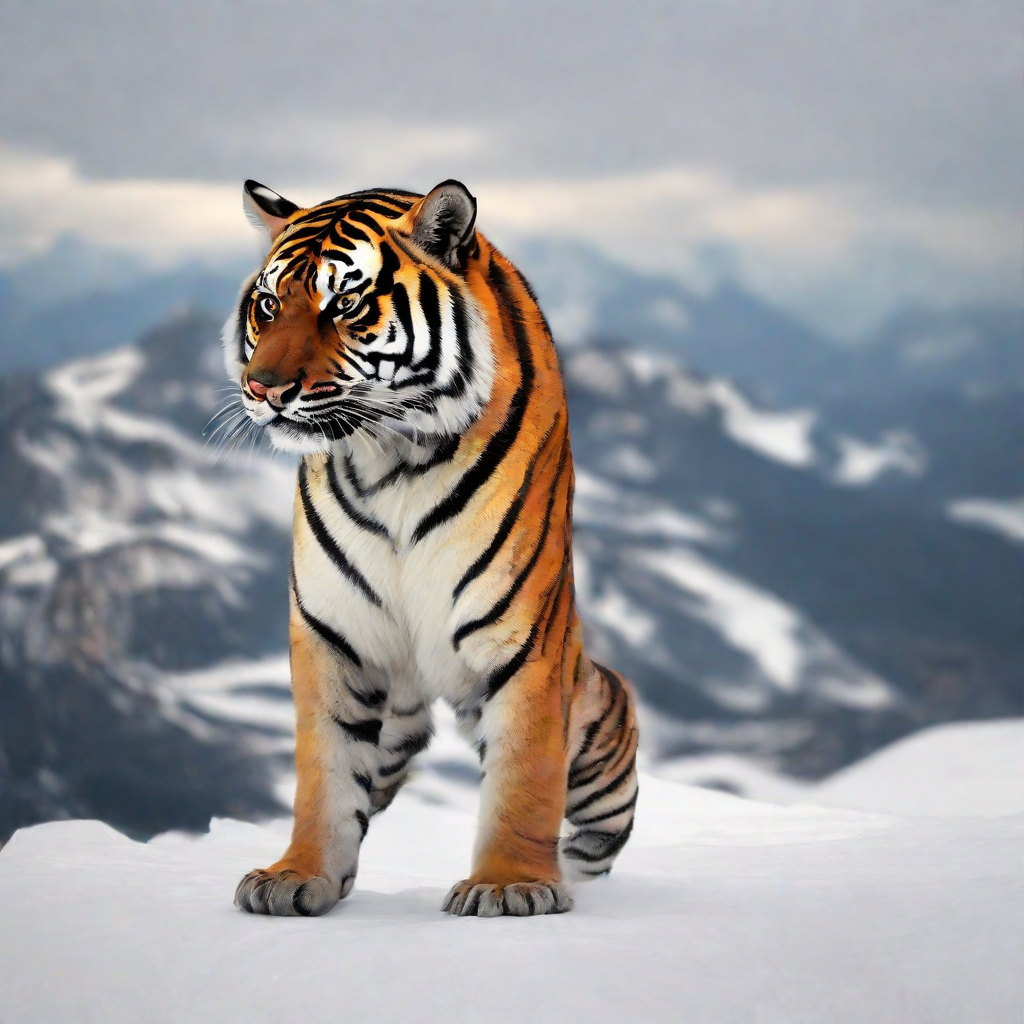}
    \end{subfigure}};
    \node[anchor=north west]  at (5.4cm,0) {
    \begin{subfigure}[b]{\locationpicsize}
        \includegraphics[width=\textwidth]{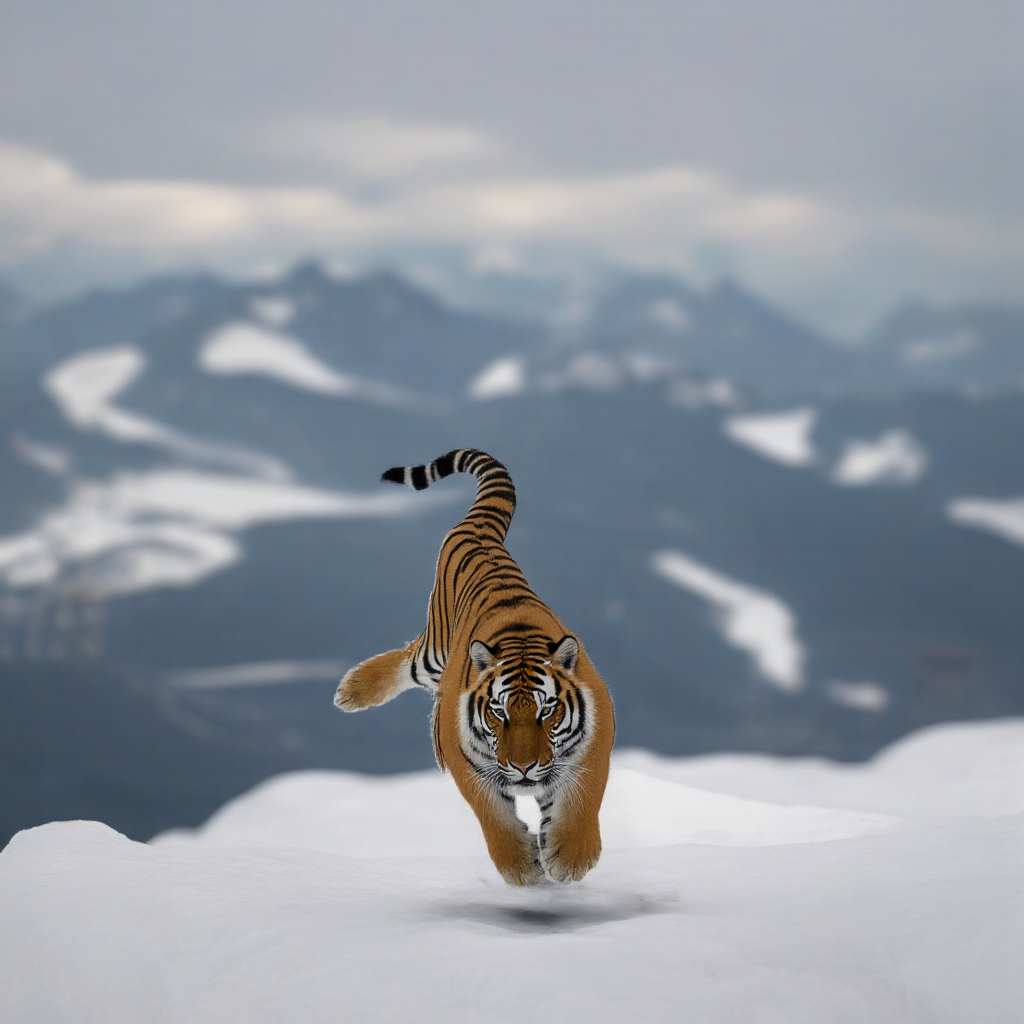}
    \end{subfigure}};
    \node[anchor=north west]  at (6.75cm,0) {
    \begin{subfigure}[b]{\locationpicsize}
        \includegraphics[width=\textwidth]{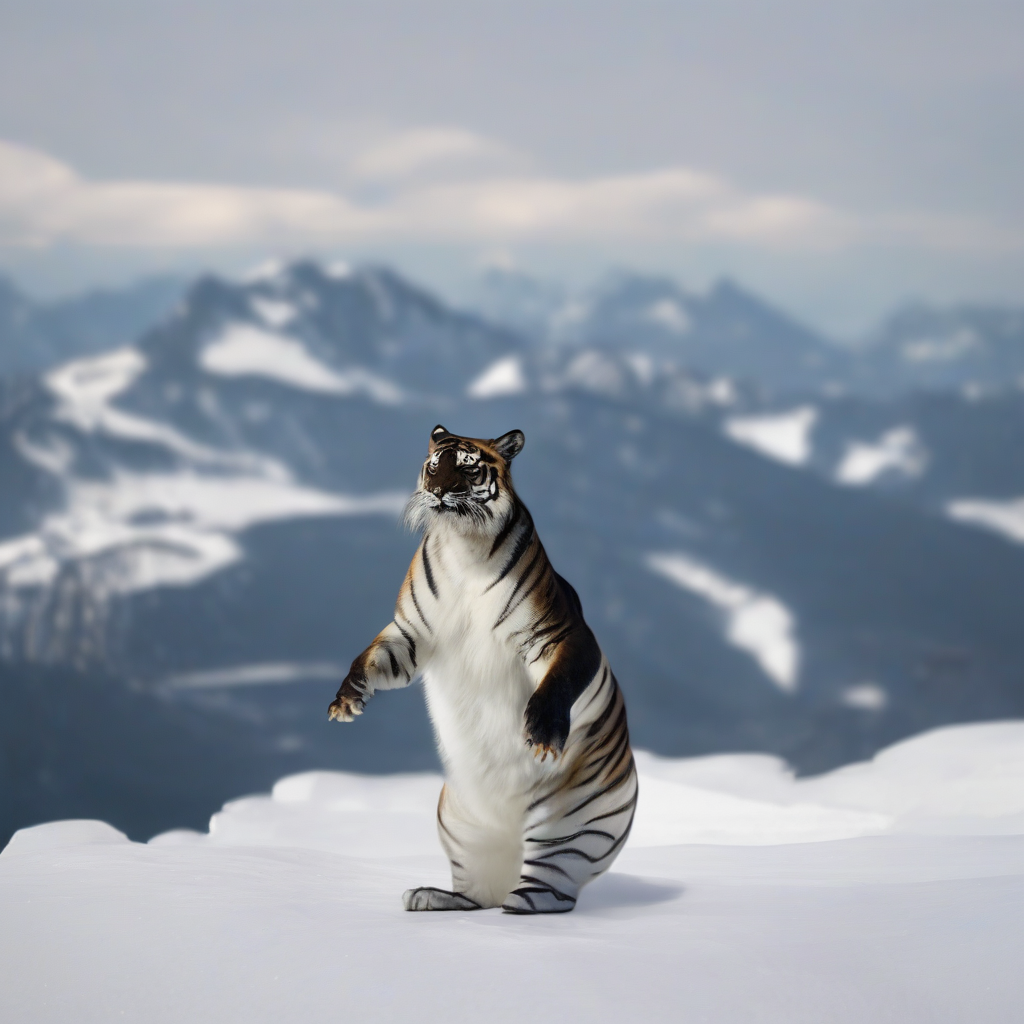}
    \end{subfigure}};

    \node[anchor=north west]  at (0,-1.4cm) {
    \begin{subfigure}[b]{\locationpicsize}
        \includegraphics[width=\textwidth]{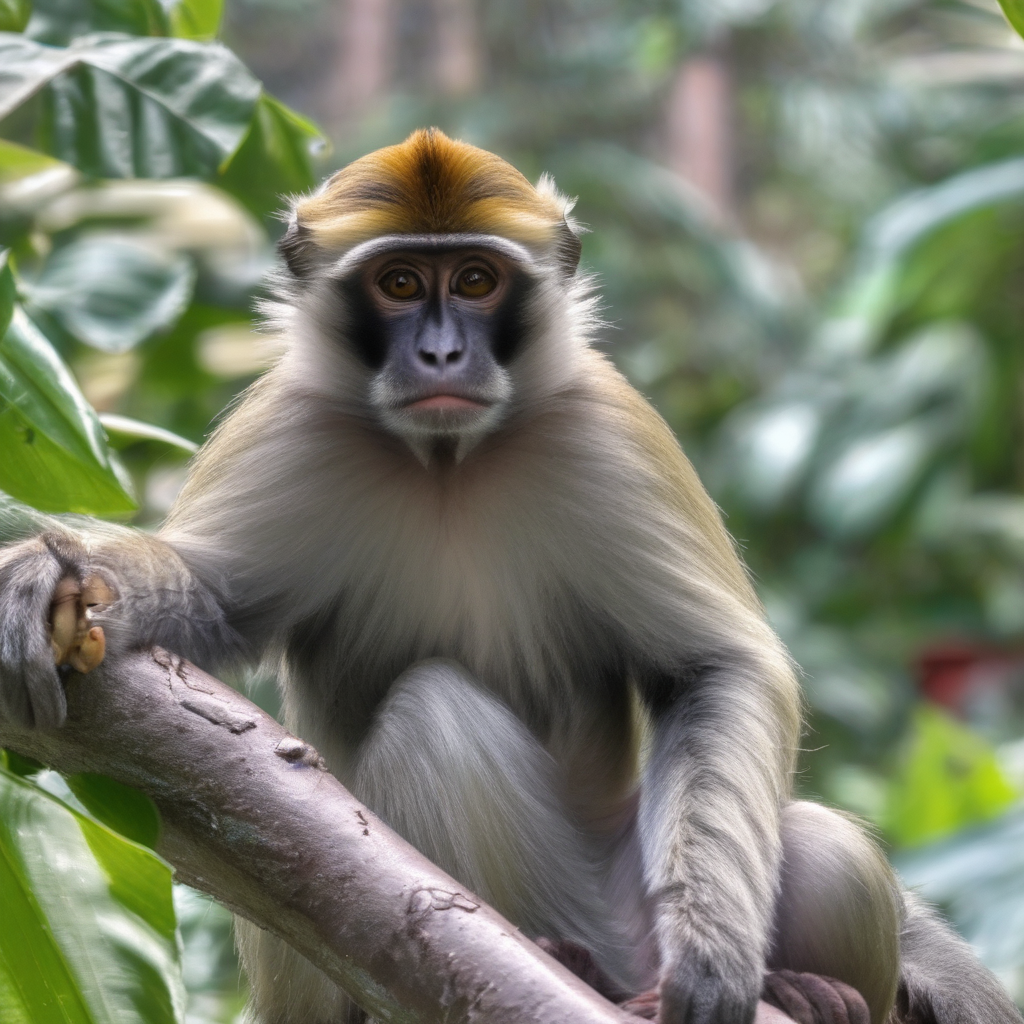}
        \vspace{-18pt}
        \caption*{Direct}
    \end{subfigure}};
    \node[anchor=north west]  at (1.35cm,-1.4cm) {
    \begin{subfigure}[b]{\locationpicsize}
        \includegraphics[width=\textwidth]{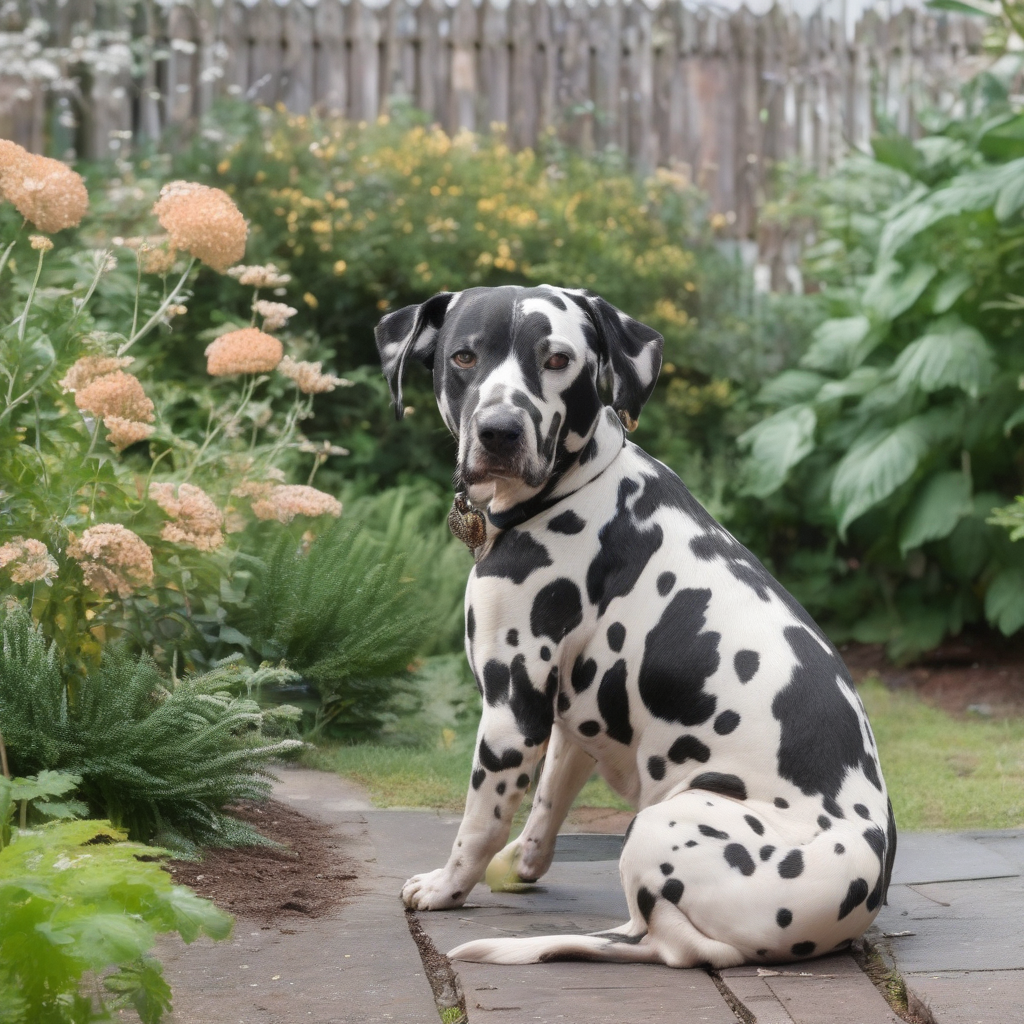}
        \vspace{-18pt}
        \caption*{Bridge}
    \end{subfigure}};
    \node[anchor=north west]  at (2.7cm,-1.4cm) {
    \begin{subfigure}[b]{\locationpicsize}
        \includegraphics[width=\textwidth]{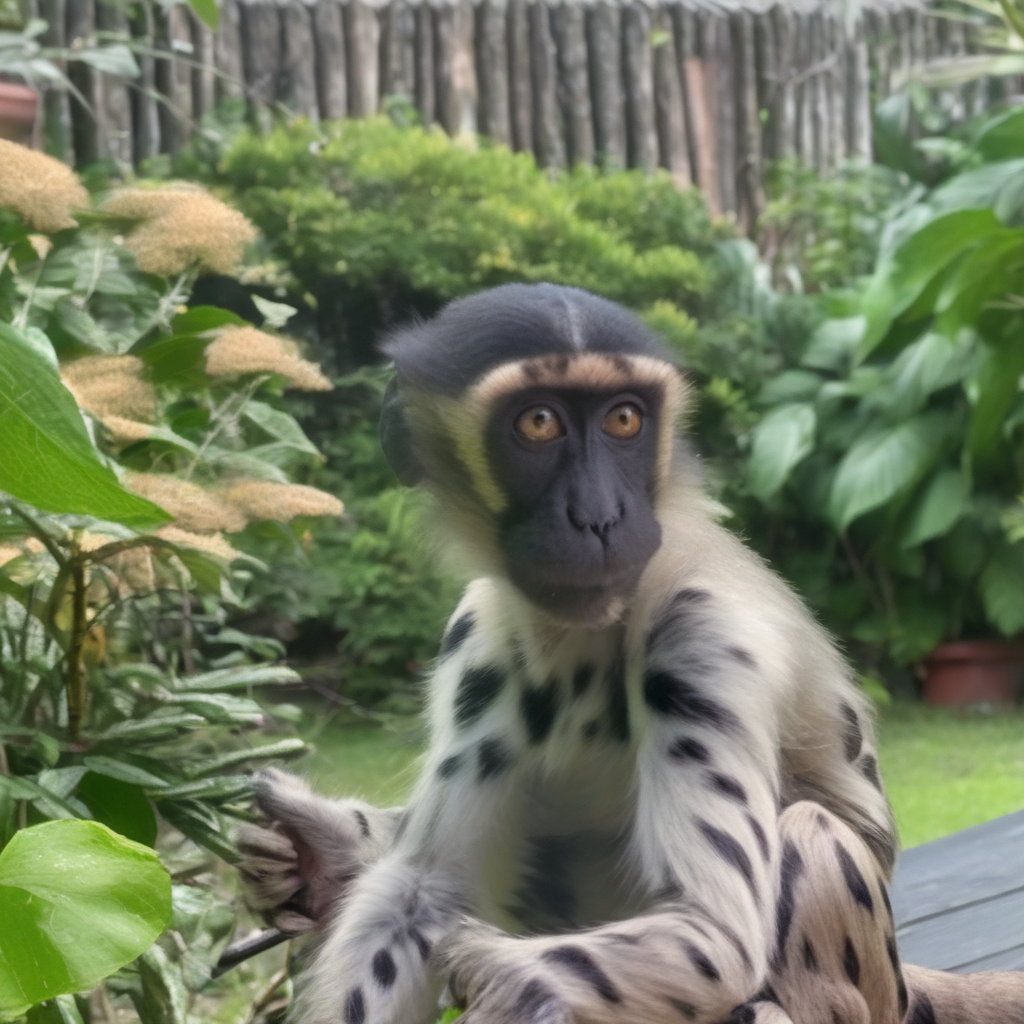}
        \vspace{-18pt}
        \caption*{P2P}
    \end{subfigure}};
    \node[anchor=north west]  at (4.05cm,-1.4cm) {
    \begin{subfigure}[b]{\locationpicsize}
        \includegraphics[width=\textwidth]{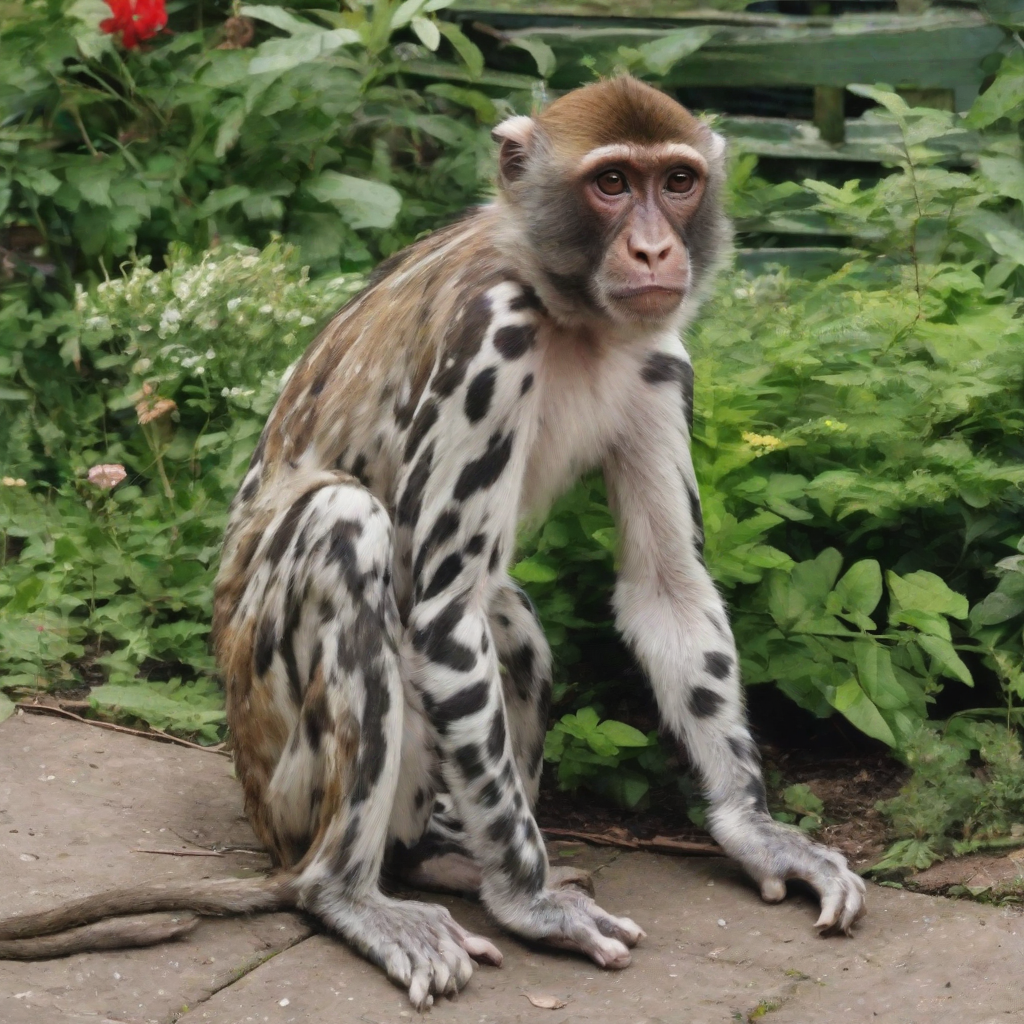}
        \vspace{-18pt}
        \caption*{Masactrl}
    \end{subfigure}};
    \node[anchor=north west]  at (5.4cm,-1.4cm) {
    \begin{subfigure}[b]{\locationpicsize}
        \includegraphics[width=\textwidth]{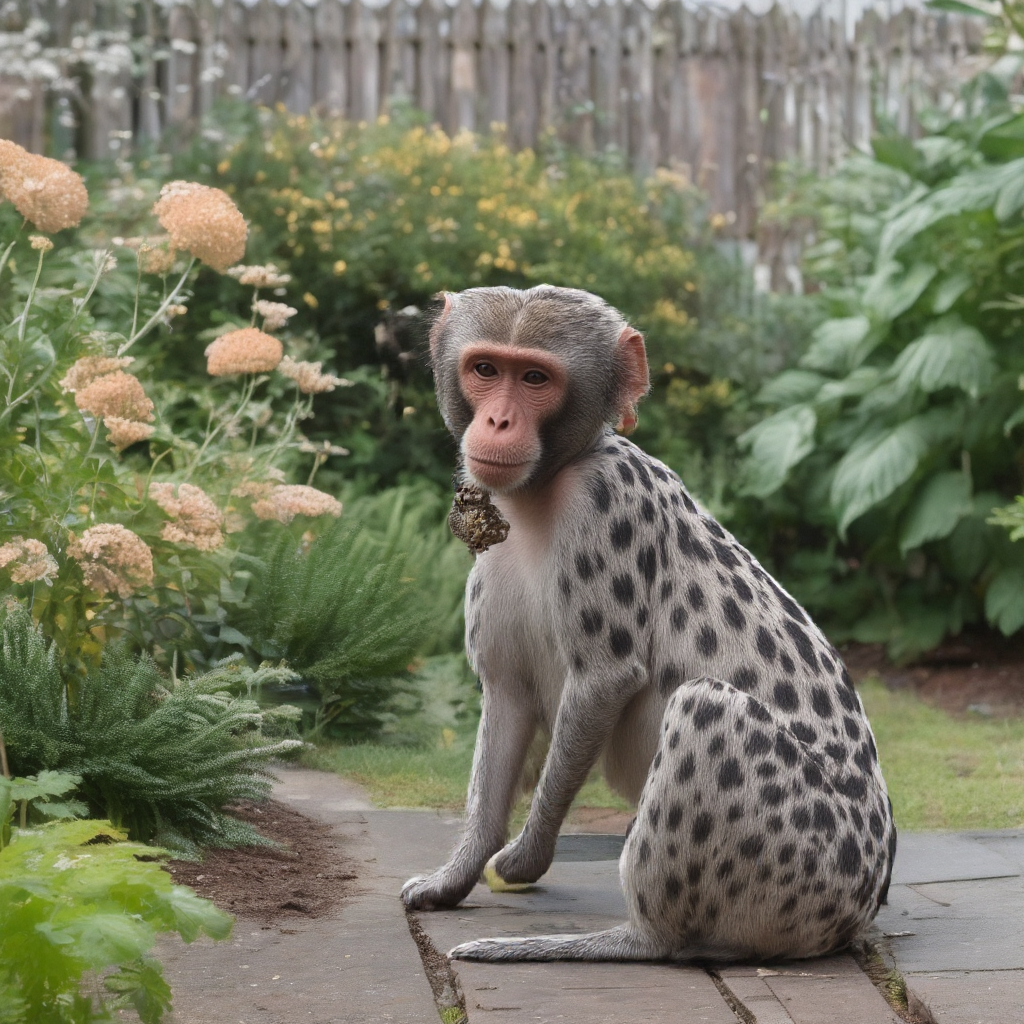}
        \vspace{-18pt}
        \caption*{FLUX.Fill}
    \end{subfigure}};
    \node[anchor=north west]  at (6.75cm,-1.4cm) {
    \begin{subfigure}[b]{\locationpicsize}
        \includegraphics[width=\textwidth]{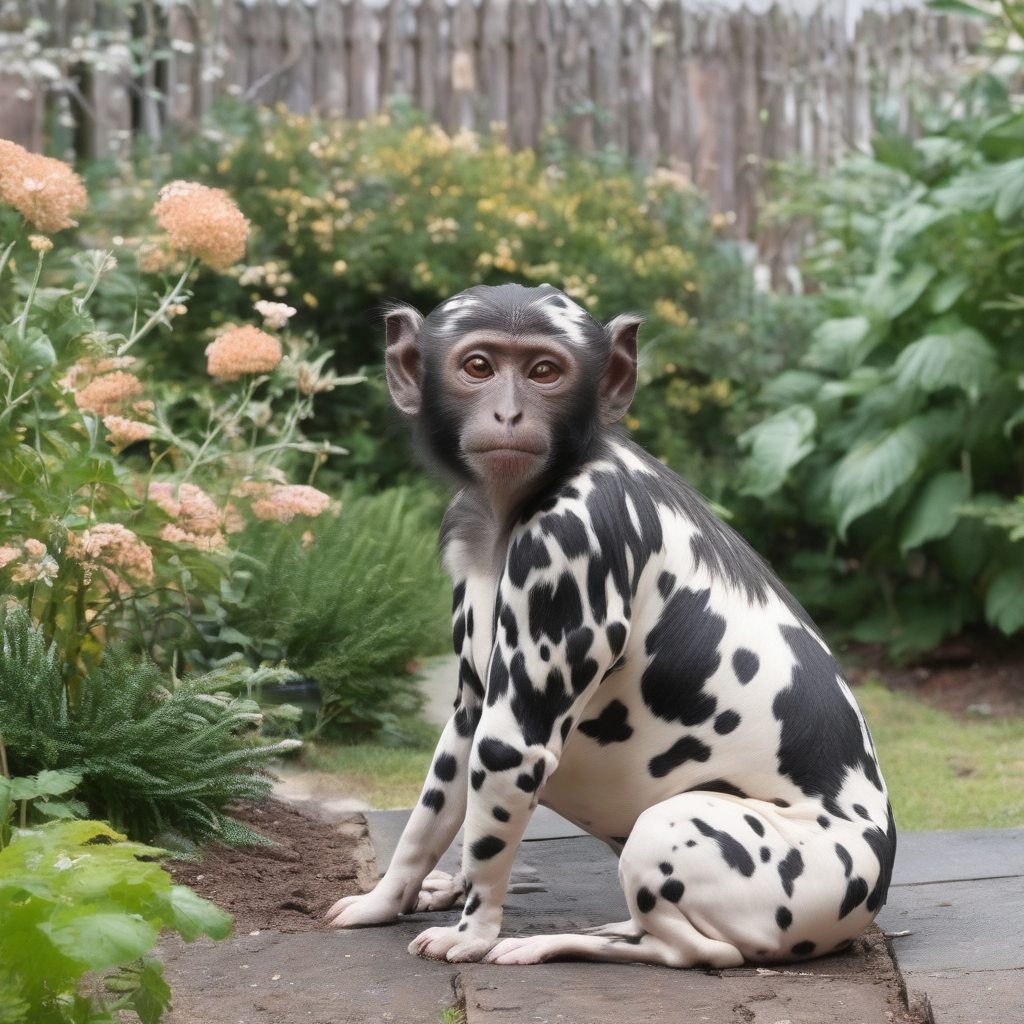}
        \vspace{-18pt}
        \caption*{Ours}
    \end{subfigure}};
    \end{tikzpicture}
    \vspace{-5pt}
    \caption{Comparison of novel content generation using other methods. The first column shows suboptimal images generated directly from text, while the second column serves as a bridge, edited to achieve the final expected result.}
    \vspace{-10pt}
    \label{fig:unseen}
\end{figure}

\begin{figure}[!t]
	\centering
	\begin{tikzpicture}
		\node[anchor=north west,align=left,text width=0.8*\replacetextwidth] at (0, 0) {\scriptsize $P$: best quality, an {\color{red}astronaut} is riding a {\color{blue} horse} in the space in a photorealistic style. \\ $P'$: *};

		\node[anchor=north west] (img0) at (1.8cm,0) {
			\begin{subfigure}[t]{\replacepicsize}
				\includegraphics[width=\textwidth]{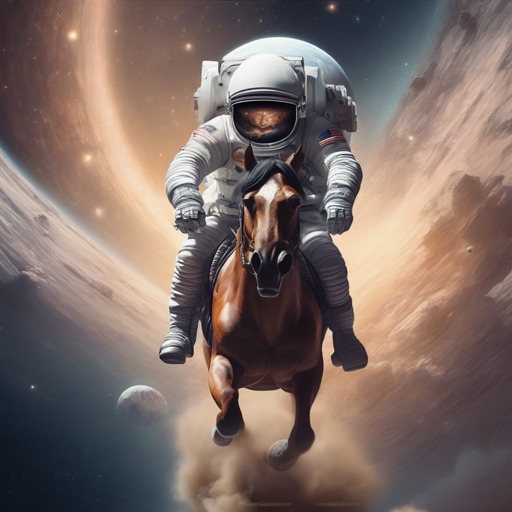}
                \vspace{-18pt}
				\caption*{\footnotesize Origin}
			\end{subfigure}
		};
        \node[anchor=north west] (img6) at (3.45cm,0) {
			\begin{subfigure}[t]{\replacepicsize}
				\includegraphics[width=\textwidth]{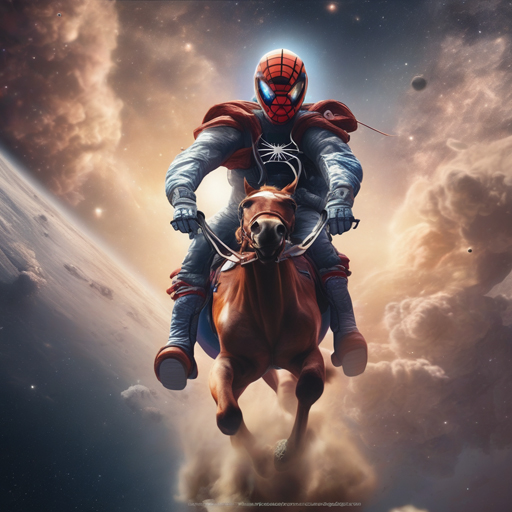}
                \vspace{-18pt}
				\caption*{\footnotesize{\color{red} Spiderman}}
			\end{subfigure}
		};
		\node[anchor=north west] (img1) at (5.1cm,0) {
			\begin{subfigure}[t]{\replacepicsize}
				\includegraphics[width=\textwidth]{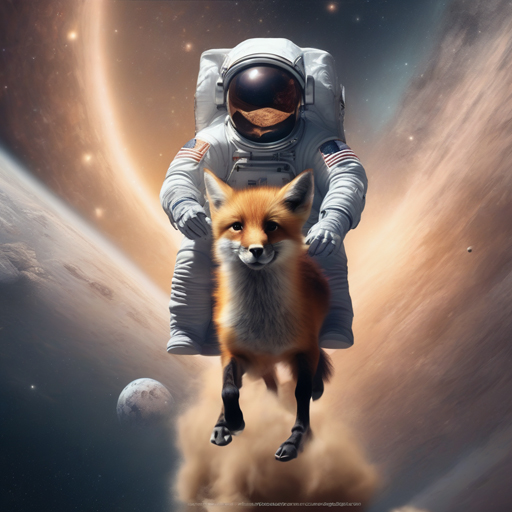}
                \vspace{-18pt}
				\caption*{\footnotesize{\color{blue} Fox}}
			\end{subfigure}
		};
		
		\node[anchor=north west] (img2) at (6.75cm,0) {
			\begin{subfigure}[t]{\replacepicsize}
				\includegraphics[width=\textwidth]{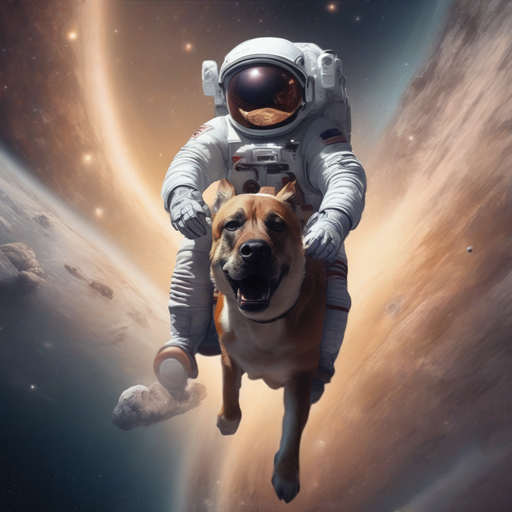}
                \vspace{-18pt}
				\caption*{\footnotesize{\color{blue} Dog}}
			\end{subfigure}
		};
		
		\node[anchor=north west,align=left,text width=0.8*\replacetextwidth] at (0, -1.9cm) {\scriptsize $P$: best quality, a {\color{red}dog} wear a {\color{blue} blue sweater} in cartoon style. \\ $P'$: *};
		\node[anchor=north west] (img7) at (1.8cm,-1.9cm) {
			\begin{subfigure}[t]{\replacepicsize}
				\includegraphics[width=\textwidth]{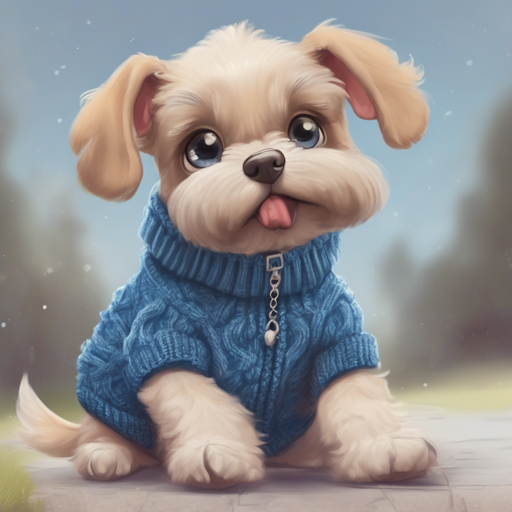}
                \vspace{-18pt}
				\caption*{\footnotesize Origin}
			\end{subfigure}
		};
            \node[anchor=north west] (img12) at (3.45cm,-1.9cm) {
			\begin{subfigure}[t]{\replacepicsize}
				\includegraphics[width=\textwidth]{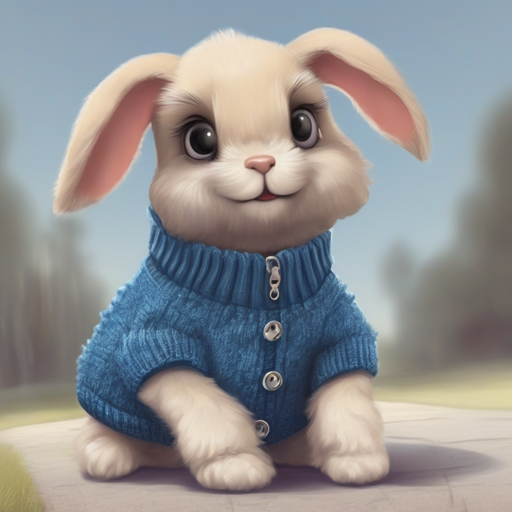}
                \vspace{-18pt}
				\caption*{\footnotesize{\color{red} Rabbit}}
			\end{subfigure}
		};
		\node[anchor=north west] (img8) at (5.1cm,-1.9cm) {
			\begin{subfigure}[t]{\replacepicsize}
				\includegraphics[width=\textwidth]{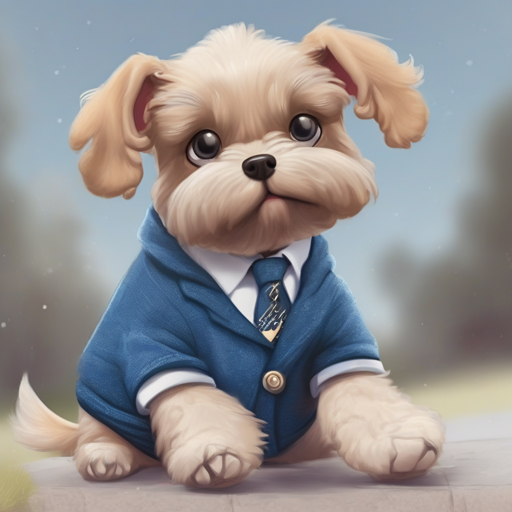}
                \vspace{-18pt}
				\caption*{\footnotesize{\color{blue} Suit}}
                
			\end{subfigure}
		};
		\node[anchor=north west] (img10) at (6.75cm,-1.9cm) {
			\begin{subfigure}[t]{\replacepicsize}
				\includegraphics[width=\textwidth]{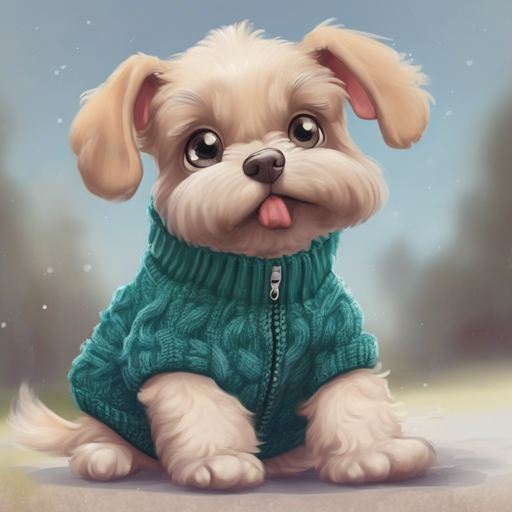}
                \vspace{-18pt}
				\caption*{\footnotesize  {\color{blue} Green sweater}}
			\end{subfigure}
		};
        
		\node[anchor=north west,align=left,text width=0.8*\replacetextwidth] at (0, -3.8cm) {\scriptsize $P$: {\color{red} happy} woman with a {\color{blue} black hair}, studio, portrait, facing camera, dark bg. \\ $P'$: * woman};
		
		\node[anchor=north west] (img14) at (1.8cm,-3.8cm) {
			\begin{subfigure}[t]{\replacepicsize}
				\includegraphics[width=\textwidth]{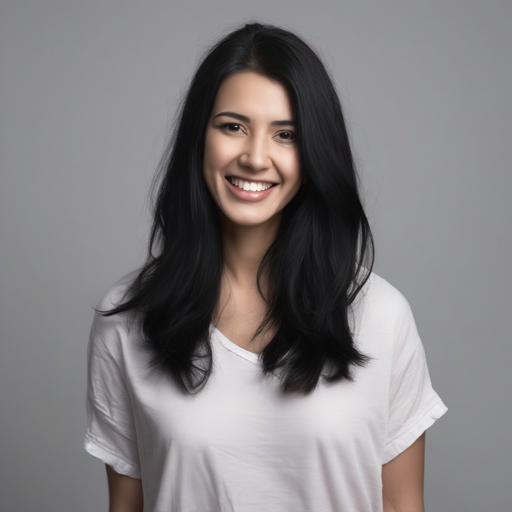}
                \vspace{-18pt}
				\caption*{\footnotesize Origin}
			\end{subfigure}
		};
		\node[anchor=north west] (img15) at (3.45cm,-3.8cm) {
			\begin{subfigure}[t]{\replacepicsize}
				\includegraphics[width=\textwidth]{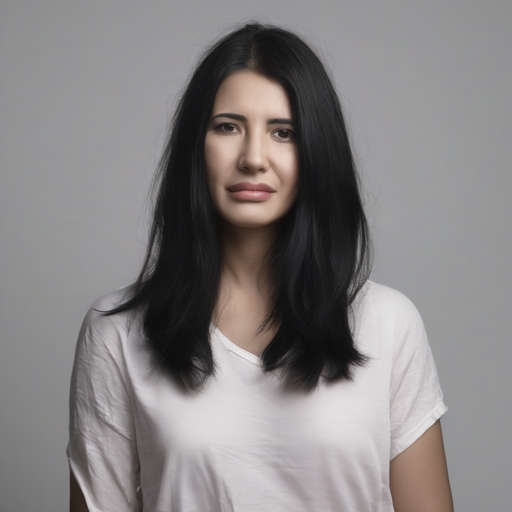}
                \vspace{-18pt}
				\caption*{\footnotesize{\color{red} Crying}}
			\end{subfigure}
		};
		\node[anchor=north west] (img19) at (5.1cm,-3.8cm) {
			\begin{subfigure}[t]{\replacepicsize}
				\includegraphics[width=\textwidth]{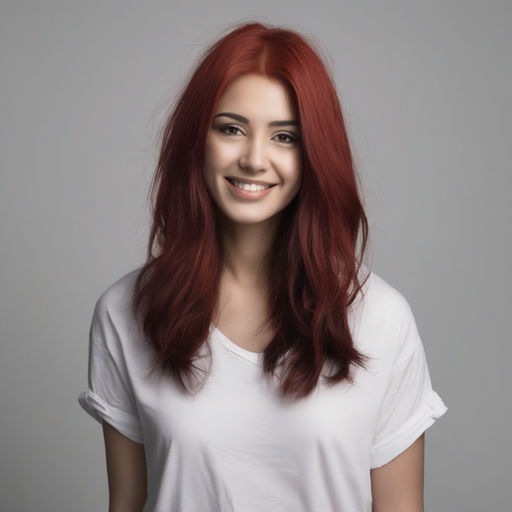}
                \vspace{-18pt}
				\caption*{\footnotesize{\color{blue} Red hair}}
			\end{subfigure}
		};
		\node[anchor=north west] (img20) at (6.75cm,-3.8cm) {
			\begin{subfigure}[t]{\replacepicsize}
				\includegraphics[width=\textwidth]{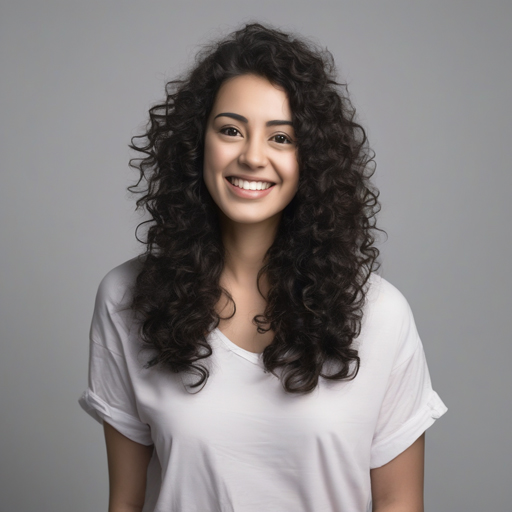}
                \vspace{-18pt}
				\caption*{\footnotesize{\color{blue} Curly hair}}
			\end{subfigure}
		};
		
		\node[anchor=north west,align=left,text width=0.8*\replacetextwidth] at (0, -5.7cm) {\scriptsize $P$: {\color{red} Kobe Bryant} wearing his iconic Lakers jersey while playing basketball, realistic, high resolution. \\ $P'$: *};
		
		\node[anchor=north west] (img21) at (1.8cm,-5.7cm) {
			\begin{subfigure}[t]{\replacepicsize}
				\includegraphics[width=\textwidth]{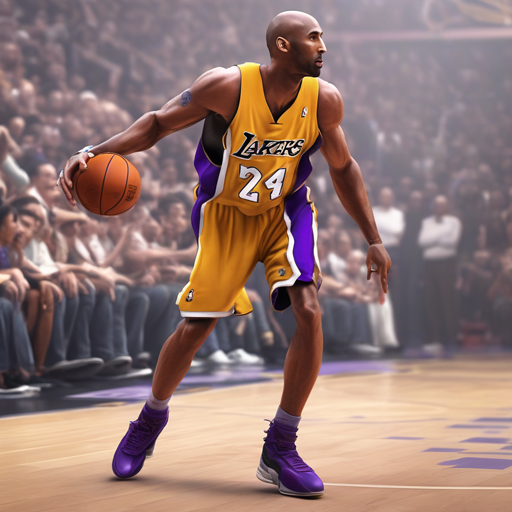}
                \vspace{-18pt}
				\caption*{\footnotesize Origin}
			\end{subfigure}
		};
		\node[anchor=north west] (img22) at (3.45cm,-5.7cm) {
			\begin{subfigure}[t]{\replacepicsize}
				\includegraphics[width=\textwidth]{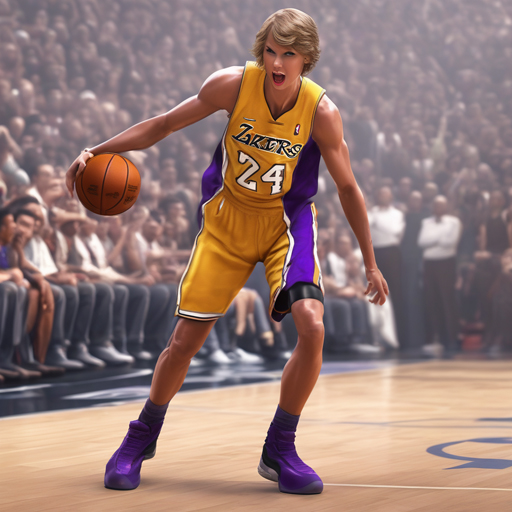}
                \vspace{-18pt}
				\caption*{\footnotesize{\color{red}Taylor Swift}}
			\end{subfigure}
		};
		\node[anchor=north west] (img25) at (5.1cm,-5.7cm) {
			\begin{subfigure}[t]{\replacepicsize}
				\includegraphics[width=\textwidth]{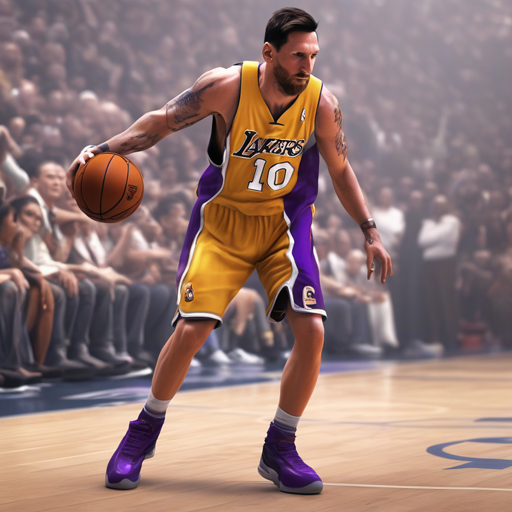}
                \vspace{-18pt}
				\caption*{\footnotesize{\color{red}Lionel Messi}}
			\end{subfigure}
		};
		\node[anchor=north west] (img26) at (6.75cm,-5.7cm) {
			\begin{subfigure}[t]{\replacepicsize}
				\includegraphics[width=\textwidth]{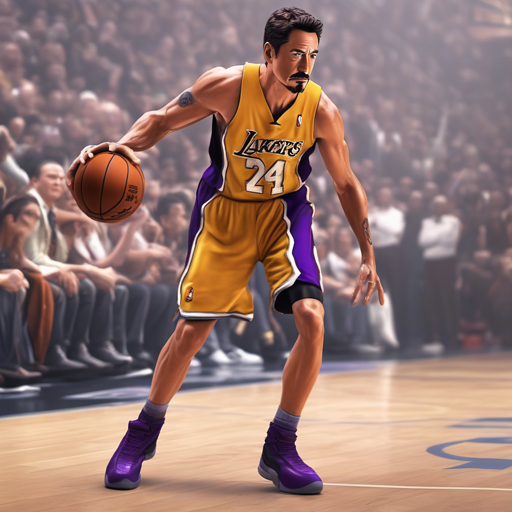}
                \vspace{-18pt}
				\caption*{\footnotesize{\color{red}Tony Stark}}
			\end{subfigure}
		};

		
	\end{tikzpicture}
    \vspace{-15pt}
	\caption{Example of object replacement. ``*'' in $P^{'}$ denotes the word below each image. Words of the same color indicate corresponding object changes. Zoom in for a better view.}
	\label{fig:replace}
    \vspace{-5pt}
\end{figure}

\noindent \textbf{Local Object Replacement}. Localized object editing focuses on replacing objects in specific regions, with comparative results across methods illustrated in Fig.~\ref{fig:location}. For prompts containing multiple instances of the same object (e.g., ``two dogs''), P2P and Masactrl inherently replace all instances due to their limitations, making them unsuitable for localized object editing tasks. In contrast, our method leverages Softbox to precisely edit targeted regions while preserving identical objects in other areas.

\noindent \textbf{Novel Image Generation}. As shown in Fig.~\ref{fig:unseen}, word-level replacement often fails to produce desired images (first column). Satisfactory results typically require extensive random seed experimentation, being time-consuming and inefficient. In contrast, our editing approach uses intermediate images as bridges to generate novel content. Other methods remain constrained by target prompt modifications without overcoming inherent biases (e.g., the second row shows nearly identical spotted monkeys). Our method seamlessly blends features like spotted dogs and monkeys for novel compositions. Moreover, even with precise prompts, pre-trained models' limitations hinder satisfactory outcomes for unknown or complex scenes. Our approach provides an elegant solution that effectively reduces novel content generation difficulty (see Appendix A.4).
\begin{figure}[!t]
    \centering
    \begin{tikzpicture}
    \node[anchor=north west]  at (0,0) {
    \begin{subfigure}[t]{\addpicsize}
        \includegraphics[width=\textwidth]{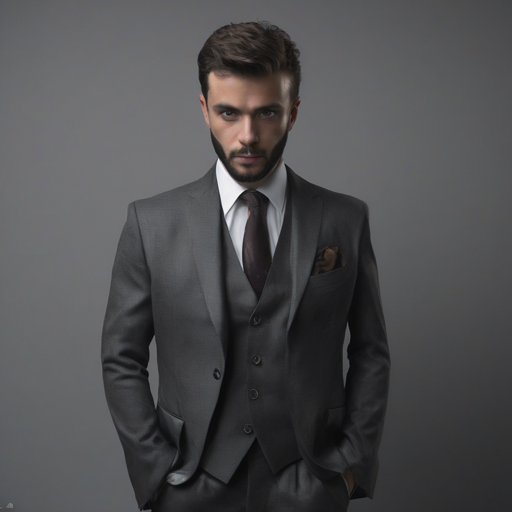}
    \end{subfigure}
    };
    \node[anchor=north west]  at (1.35cm,0) {
    \begin{subfigure}[t]{\addpicsize}
        \includegraphics[width=\textwidth]{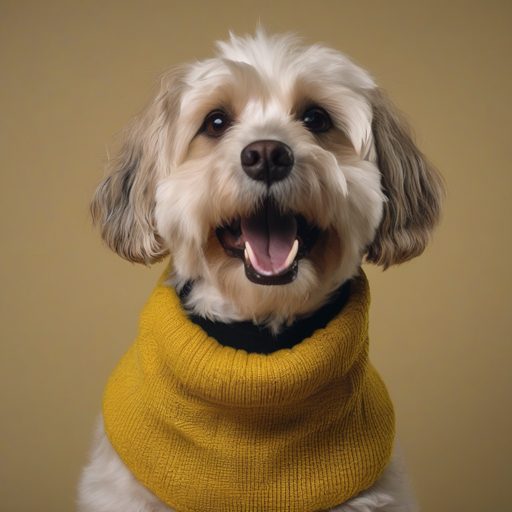}
    \end{subfigure}
    };
    \node[anchor=north west]  at (2.7cm,0) {
    \begin{subfigure}[t]{\addpicsize}
        \includegraphics[width=\textwidth]{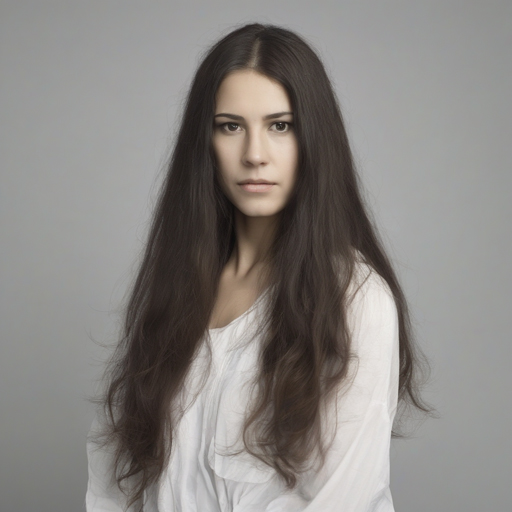}
    \end{subfigure}
    };
    \node[anchor=north west]  at (4.05cm,0) {
    \begin{subfigure}[t]{\addpicsize}
        \includegraphics[width=\textwidth]{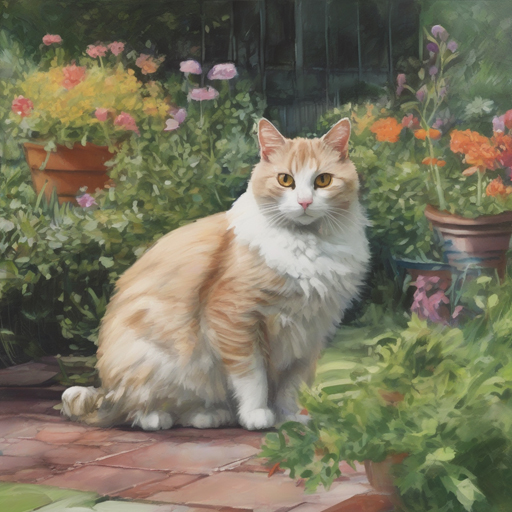}
    \end{subfigure}
    };
    \node[anchor=north west]  at (5.4cm,0) {
    \begin{subfigure}[t]{\addpicsize}
        \includegraphics[width=\textwidth]{sec/image/add/originD-2.jpg}
    \end{subfigure}
    };
    \node[anchor=north west]  at (6.75cm,0) {
    \begin{subfigure}[t]{\addpicsize}
        \includegraphics[width=\textwidth]{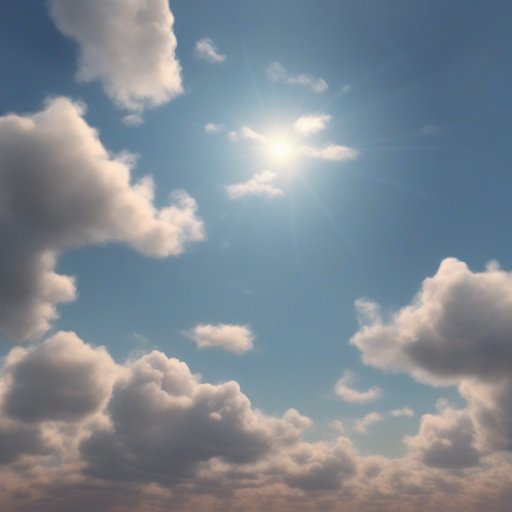}
    \end{subfigure}
    };

    \node[anchor=north west]  at (0cm,-1.4cm) {
    \begin{subfigure}[t]{\addpicsize}
        \includegraphics[width=\textwidth]{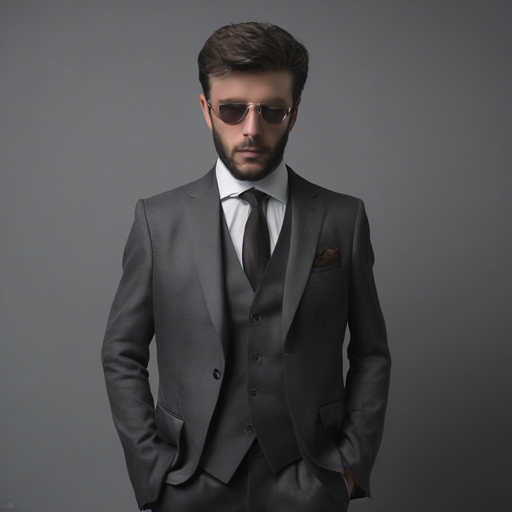}
        \vspace{-15pt}
        \captionsetup{font=addfont,justification=centering, singlelinecheck=false}
        \caption*{Sunglasses}
    \end{subfigure}
    };
    \node[anchor=north west]  at (1.35cm,-1.4cm) {
    \begin{subfigure}[t]{\addpicsize}
        \includegraphics[width=\textwidth]{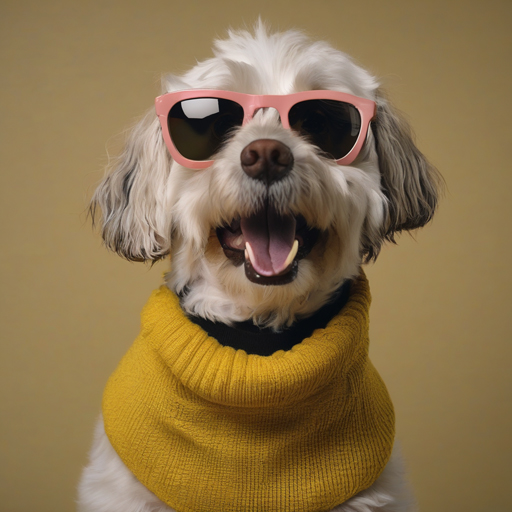}
        \vspace{-15pt}
        \captionsetup{font=addfont,justification=centering, singlelinecheck=false}
        \caption*{Sunglasses}
    \end{subfigure}
    };
    \node[anchor=north west]  at (2.7cm,-1.4cm) {
    \begin{subfigure}[t]{\addpicsize}
        \includegraphics[width=\textwidth]{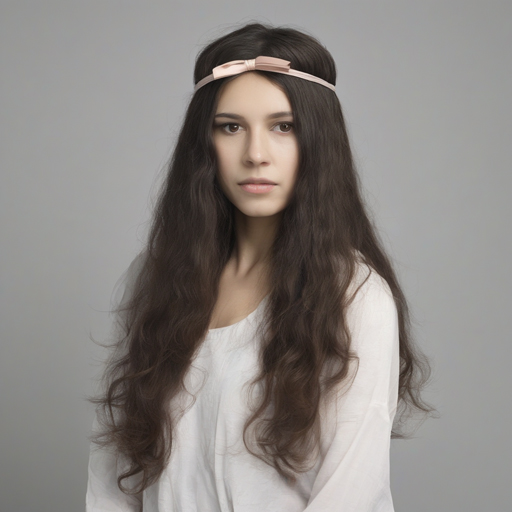}
        \vspace{-15pt}
        \captionsetup{font=addfont,justification=centering, singlelinecheck=false}
        \caption*{Bow hairband}
    \end{subfigure}
    };
    \node[anchor=north west]  at (4.05cm,-1.4cm) {
    \begin{subfigure}[t]{\addpicsize}
        \includegraphics[width=\textwidth]{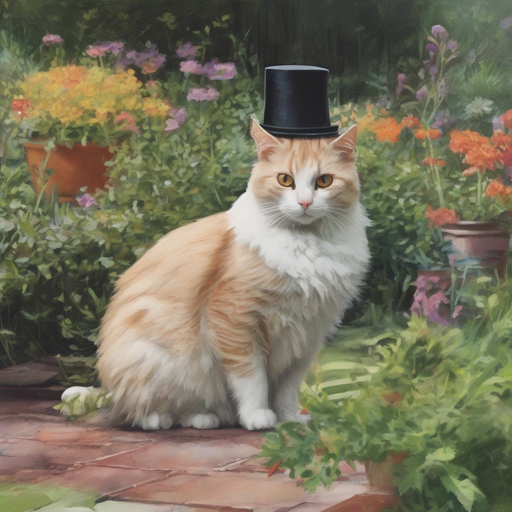}
        \vspace{-15pt}
        \captionsetup{font=addfont,justification=centering, singlelinecheck=false}
        \caption*{Magic Hat}
    \end{subfigure}
    };
    \node[anchor=north west]  at (5.4cm,-1.4cm) {
    \begin{subfigure}[t]{\addpicsize}
        \includegraphics[width=\textwidth]{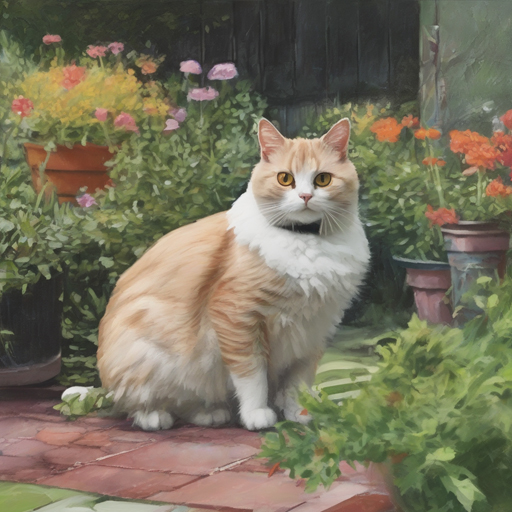}
        \vspace{-15pt}
        \captionsetup{font=addfont,justification=centering, singlelinecheck=false}
        \caption*{Cat collar}
    \end{subfigure}
    };
    \node[anchor=north west]  at (6.75cm,-1.4cm) {
    \begin{subfigure}[t]{\addpicsize}
        \includegraphics[width=\textwidth]{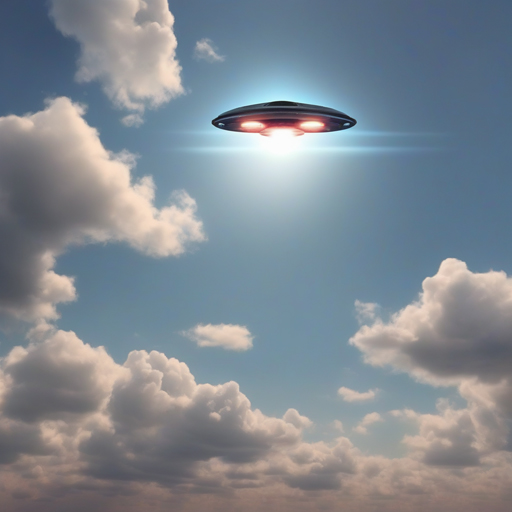}
        \vspace{-15pt}
        \captionsetup{font=addfont,justification=centering, singlelinecheck=false}
        \caption*{UFO}
    \end{subfigure}
    };
    \end{tikzpicture}
    \vspace{-5pt}
    \caption{Example of object addition. Top row: original images; bottom row: edited images.}
    \label{fig:add}
    \vspace{-10pt}
\end{figure}

\begin{figure}[!t]
	\centering
	\begin{subfigure}[t]{\stylepicsize}
		\includegraphics[width=\textwidth]{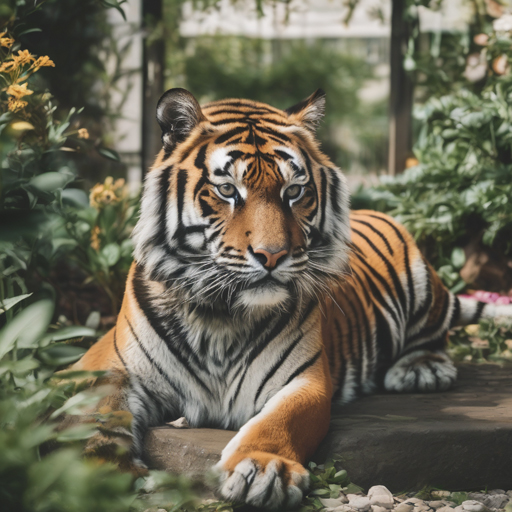}
        \vspace{-16pt}
		\captionsetup{font=stylefont,justification=centering, singlelinecheck=false}
        \caption*{origin}
	\end{subfigure}
	\begin{subfigure}[t]{\stylepicsize}
		\includegraphics[width=\textwidth]{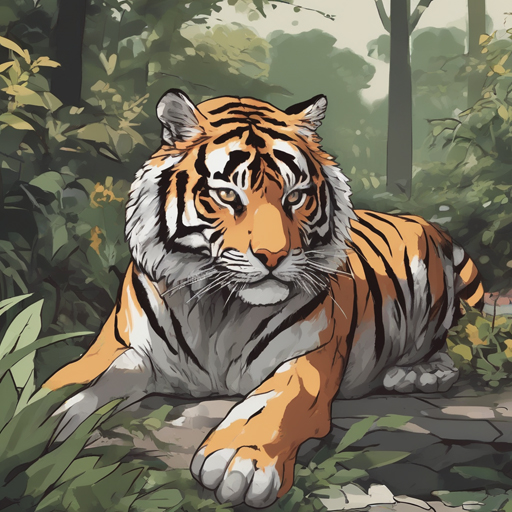}
        \vspace{-16pt}
		\captionsetup{font=stylefont,justification=centering, singlelinecheck=false}
		\caption*{in anime artwork style}
	\end{subfigure}
	\begin{subfigure}[t]{\stylepicsize}
		\includegraphics[width=\textwidth]{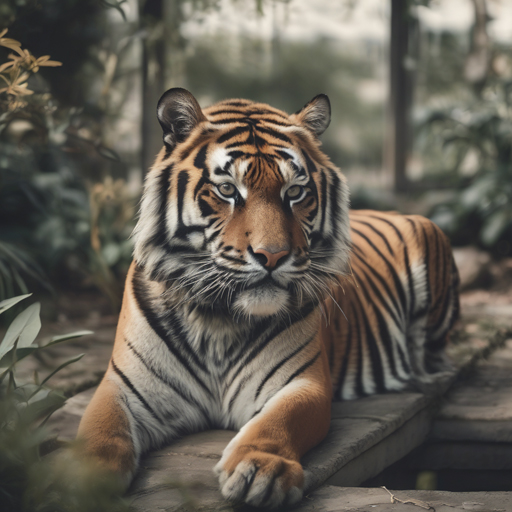}
        \vspace{-16pt}
		\captionsetup{font=stylefont,justification=centering, singlelinecheck=false}
		\caption*{in cinematic film style}
	\end{subfigure}
	\begin{subfigure}[t]{\stylepicsize}
		\includegraphics[width=\textwidth]{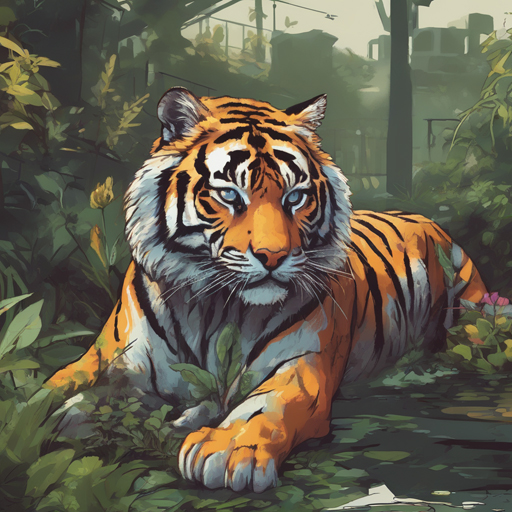}
        \vspace{-16pt}
		\captionsetup{font=stylefont,justification=centering, singlelinecheck=false}
		\caption*{in neonpunk style}
	\end{subfigure}
	\begin{subfigure}[t]{\stylepicsize}
		\includegraphics[width=\textwidth]{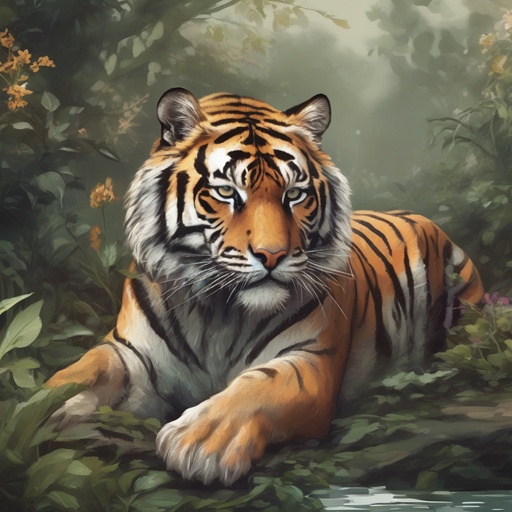}
        \vspace{-16pt}
		\captionsetup{font=stylefont,justification=centering, singlelinecheck=false}
		\caption*{in fantasy art style}
	\end{subfigure}
	\begin{subfigure}[t]{\stylepicsize}
		\includegraphics[width=\textwidth]{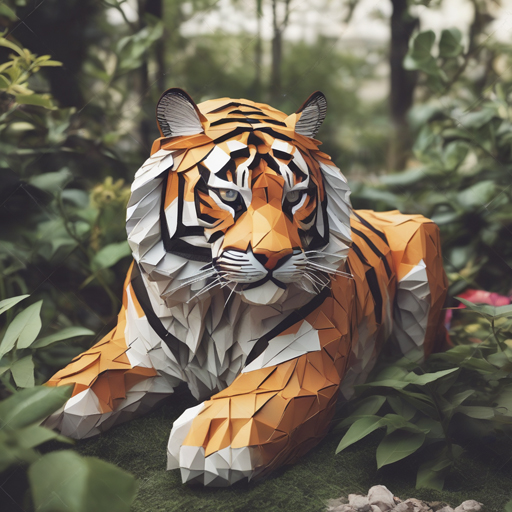}
        \vspace{-16pt}
		\captionsetup{font=stylefont,justification=centering, singlelinecheck=false}
		\caption*{in origami style}
	\end{subfigure}
    
	\begin{subfigure}[t]{\stylepicsize}
		\includegraphics[width=\textwidth]{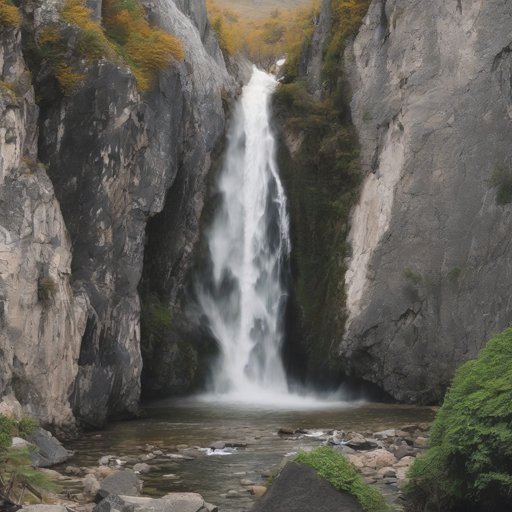}
        \vspace{-16pt}
		\captionsetup{font=stylefont,justification=centering, singlelinecheck=false}
        \caption*{origin}
	\end{subfigure}
	\begin{subfigure}[t]{\stylepicsize}
		\includegraphics[width=\textwidth]{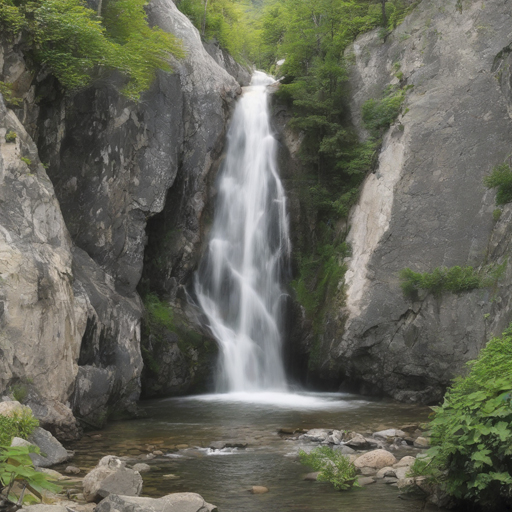}
        \vspace{-16pt}
		\captionsetup{font=stylefont,justification=centering, singlelinecheck=false}
		\caption*{in summer}
	\end{subfigure}
	\begin{subfigure}[t]{\stylepicsize}
		\includegraphics[width=\textwidth]{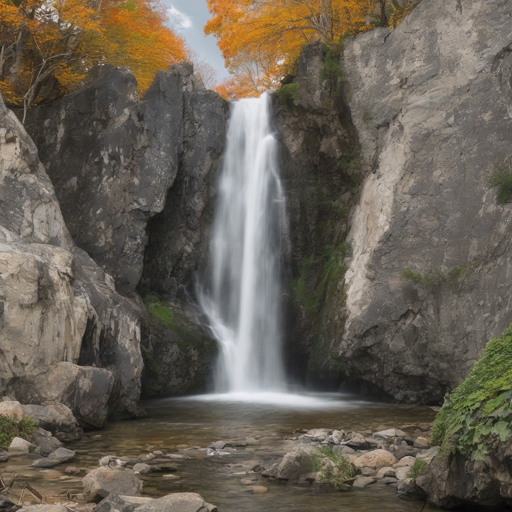}
        \vspace{-16pt}
		\captionsetup{font=stylefont,justification=centering, singlelinecheck=false}
		\caption*{in fall}
	\end{subfigure}
	\begin{subfigure}[t]{\stylepicsize}
		\includegraphics[width=\textwidth]{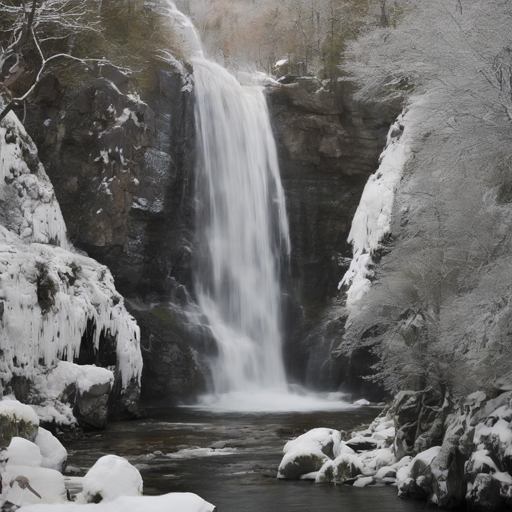}
        \vspace{-16pt}
		\captionsetup{font=stylefont,justification=centering, singlelinecheck=false}
		\caption*{in winter}
	\end{subfigure}
	\begin{subfigure}[t]{\stylepicsize}
		\includegraphics[width=\textwidth]{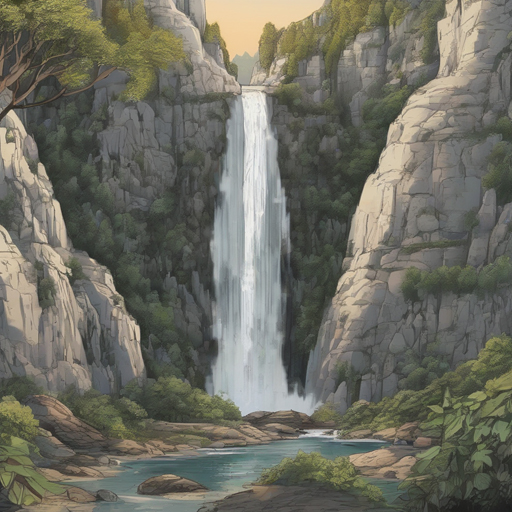}
        \vspace{-16pt}
		\captionsetup{font=stylefont,justification=centering, singlelinecheck=false}
		\caption*{in cartoon style}
	\end{subfigure}
	\begin{subfigure}[t]{\stylepicsize}
		\includegraphics[width=\textwidth]{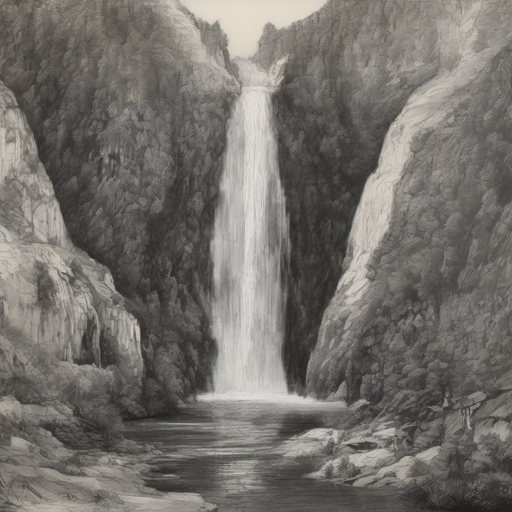}
        \vspace{-16pt}
		\captionsetup{font=stylefont,justification=centering, singlelinecheck=false}
		\caption*{in sketch}
	\end{subfigure}
	\begin{subfigure}[t]{\stylepicsize}
		\includegraphics[width=\textwidth]{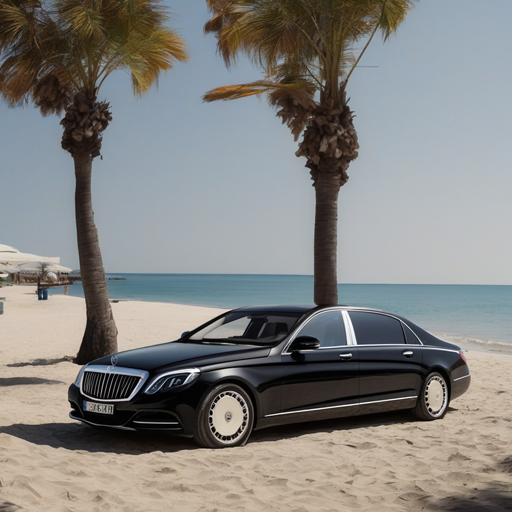}
        \vspace{-16pt}
		\captionsetup{font=stylefont,justification=centering, singlelinecheck=false}
        \caption*{origin}
	\end{subfigure}
	\begin{subfigure}[t]{\stylepicsize}
		\includegraphics[width=\textwidth]{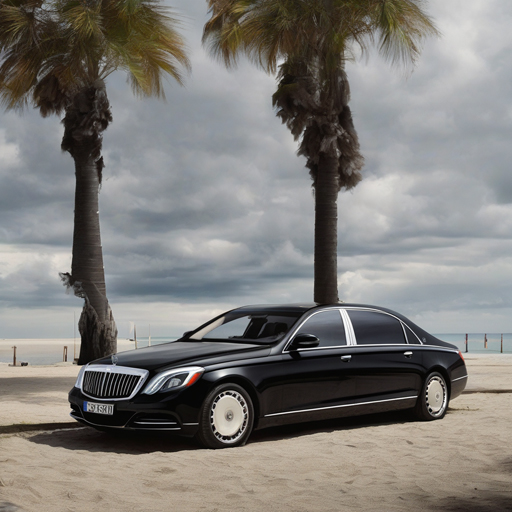}
        \vspace{-16pt}
		\captionsetup{font=stylefont,justification=centering, singlelinecheck=false}
		\caption*{in a cloudy day}
	\end{subfigure}
	\begin{subfigure}[t]{\stylepicsize}
		\includegraphics[width=\textwidth]{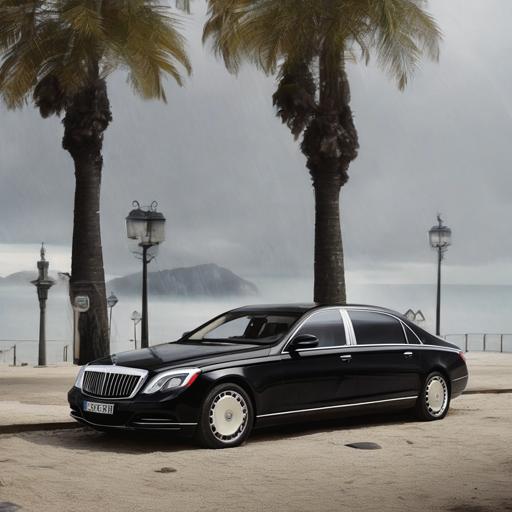}
        \vspace{-16pt}
		\captionsetup{font=stylefont,justification=centering, singlelinecheck=false}
		\caption*{in a rainy day}
	\end{subfigure}
	\begin{subfigure}[t]{\stylepicsize}
		\includegraphics[width=\textwidth]{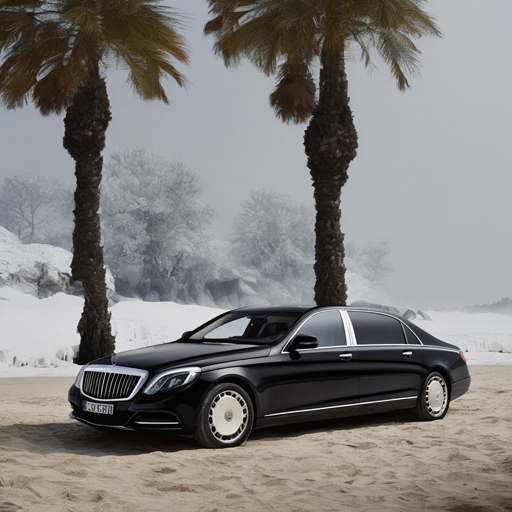}
        \vspace{-16pt}
		\captionsetup{font=stylefont,justification=centering, singlelinecheck=false}
		\caption*{in a snowy day}
	\end{subfigure}
	\begin{subfigure}[t]{\stylepicsize}
		\includegraphics[width=\textwidth]{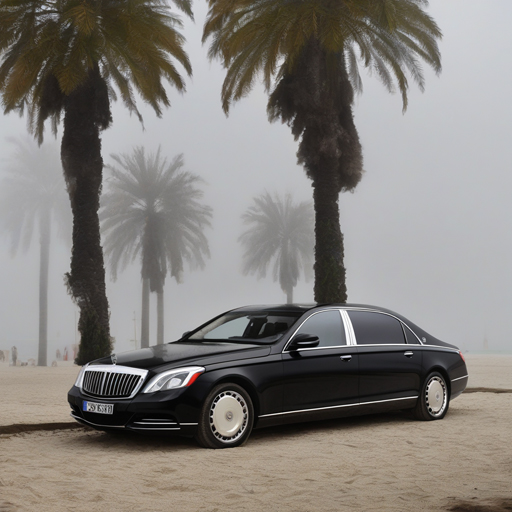}
        \vspace{-16pt}
		\captionsetup{font=stylefont,justification=centering, singlelinecheck=false}
		\caption*{in a foggy day}
	\end{subfigure}
	\begin{subfigure}[t]{\stylepicsize}
		\includegraphics[width=\textwidth]{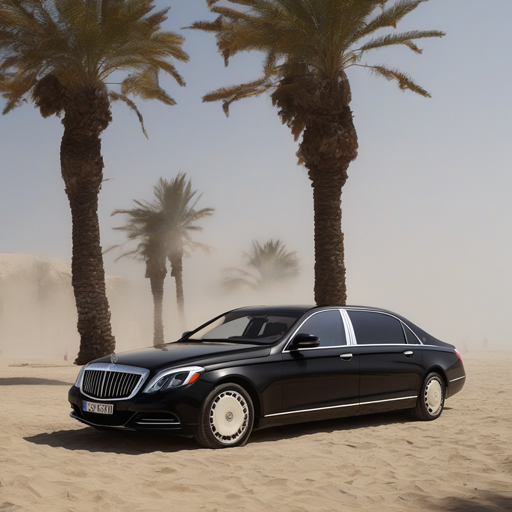}
        \vspace{-16pt}
		\captionsetup{font=stylefont,justification=centering, singlelinecheck=false}
		\caption*{in a sandstorm day}
	\end{subfigure}
    \vspace{-10pt}
	\caption{Examples of style transfer.}
	\label{fig:style}
    \vspace{-10pt}
\end{figure}
\begin{figure}[!t]
    \centering
    \begin{subfigure}[t]{\realpicsize}
        \includegraphics[width=\textwidth]{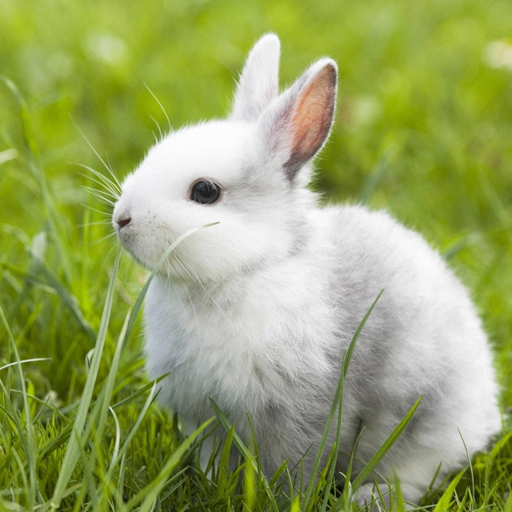}
        \captionsetup{font=realfont,justification=centering, singlelinecheck=false}
        \caption*{Rabbit}
    \end{subfigure}
    \begin{subfigure}[t]{\realpicsize}
        \includegraphics[width=\textwidth]{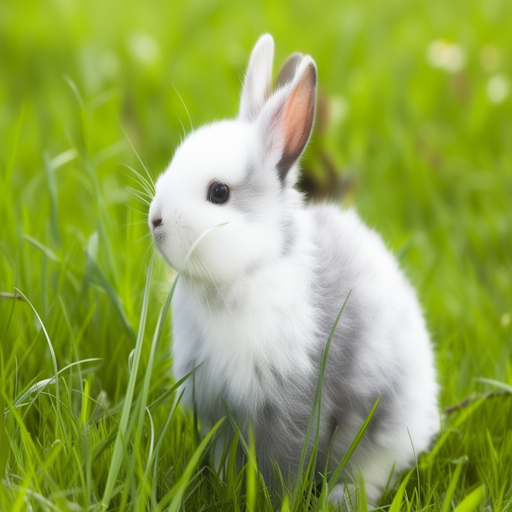}
        \captionsetup{font=realfont,justification=centering, singlelinecheck=false}
        \caption*{Reconstructed}
    \end{subfigure}
    \begin{subfigure}[t]{\realpicsize}
        \includegraphics[width=\textwidth]{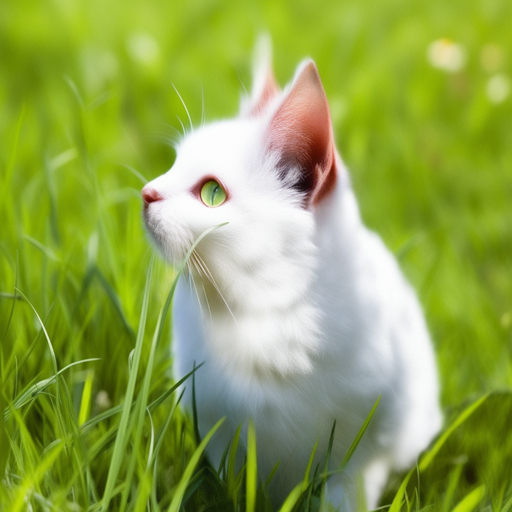}
        \captionsetup{font=realfont,justification=centering, singlelinecheck=false}
        \caption*{Cat}
    \end{subfigure}
    \begin{subfigure}[t]{\realpicsize}
        \includegraphics[width=\textwidth]{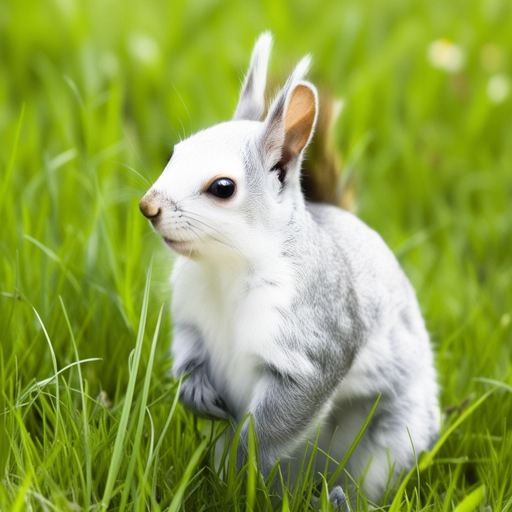}
        \captionsetup{font=realfont,justification=centering, singlelinecheck=false}
        \caption*{Squirrel}
    \end{subfigure}
    \begin{subfigure}[t]{\realpicsize}
        \includegraphics[width=\textwidth]{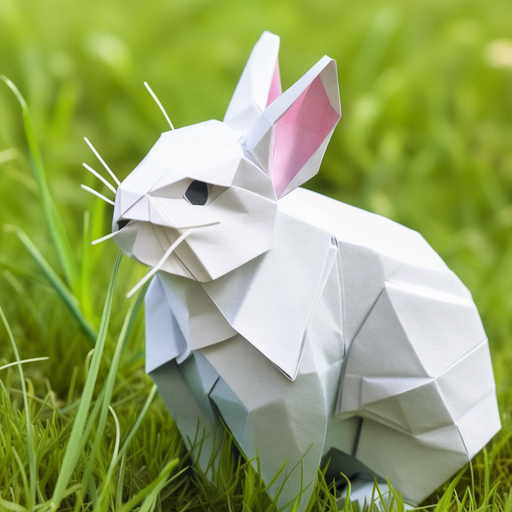}
        \captionsetup{font=realfont,justification=centering, singlelinecheck=false}
        \caption*{in origami style}
    \end{subfigure}
    \begin{subfigure}[t]{\realpicsize}
        \includegraphics[width=\textwidth]{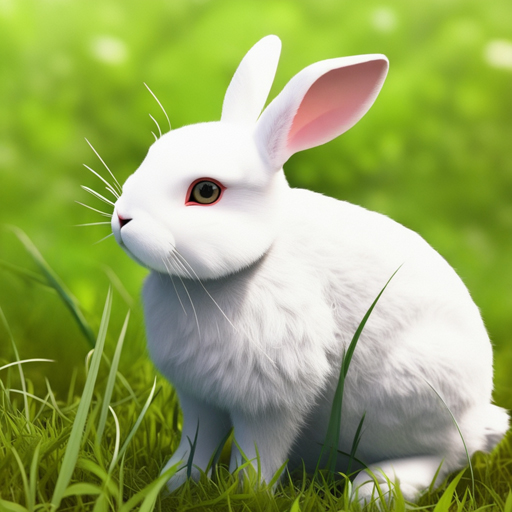}
        \captionsetup{font=realfont,justification=centering, singlelinecheck=false}
        \caption*{anime artwork, anime style}
    \end{subfigure}
    \begin{subfigure}[t]{\realpicsize}
        \includegraphics[width=\textwidth]{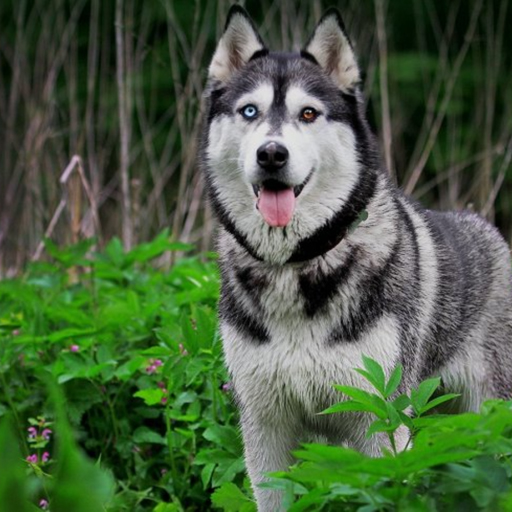}
        \captionsetup{font=realfont,justification=centering, singlelinecheck=false}
        \caption*{Dog}
    \end{subfigure}
    \begin{subfigure}[t]{\realpicsize}
        \includegraphics[width=\textwidth]{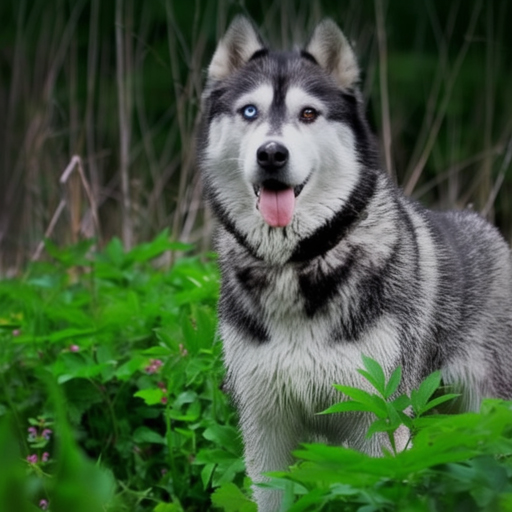}
        \captionsetup{font=realfont,justification=centering, singlelinecheck=false}
        \caption*{Reconstructed}
    \end{subfigure}
    \begin{subfigure}[t]{\realpicsize}
        \includegraphics[width=\textwidth]{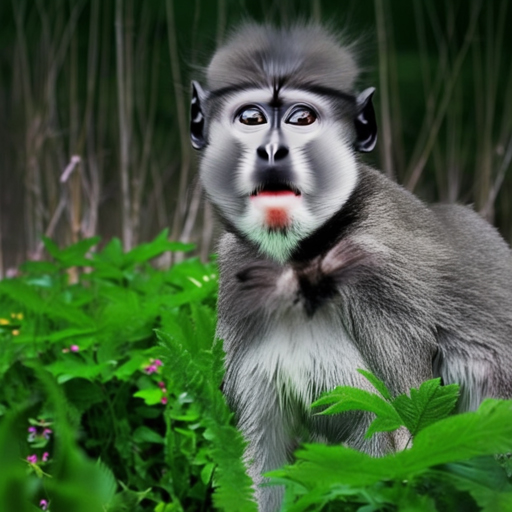}
        \captionsetup{font=realfont,justification=centering, singlelinecheck=false}
        \caption*{Monkey}
    \end{subfigure}
    \begin{subfigure}[t]{\realpicsize}
        \includegraphics[width=\textwidth]{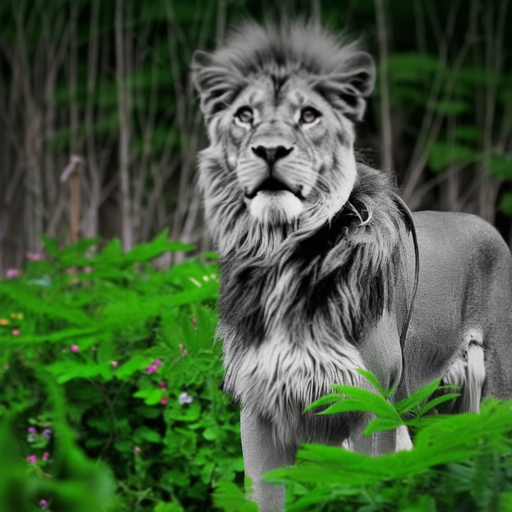}
        \captionsetup{font=realfont,justification=centering, singlelinecheck=false}
        \caption*{Lion}
    \end{subfigure}
    \begin{subfigure}[t]{\realpicsize}
        \includegraphics[width=\textwidth]{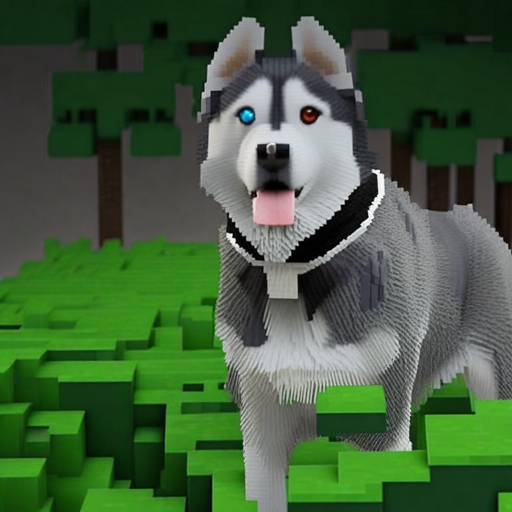}
        \captionsetup{font=realfont,justification=centering, singlelinecheck=false}
        \caption*{in Minecraft style}
    \end{subfigure}
    \begin{subfigure}[t]{\realpicsize}
        \includegraphics[width=\textwidth]{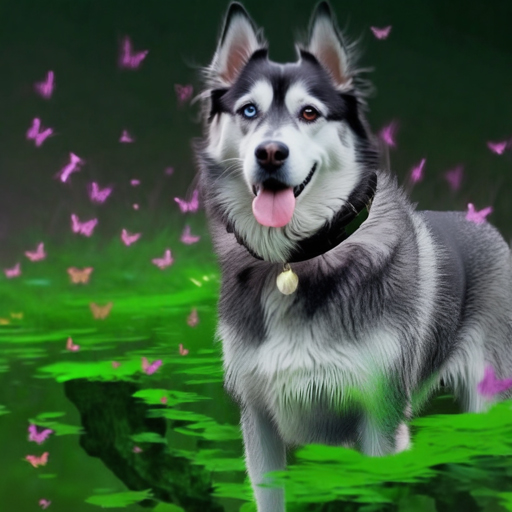}
        \captionsetup{font=realfont,justification=centering, singlelinecheck=false}
        \caption*{ethereal fantasy concept art}
    \end{subfigure}
    \vspace{-10pt}
    \caption{Examples of real image editing.}
    \label{fig:real}
    \vspace{-10pt}
\end{figure}

\begin{figure}[!t]
	\centering
	\begin{tikzpicture}
		\node[anchor=north west] (img0) at (0*\ablationfiggap,0) {
			\begin{subfigure}[t]{\ablationpicsize}
				\includegraphics[width=\textwidth]{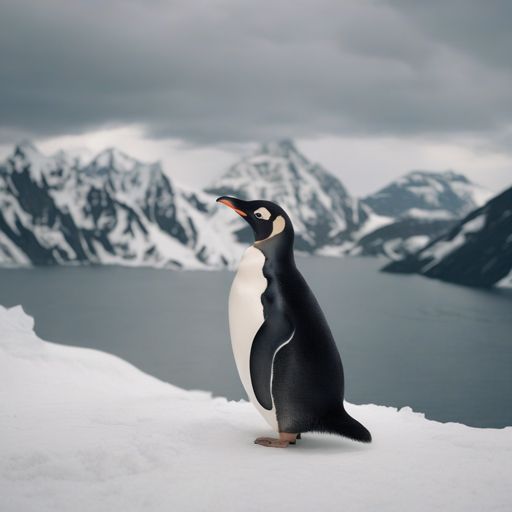}
                \vspace{-16pt}
				\caption*{Origin}
			\end{subfigure}
		};
		\node [anchor=north west, text width=3*\ablationfiggap+\ablationpicsize] at (\ablationfiggap, 0) {$P$: A penguin on the snow mountain. \\ $P'$: Bear.};
		\node[anchor=north west] (img2) at (5*\ablationfiggap,0) {
			\begin{subfigure}[t]{\ablationpicsize}
				\includegraphics[width=\textwidth]{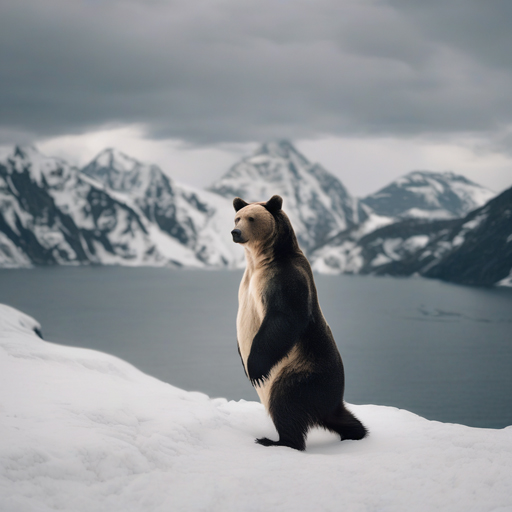}
                \vspace{-16pt}
				\caption*{Final}
			\end{subfigure}
		};
		\node[anchor=north west] (img11) at (0*\ablationfiggap,-\ablationhgap+0.2*\ablationhgap) {
			\begin{subfigure}[t]{\ablationpicsize}
				\includegraphics[width=\textwidth]{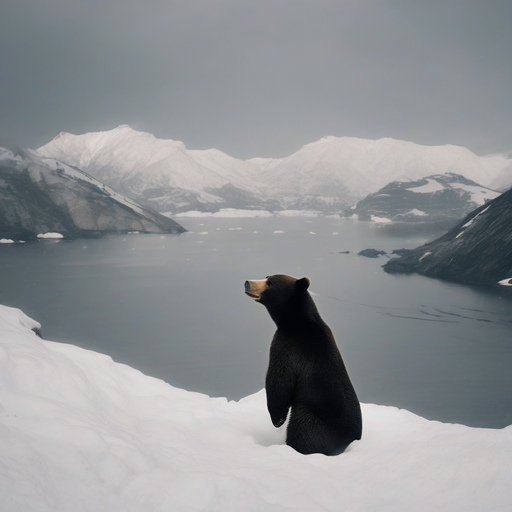}
				\captionsetup{font=ablationfont,justification=raggedright, singlelinecheck=false}
				\caption*{W/O Softbox \\ T:[0,30] \\ Box:[0.0,0.0,0.0,0.0]}
			\end{subfigure}
		};
		\node[anchor=north west] (img12) at (1*\ablationfiggap,-\ablationhgap+0.2*\ablationhgap) {
			\begin{subfigure}[t]{\ablationpicsize}
				\includegraphics[width=\textwidth]{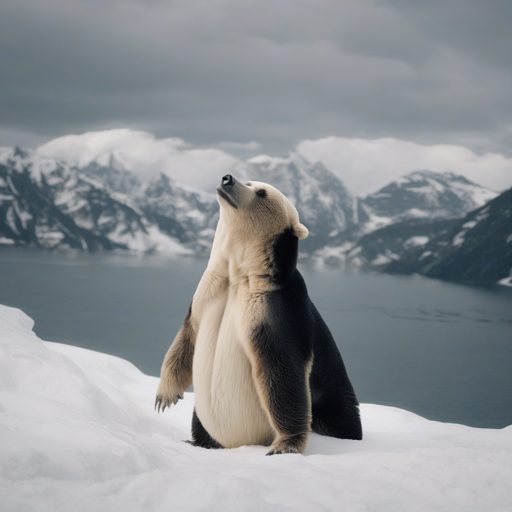}
				\captionsetup{font=ablationfont,justification=raggedright, singlelinecheck=false}
				\caption*{W/O Softbox \\ T:[5,30] \\ Box:[0.0,0.0,0.0,0.0]}
			\end{subfigure}
		};
		\node[anchor=north west] (img13) at (2*\ablationfiggap,-\ablationhgap+0.2*\ablationhgap) {
			\begin{subfigure}[t]{\ablationpicsize}
				\includegraphics[width=\textwidth]{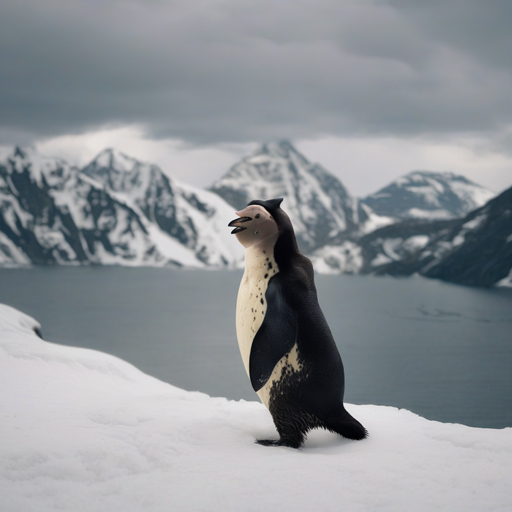}
				\captionsetup{font=ablationfont,justification=raggedright, singlelinecheck=false}
				\caption*{W/O Softbox \\ T:[8,20] \\ Box:[0.0,0.0,0.0,0.0]}
			\end{subfigure}
		};
		\node[anchor=north west] (img14) at (3*\ablationfiggap,-\ablationhgap+0.2*\ablationhgap) {
			\begin{subfigure}[t]{\ablationpicsize}
				\includegraphics[width=\textwidth]{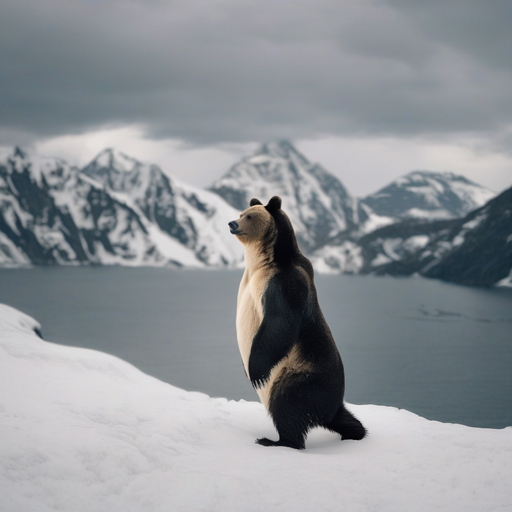}
				\captionsetup{font=ablationfont,justification=raggedright, singlelinecheck=false}
				\caption*{W/O Softbox \\ T:[8,30] \\ Box:[0.0,0.0,0.0,0.0]}
			\end{subfigure}
		};
		\node[anchor=north west] (img15) at (4*\ablationfiggap,-\ablationhgap+0.2*\ablationhgap) {
			\begin{subfigure}[t]{\ablationpicsize}
				\includegraphics[width=\textwidth]{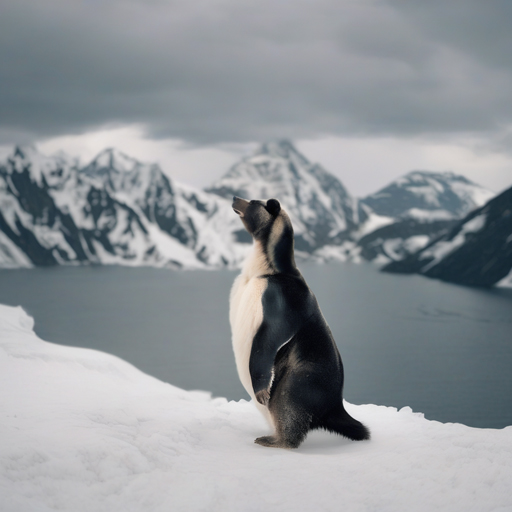}
				\captionsetup{font=ablationfont,justification=raggedright, singlelinecheck=false}
				\caption*{W/O Softbox \\ T:[15,30] \\ Box:[0.0,0.0,0.0,0.0]}
			\end{subfigure}
		};
		\node[anchor=north west] (img16) at (5*\ablationfiggap,-\ablationhgap+0.2*\ablationhgap) {
			\begin{subfigure}[t]{\ablationpicsize}
				\includegraphics[width=\textwidth]{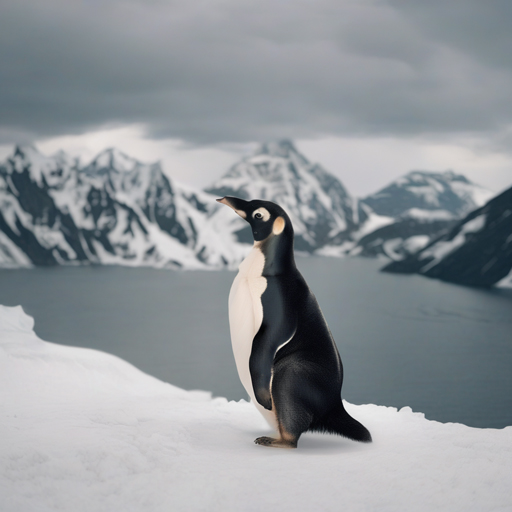}
				\captionsetup{font=ablationfont,justification=raggedright, singlelinecheck=false}
				\caption*{W/O Softbox \\ T:[20,30] \\ Box:[0.0,0.0,0.0,0.0]}
			\end{subfigure}
		};
		\node[anchor=north west] (img21) at (0*\ablationfiggap,-2*\ablationhgap+0.2*\ablationhgap) {
			\begin{subfigure}[t]{\ablationpicsize}
				\includegraphics[width=\textwidth]{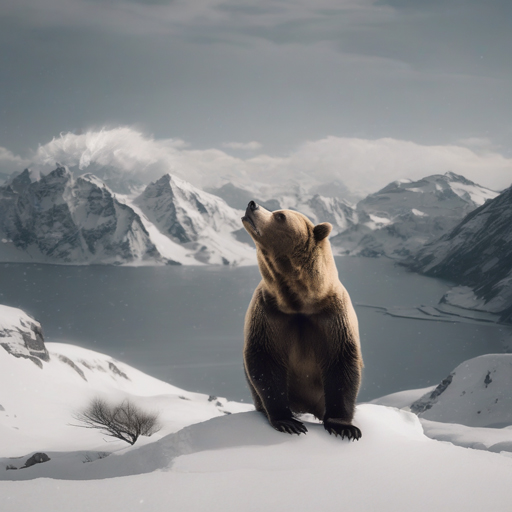}
				\captionsetup{font=ablationfont,justification=raggedright, singlelinecheck=false}
				\caption*{T:[0,30] \\ Box:[0.4,0.7,0.4,0.6]}
			\end{subfigure}
		};
		\node[anchor=north west] (img22) at (1*\ablationfiggap,-2*\ablationhgap+0.2*\ablationhgap) {
			\begin{subfigure}[t]{\ablationpicsize}
				\includegraphics[width=\textwidth]{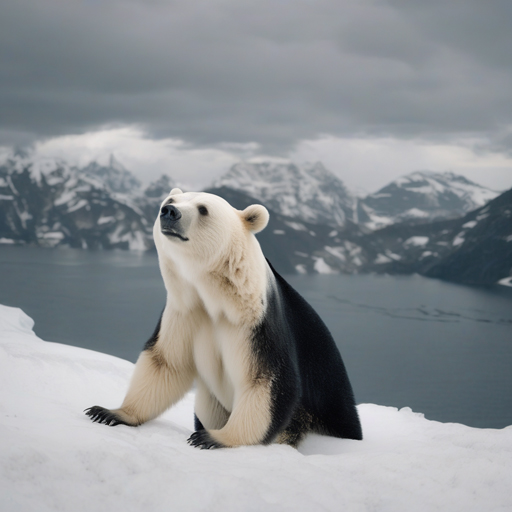}
				\captionsetup{font=ablationfont,justification=raggedright, singlelinecheck=false}
				\caption*{T:[5,30] \\ Box:[0.4,0.7,0.4,0.6]}
			\end{subfigure}
		};
		\node[anchor=north west] (img23) at (2*\ablationfiggap,-2*\ablationhgap+0.2*\ablationhgap) {
			\begin{subfigure}[t]{\ablationpicsize}
				\includegraphics[width=\textwidth]{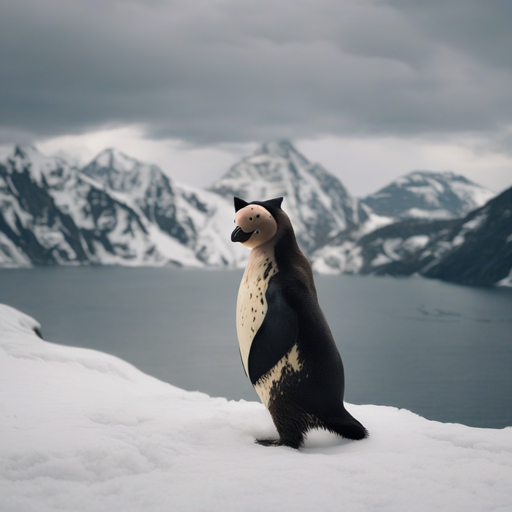}
				\captionsetup{font=ablationfont,justification=raggedright, singlelinecheck=false}
				\caption*{T:[8,20] \\ Box:[0.4,0.7,0.4,0.6]}
			\end{subfigure}
		};
		\node[anchor=north west] (img24) at (3*\ablationfiggap,-2*\ablationhgap+0.2*\ablationhgap) {
			\begin{subfigure}[t]{\ablationpicsize}
				\includegraphics[width=\textwidth]{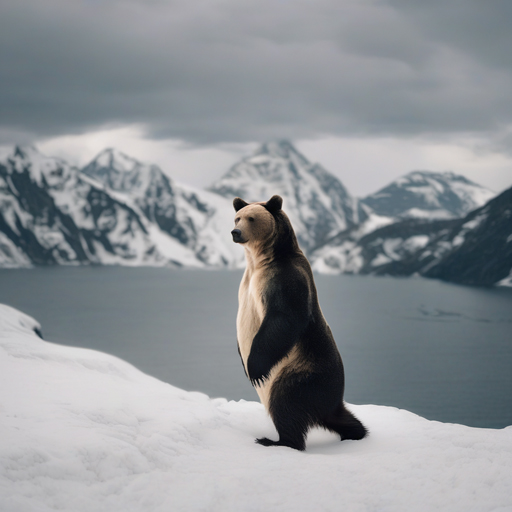}
				\captionsetup{font=ablationfont,justification=raggedright, singlelinecheck=false}
				\caption*{T:[8,30] \\ Box:[0.4,0.7,0.4,0.6]}
			\end{subfigure}
		};
		\node[anchor=north west] (img25) at (4*\ablationfiggap,-2*\ablationhgap+0.2*\ablationhgap) {
			\begin{subfigure}[t]{\ablationpicsize}
				\includegraphics[width=\textwidth]{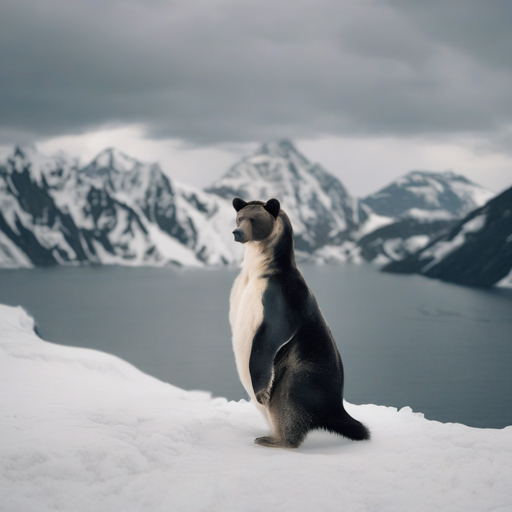}
				\captionsetup{font=ablationfont,justification=raggedright, singlelinecheck=false}
				\caption*{T:[15,30] \\ Box:[0.4,0.7,0.4,0.6]}
			\end{subfigure}
		};
		\node[anchor=north west] (img26) at (5*\ablationfiggap,-2*\ablationhgap+0.2*\ablationhgap) {
			\begin{subfigure}[t]{\ablationpicsize}
				\includegraphics[width=\textwidth]{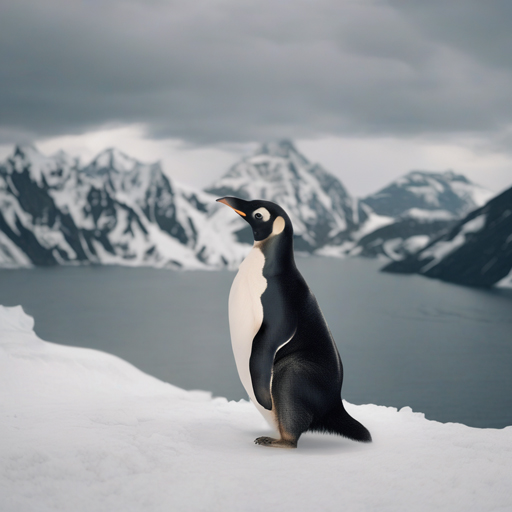}
				\captionsetup{font=ablationfont,justification=raggedright, singlelinecheck=false}
				\caption*{T:[20,30] \\ Box:[0.4,0.7,0.4,0.6]}
			\end{subfigure}
		};
		\node[anchor=north west] (img31) at (0*\ablationfiggap,-3*\ablationhgap+0.2*\ablationhgap) {
			\begin{subfigure}[t]{\ablationpicsize}
				\includegraphics[width=\textwidth]{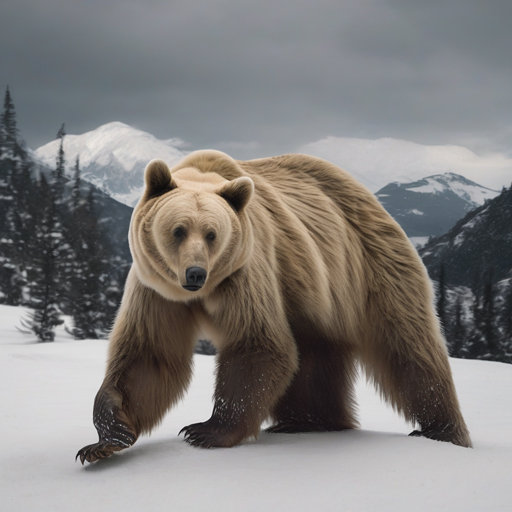}
				\captionsetup{font=ablationfont,justification=raggedright, singlelinecheck=false}
				\caption*{T:[0,30] \\ Box:[0.0,1.0,0.0,1.0]}
			\end{subfigure}
		};
		\node[anchor=north west] (img32) at (1*\ablationfiggap,-3*\ablationhgap+0.2*\ablationhgap) {
			\begin{subfigure}[t]{\ablationpicsize}
				\includegraphics[width=\textwidth]{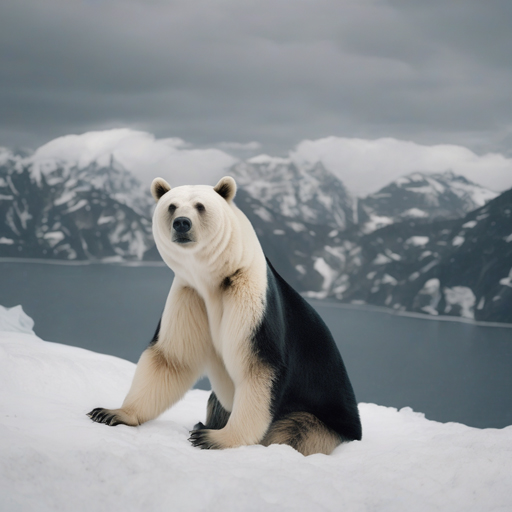}
				\captionsetup{font=ablationfont,justification=raggedright, singlelinecheck=false}
				\caption*{T:[5,30] \\ Box:[0.0,1.0,0.0,1.0]}
			\end{subfigure}
		};
		\node[anchor=north west] (img33) at (2*\ablationfiggap,-3*\ablationhgap+0.2*\ablationhgap) {
			\begin{subfigure}[t]{\ablationpicsize}
				\includegraphics[width=\textwidth]{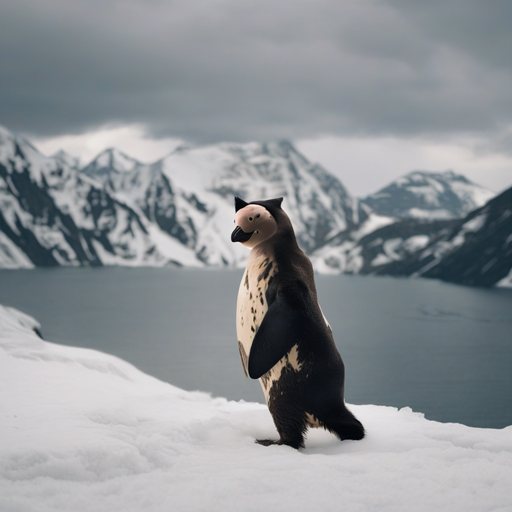}
				\captionsetup{font=ablationfont,justification=raggedright, singlelinecheck=false}
				\caption*{T:[8,20] \\ Box:[0.0,1.0,0.0,1.0]}
			\end{subfigure}
		};
		\node[anchor=north west] (img34) at (3*\ablationfiggap,-3*\ablationhgap+0.2*\ablationhgap) {
			\begin{subfigure}[t]{\ablationpicsize}
				\includegraphics[width=\textwidth]{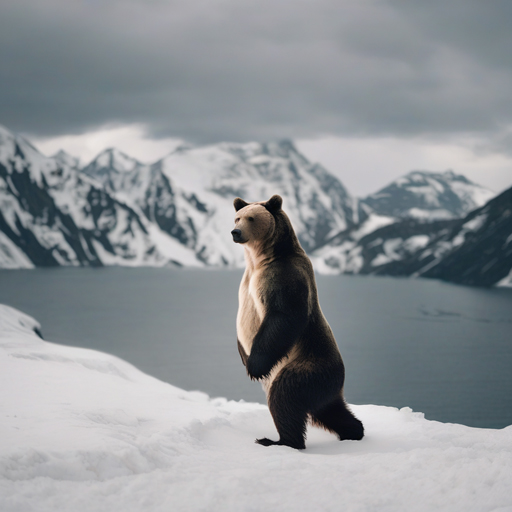}
				\captionsetup{font=ablationfont,justification=raggedright, singlelinecheck=false}
				\caption*{T:[8,30] \\ Box:[0.0,1.0,0.0,1.0]}
			\end{subfigure}
		};
		\node[anchor=north west] (img35) at (4*\ablationfiggap,-3*\ablationhgap+0.2*\ablationhgap) {
			\begin{subfigure}[t]{\ablationpicsize}
				\includegraphics[width=\textwidth]{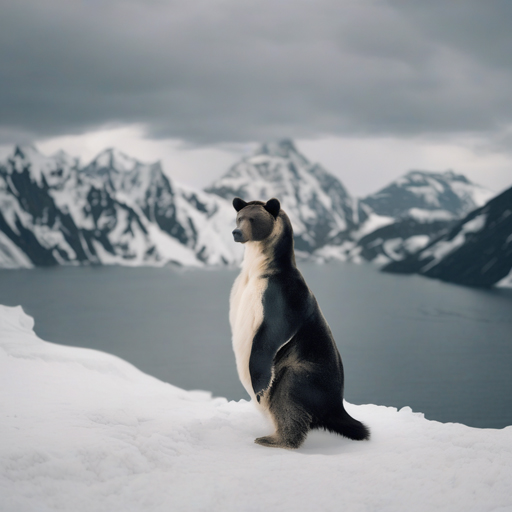}
				\captionsetup{font=ablationfont,justification=raggedright, singlelinecheck=false}
				\caption*{T:[15,30] \\ Box:[0.0,1.0,0.0,1.0]}
			\end{subfigure}
		};
		\node[anchor=north west] (img36) at (5*\ablationfiggap,-3*\ablationhgap+0.2*\ablationhgap) {
			\begin{subfigure}[t]{\ablationpicsize}
				\includegraphics[width=\textwidth]{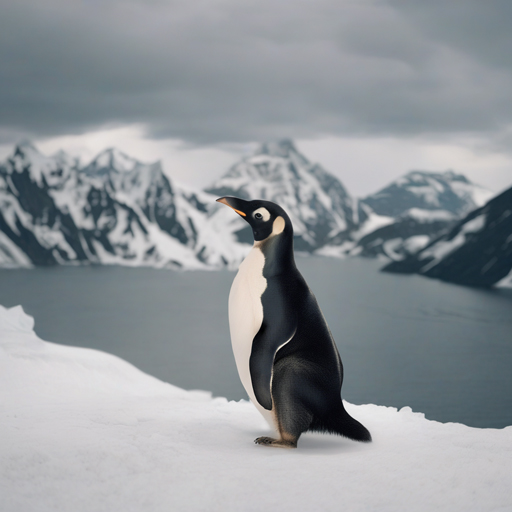}
				\captionsetup{font=ablationfont,justification=raggedright, singlelinecheck=false}
				\caption*{T:[20,30] \\ Box:[0.0,1.0,0.0,1.0]}
			\end{subfigure}
		};
		
		\draw[color=red,rounded corners,dashed,very thick] (3.05*\ablationfiggap,-2*\ablationhgap+0.2*\ablationhgap) rectangle (4.05*\ablationfiggap,-3.9*\ablationhgap+0.2*\ablationhgap);
	\end{tikzpicture}
    \vspace{-5pt}
	\caption{Ablation study. ``T'' represents the time step interval. ``Box'' represents the rectangular area defined by Softbox. The second row does not use Softbox. The red dashed box shows the impact of Softbox.}
	\label{fig:ablation2}
    \vspace{-15pt}
\end{figure}

\subsection{More Results}
\noindent \textbf{Object Replacement}. Our method enables precise object replacement in image generation, allowing direct editing of source prompts with arbitrary target prompts without requiring the same formatting constraints or identical random seeds as the source prompt. As illustrated in Fig.~\ref{fig:replace} (see Appendix A.5 for more results), this approach enables flexible replacement of any object while preserving the background, geometry, and semantics of other elements in the source image. For instance, we can replace ``horse'' with ``fox'' or ``dog'', swap ``sweater'' for ``suit'', and even substitute ``astronaut'' with ``spiderman''. Additionally, our method provides more granular control such as color adjustments, character replacements, and subtle expression modifications, as demonstrated with various examples in the figure. Our experiments yielded excellent results even for highly challenging fine-grained tasks. For example, we can alter an object's color from blue to green, or from black to red, while maintaining texture details. Similarly, subtle facial expression adjustments can be made without altering the character's distinct features, resulting in satisfactory and natural outcomes. Notably, our method also supports modifying character appearances in images by replacing names. We select well-known figures from various domains to ensure the model can understand the semantics of these individuals. 

\noindent \textbf{Object Addition}. Fig.~\ref{fig:add} highlights the exceptional capabilities of PSP in object addition. The first two columns demonstrate its ability to add ``sunglasses'' to various objects, showing remarkable detail and ensuring a perfect fit with the facial features. In the third column, we add a ``bow hairband'' to the ``woman'', achieving not merely the placement of the object on the head but also a natural interaction with the hair, reflecting subtle traces of the hair being tied. This indicates that our method effectively integrates added objects with the background, adhering to real-world logic and consistency. The fourth and fifth columns illustrate that even with identical backgrounds, our approach can accurately add specified objects to the image while maintaining background coherence. The sixth column presents additional examples, emphasizing our method’s capability to seamlessly blend novel objects with the background. This further validates that, in complex scenes, our technique ensures harmonious integration of newly added objects with the existing background, creating highly realistic images.

\noindent \textbf{Style Transfer}. In style transfer tasks, our method adjusts the overall artistic style of an image while preserving details. As shown in Fig.~\ref{fig:style}, it seamlessly achieves this by transforming the style and scene without altering the primary object. This ensures the core elements remain unchanged while infusing a new visual style.

\noindent \textbf{Real Image Editing}. Our PSP method seamlessly extends to real image editing via DDIM inversion~\cite{song2020denoising}. Fig.~\ref{fig:real} presents examples of real image editing. Our method excels in both style transfer and object replacement tasks.

\subsection{Ablation Study}
We conducted an ablation study to evaluate the effectiveness of key components in our method, including the time step interval and Softbox. Each is integrated sequentially, and the results, visually presented in Fig.~\ref{fig:ablation2}, validate their effectiveness. The Softbox is defined in the format \([h1, h2, w1, w2]\), where the region enclosed by the coordinates \((h1, w1)\) and \((h2, w2)\) represents the box area. These coordinates are represented as proportions of width and height, such as \([0.1, 0.4, 0.2, 0.4]\). An appropriate Softbox constraint prevents background distortion during object replacement while ensuring high fidelity of the generated object. Therefore, our Softbox is typically obtained using a segmentation model. Improper selection of the time step interval can lead to significant issues: setting it too early may cause the final image layout to diverge significantly from the source image, while setting it too late can hinder effective object replacement. The synergy of these components enhances our method’s performance in image editing tasks.

\section{Conclusion}
In this paper, we conduct a comprehensive investigation into the role of key components in text embeddings and derive three key insights. Building on these insights, we propose a novel training-free image editing method called Prompt-Softbox-Prompt (PSP), which leverages free-form text embeddings for object manipulation without requiring format alignment with source prompts. Our method enables controllable image editing by inserting, replacing, or removing text embeddings within cross-attention layers and employing the Softbox mechanism to precisely constrain the injection region of object semantics, thereby preserving the image background. Extensive experiments have demonstrated the effectiveness and practicality of our method across various tasks. 


\begin{acks}
This project is sponsored by Shanghai Pujiang Programme 24PJD030 and Natural Science Foundation of Shanghai 25ZR1402138.
\end{acks}

\bibliographystyle{ACM-Reference-Format}
\balance
\bibliography{sample-base}


\end{document}